\documentclass[singlespacetitle,tocpage]{usmthesis}

\usepackage{pdfpages}
\usepackage{afterpage}
\usepackage{import}
\usepackage{scrextend}
\changefontsizes{13pt}
\usepackage{marvosym}
\usepackage{enumitem}
\usepackage{tabu}
\usepackage{booktabs}
\usepackage{calligra}
\usepackage{multirow}
\usepackage{multicol}
\usepackage{bm}
\usepackage{bibentry}
\usepackage{listings}
\usepackage{algpseudocode}
\usepackage{algorithm}
\usepackage{indentfirst}
\usepackage{caption}
\usepackage{graphicx}
\usepackage{amssymb}
\usepackage{pifont}
\usepackage{svg}
\usepackage{adjustbox}
\usepackage{multibib}
\usepackage{soul}
\usepackage{ulem}
\newcommand{\etal}{\textit{et al. }}
\newcolumntype{M}[1]{>{\centering\arraybackslash}m{#1}}

\author{HUỲNH LÂM HẢI ĐĂNG \hspace{5mm} 19125003 \\ HỒ THỊ NGỌC PHƯỢNG \hspace{5mm} 19125014}
\title{ENHANCING VIDEO SUMMARIZATION WITH CONTEXT AWARENESS}
\submityear{2023}
\submitmonth{July}
\degreetype{BACHELOR OF SCIENCE IN COMPUTER SCIENCE}
\lecturer{ASSOC. PROF. TRẦN MINH TRIẾT \\ DR. LÊ TRUNG NGHĨA}
\newcites{own}{List of Publications}

\usepackage[plainpages=false,bookmarksnumbered,
bookmarksdepth=section,breaklinks=true]{hyperref}
\hypersetup{
    colorlinks,
    citecolor=black,
    filecolor=black,
    linkcolor=black,
    urlcolor=black
}

\ifpdf
  \makeatletter
  \hypersetup{hypertexnames=false,%
    pdfauthor={\@author},pdftitle={\@title}}
  \makeatother
\fi

\usepackage{ragged2e}
\usepackage{tikz}
\usepackage{multido}
\usetikzlibrary{calc}
\usepackage{longtable}
\usepackage{makecell}
\usepackage{url}
\usepackage{breakurl}

\begin{document}
\bibliographystyle{ieeetr}
\bibliographystyleown{ieeetr}

\makecover
\frontmatter

\chapter{Acknowledgements}
\label{chapter:acknowledgements}

We extend our deepest gratitude to our first advisor, Associate Professor Tran Minh Triet, for his invaluable guidance and unwavering support, which significantly influenced the direction and outcomes of this research.

Our sincere appreciation goes to our second advisor, Dr. Le Trung Nghia, for his expertise and constructive criticism, enriching the research process and outcomes.

We also thank Doctor Nguyen Ngoc Thao for her invaluable feedback, which played a pivotal role in refining our study.

Acknowledgments to the Faculty of Information and Technology at the University of Science, VNU-HCM, for their unwavering support and commitment to academic excellence.

Heartfelt thanks to the participants whose insights added depth to our findings.

Lastly, our families and friends deserve our deepest gratitude for their unwavering support throughout this academic journey.

To all those mentioned and those inadvertently omitted, we sincerely thank you for your indispensable contributions to our academic and personal growth.
\begin{flushright}
  \begin{minipage}{10cm}
  \centering
  Authors \\
  Huỳnh Lâm Hải Đăng \& Hồ Thị Ngọc Phượng
  \end{minipage}
\end{flushright}

\tableofcontents \clearpage
\listoftables \clearpage
\listoffigures \clearpage
\setlength\intextsep{24pt}

\begin{EnAbstract}
    Video summarization is a crucial research area that aims to efficiently browse and retrieve relevant information from the vast amount of video content available today. With the exponential growth of multimedia data, the ability to extract meaningful representations from videos has become essential. Video summarization techniques automatically generate concise summaries by selecting keyframes, shots, or segments that capture the video's essence. This process improves the efficiency and accuracy of various applications, including video surveillance, education, entertainment, and social media.
    
    Despite the importance of video summarization, there is a lack of diverse and representative datasets, hindering comprehensive evaluation and benchmarking of algorithms. Existing evaluation metrics also fail to fully capture the complexities of video summarization, limiting accurate algorithm assessment and hindering the field's progress.
    
    To overcome data scarcity challenges and improve evaluation, we propose an unsupervised approach that leverages video data structure and information for generating informative summaries. By moving away from fixed annotations, our framework can produce representative summaries effectively.
    
    Moreover, we introduce an innovative evaluation pipeline tailored specifically for video summarization. Human participants are involved in the evaluation, comparing our generated summaries to ground truth summaries and assessing their informativeness. This human-centric approach provides valuable insights into the effectiveness of our proposed techniques.

    Experimental results demonstrate that our training-free framework outperforms existing unsupervised approaches and achieves competitive results compared to state-of-the-art supervised methods.
    
\end{EnAbstract}

\mainmatter
\chapter{Introduction}
\label{chapter:introduction}

\begin{ChapAbstract}
    In this chapter, we provide general information about our work in four sections before getting into details in the following chapters. Section \ref{section:intro-overview} introduces the practicality and applicability of Video Summarization. We then discuss our motivation for applying unsupervision and introducing a new evaluation metric in Section \ref{section:intro-motivation}. Section \ref{section:intro-objectives} presents our objectives in developing the model as well as the evaluation pipeline. Finally, we describe the outline content of our work in Section \ref{section:intro-content}.
\end{ChapAbstract}

\section{Overview}
\label{section:intro-overview}

In recent years, the consumption of video content has experienced a remarkable upsurge, driven by the proliferation of multimedia platforms such as TikTok, YouTube, Instagram, and others. A striking example of this growth can be observed in the case of YouTube, where the number of video content hours uploaded per minute has witnessed a substantial increase. Between 2014 and 2020, there was an approximate 40 percent rise in the rate of uploads, with over 500 hours of video being uploaded every minute as of June 2022 \cite{YoutubeHours}. This surge in video content on platforms like YouTube reflects the expanding demand among consumers for online video consumption. With an approximation of 2.5 quintillion bytes of data created every day \cite{Meena2023Review}, there is a pressing need for effective methods that can automatically generate concise and informative summaries of videos, enabling users to quickly comprehend the content without having to watch the entire video. 

Video summarization, as a research area, focuses on generating concise summaries that effectively capture the temporal and semantic aspects of a video, while preserving its salient content. Achieving this objective involves addressing several fundamental challenges, such as identifying key frames or representative shots, detecting important events, recognizing significant objects or actions, and preserving the overall context and coherence of the video. 

The task plays a crucial role in facilitating efficient browsing, indexing, and retrieval of video data, offering users the ability to preview and comprehend video content without investing significant time and effort. Moreover, video summarization finds applications in various domains, including video surveillance, multimedia retrieval, video archiving, and online video platforms, where it serves as a valuable tool for enhancing user experience and content accessibility \cite{Apostolidis2021Video}.

\section{Motivations}
\label{section:intro-motivation}

Despite the widespread usage of video summarization, the availability of datasets for this task remains limited. Currently, there are only a few prominent datasets available, namely SumMe \cite{SumMe} and TVSum \cite{TVSum}. This scarcity of diverse and representative datasets poses a challenge for comprehensive evaluation and benchmarking of video summarization algorithms. The lack of varied datasets restricts the ability to assess the generalizability and effectiveness of developed techniques.

Supervised approaches for video summarization face difficulties due to the nature of the task. Traditional evaluation metrics, such as F-measure and precision-recall curves, heavily rely on frame-level matching. However, these metrics do not adequately account for the temporal coherence and semantic understanding required in generating high-quality video summaries. The limitations of these metrics make it challenging to fully capture the inherent complexities and challenges involved in the video summarization process.

Moreover, supervised approaches heavily rely on large amounts of annotated data. However, annotating video summaries is a labor-intensive process, making it challenging to collect a sufficient quantity of annotated data for training and evaluation purposes. This scarcity of annotated data further limits the effectiveness and scalability of supervised approaches in video summarization.

These challenges highlight the need for the development of more diverse and representative datasets for video summarization. Additionally, there is a demand for the exploration and adoption of evaluation metrics that can better capture the temporal coherence and semantic understanding of video summaries. Finding alternative approaches to address the data scarcity issue, such as weakly supervised or unsupervised learning techniques, could also pave the way for advancements in video summarization research.

Through our research, we aim to contribute to the advancement of video summarization by investigating unsupervised learning techniques and developing an evaluation pipeline that captures the nuanced aspects of video summarization. Our work strives to overcome the limitations of traditional supervised approaches and evaluation metrics, ultimately leading to more effective and robust video summarization algorithms.

\section{Objectives and Main Contributions}
\label{section:intro-objectives}

    In this work, we aim to develop an unsupervised video summarization approach that eliminates the need for labor-intensive annotations. By leveraging deep pre-trained models to extract visual representations, our goal is to create a framework capable of generating comprehensive video summaries from unlabeled video data. This approach addresses the challenges of data scarcity and reduces reliance on annotated datasets.

    Furthermore, we propose a novel evaluation pipeline customized for video summarization. Understanding the limitations of conventional evaluation metrics, we design a unique framework that incorporates human evaluation. This mimics how humans summarize videos into short-form content, considering subjective factors like semantic relevance, coherence, and overall quality, which are challenging to quantify objectively. Integrating human judgment enhances the assessment of video summarization algorithms, providing a more comprehensive and reliable evaluation.

    Our main contributions are as follows:
    
    \begin{itemize}
      \item Introduction of a novel \textit{unsupervised method} explicitly designed for video summarization. Despite lacking learning aspects, our model outperforms existing unsupervised methods and approximates the performance of state-of-the-art supervised approaches.
      \item Proposal of an \textit{evaluation pipeline tailored to human-centric criteria}. This pipeline goes beyond traditional evaluation metrics, incorporating aspects more relevant and meaningful to human viewers. It ensures that the generated summaries align with human preferences and expectations.
    \end{itemize}

\section{Project Content}
\label{section:intro-content}


    After \textbf{Chapter \ref{chapter:introduction}: Introduction}, the remainder of our work is composed of 5 chapters as follows:



    \textbf{Chapter \ref{chapter:related}: Literature Review}

        This chapter first provides an overview of three main deep learning approaches for solving video summarization task: supervised methods, weakly supervised methods, and unsupervised methods. At each approach, we discuss the leading papers and explain how the follow-up papers could improve the baseline in many aspects. Finally, we analyze the advantages and disadvantages of each method. 

    \textbf{Chapter \ref{chapter:method}: Context-Aware Video Summarization}

        A detailed explanation of our method is described in the chapter. We begin by outlining the overall motivation and intuition behind our approach. Subsequently, we delve into the specifics of our proposed architecture and analyze its components from various perspectives.

    \textbf{Chapter \ref{chapter:experiments}: Experiments}

        After exploring the possible solution for the aforementioned problems, we conduct comprehensive experiments to evaluate our proposed method using both qualitative and quantitative approaches. We compare our approach with state-of-the-art architectures in terms of accuracy and efficiency, highlighting its strengths. Furthermore, we provide a detailed and precise ablation study to further validate the effectiveness of our method empirically.

    \textbf{Chapter \ref{chapter:conclusion}: Conclusions}

        At the end of this report, we summarize our work and briefly discuss the disadvantages of the current approach to pave the way for future research.
\chapter{Literature Review}
\label{chapter:related}

\begin{ChapAbstract}
  Deep learning methods have dominated the video summarization task for a long time due to their remarkable ability to automatically learn relevant features and representations from large-scale video data. In this chapter, we initially cover the fundamental aspects, encompassing the problem statement, datasets used, and evaluation metrics employed in video summarization research in Section \ref{section:rel-preliminary}. Subsequently, we conduct a thorough examination of the existing literature in video summarization, emphasizing three principal categories of approaches: supervised methods (Section \ref{section:rel-supervised}), unsupervised methods (Section \ref{section:rel-unsupervised}), and weakly supervised methods (Section \ref{section:rel-weakly}).
\end{ChapAbstract}

\section{Preliminary}
\label{section:rel-preliminary}

In this section, we will provide a foundation of fundamental knowledge upon which we will build our proposal. This preliminary discussion encompasses the problem statement, problem formulation, the datasets utilized and the evaluation metrics employed in previous works.

\subsection{Problem Statement}
\label{subsec:rel-statement}

Video summarization aims to generate a concise overview of video content by selecting the most informative and significant parts. The resulting summary can take the form of either a set of representative video frames, known as a video storyboard, or a compilation of video fragments stitched together in chronological order, referred to as a video skim. Video skims have an advantage over static frame sets as they can include audio and motion elements, allowing for a more natural storytelling experience and potentially conveying more information. Moreover, watching a video skim is often more engaging and captivating for viewers compared to a slideshow of frames \cite{Li2001Overview}. On the other hand, storyboards offer greater flexibility in terms of data organization for browsing and navigation purposes, as they are not bound by timing or synchronization constraints \cite{Calic2007Comic,Wang2007VideoCollage}.

Our problem statement aligns closely with the concept of video storyboards, which involve selecting a subset of representative video frames to summarize the content. By focusing on these key frames, we aim to capture the essence and important aspects of the video in a condensed form. This approach allows for efficient browsing and navigation through the video data while providing a comprehensive overview of its content.

\subsection{Problem Formulation}
\label{subsec:rel-formulation}

In the problem of video summarization, we are given an input video $\textbf{I}=\{I_t\}_{t=1}^T$, where each frame $I_t \in \mathbb{R} ^{W \times H \times C}$ has $C$ channels, and its height and width are denoted by $H$ and $W$ respectively. The objective of a video summarizer is to generate a concise summary $\textbf{S}$ that contains a subset of representative frames from the input video. The summary $\textbf{S}$ is represented as $\textbf{S} = \{I_{t_i}\}_{i=1}^L$, where $L$ is typically much smaller than $T$, and the frame indices are arranged in ascending order, i.e., $t_i < t_{i + 1}$ for all valid \(i \in [1, L)\). The goal is to select a set of frames that effectively capture the essence of the video content and convey the most relevant information in a compact form.

\subsection{Datasets}
\label{subsec:rel-datasets}

As referenced in Section \ref{section:intro-motivation}, two datasets that prevail in the video summarization bibliography are SumMe \cite{SumMe} and TVSum \cite{TVSum}. 

SumMe dataset comprises 25 videos, ranging from 1 to 6 minutes in duration, encompassing diverse content and captured from both first-person and third-person perspectives. Each video has been annotated by 15 to 18 users, resulting in multiple fragment-level user summaries. These summaries typically span 5\% to 15\% of the original video duration. 

TVSum dataset comprises 50 videos, with durations ranging from 1 to 11 minutes. These videos cover content from 10 categories of the TRECVid MED dataset. Each video in TVSum has been annotated by 20 users, providing shot- and frame-level importance scores on a scale of 1 to 5.

In addition to SumMe and TVSum, two common datasets for evaluating video summaries are OVP \cite{De2011VSUMM} and YouTube \cite{De2011VSUMM}. Each dataset comprises 50 videos, with annotations consisting of sets of key-frames generated by 5 users. The video durations span from 1 to 4 minutes for OVP and 1 to 10 minutes for YouTube. These datasets encompass a wide variety of video content, including documentaries, educational videos, ephemeral videos, historical footage, and lectures in the case of OVP, and cartoons, news clips, sports highlights, commercials, TV shows, and home videos in the case of YouTube.

Considering the size of these datasets, it is evident that there is a scarcity of large-scale annotated datasets, which limits their utility in enhancing the training of sophisticated supervised deep learning architectures.

Some less commonly used datasets for video summarization include CoSum \cite{Chu2015CoSum}, MED-summaries \cite{Potapov2014MEDSummaries}, Video Titles in the Wild (VTW) \cite{Zeng2016TitleWild}, League of Legends (LoL) \cite{Fu2017VideoLoL}, and FVPSum \cite{Ho2018FVPSum}. 

CoSum focuses on video co-summarization. It consists of 50 videos obtained from Youtube using 10 query terms related to the content of SumMe dataset. Each video has an approximate duration of 4 minutes, from which sets of key-fragments are selected by 3 different annotators.

MED-Summaries consists of 160 videos from TRECVID 2011 MED dataset. The dataset is divided into a validation set with 60 videos from 15 event categories and a test set with 100 videos from 10 event categories. The majority of videos has durations range from 1 to 5 minutes, with each being annotated with a set of importance scores, averaged over 1 to 4 annotators.

The VTW dataset consists of 18100 open domain videos, out of which 2000 videos are annotated at the sub-shot level with highlight scores. These user-generated videos are untrimmed and typically contain a highlight event. On average, the videos in the dataset have a duration of 1.5 minutes.

The LoL dataset consists of 218 long videos, ranging from 30 to 50 minutes in duration. These videos showcase game matches from the North American League of Legends Championship Series (NALCS). The annotations for this dataset are derived from a YouTube channel that features community-generated highlights, with the highlight videos typically having a duration of 5 to 7 minutes. As a result, there is one set of key-fragments available for each video in the dataset.

The FPVSum dataset focuses on first-person video summarization and comprises 98 videos, totaling over 7 hours of content. These videos are sourced from 14 categories of GoPro viewer-friendly videos. For each category, approximately 35\% of the video sequences have been annotated with ground-truth scores by at least 10 users, while the remaining sequences are considered unlabeled examples. This dataset provides valuable resources for evaluating and developing first-person video summarization algorithms.

Apostolidis \etal~\cite{Apostolidis2021Video} have compiled a comprehensive summarization table, showcasing the main characteristics of the aforementioned datasets. For reference, Table \ref{table:dataset-characteristics} presents an overview of the dataset attributes, such as video count, annotation types, video duration, and user involvement.
\begin{table}
  \caption{Datasets for video summarization and their characteristics.}
  \scriptsize
\begin{tabular}{|c|c|c|c|c|c|}
\hline
\textbf{Dataset}                                                             & \textbf{no. videos} & \textbf{\begin{tabular}[c]{@{}c@{}}duration \\ (min)\end{tabular}} & \textbf{content}                                                                                                          & \textbf{\begin{tabular}[c]{@{}c@{}}type of \\ annotations\end{tabular}}    & \textbf{\begin{tabular}[c]{@{}c@{}}annotators \\ per video\end{tabular}} \\ \hline
\begin{tabular}[c]{@{}c@{}}SumMe \\ \cite{SumMe}\end{tabular}                & 25                  & 1 - 6                                                              & holidays, events, sports                                                                                                  & \begin{tabular}[c]{@{}c@{}}multiple sets of \\ key-fragments\end{tabular}  & 15 - 18                                                                  \\ \hline
\begin{tabular}[c]{@{}c@{}}TVSum \\ \cite{TVSum}\end{tabular}                & 50                  & 2 - 10                                                             & \begin{tabular}[c]{@{}c@{}}news, how-tos, \\ user-generated, documentaries\\ (10 categories - 5 videos each)\end{tabular} & \begin{tabular}[c]{@{}c@{}}multiple \\ fragment level scores\end{tabular}  & 20                                                                       \\ \hline
\begin{tabular}[c]{@{}c@{}}OVP \\ \cite{De2011VSUMM}\end{tabular}            & 50                  & 1 - 4                                                              & \begin{tabular}[c]{@{}c@{}}documentary, educational, \\ ephemeral, historical, lecture\end{tabular}                       & \begin{tabular}[c]{@{}c@{}}multiple sets of \\ key frames\end{tabular}     & 5                                                                        \\ \hline
\begin{tabular}[c]{@{}c@{}}YouTube \\ \cite{De2011VSUMM}\end{tabular}        & 50                  & 1 - 10                                                             & \begin{tabular}[c]{@{}c@{}}cartoons, sports, tv-shows, \\ commercial. home videos\end{tabular}                            & \begin{tabular}[c]{@{}c@{}}multiple sets of \\ key frames\end{tabular}     & 5                                                                        \\ \hline
\begin{tabular}[c]{@{}c@{}}CoSum \\ \cite{Chu2015CoSum}\end{tabular}         & 51                  & ~ 4                                                                & \begin{tabular}[c]{@{}c@{}}holidays, events, \\ sports (10 categories)\end{tabular}                                       & \begin{tabular}[c]{@{}c@{}}multiple sets of \\ key fragments\end{tabular}  & 3                                                                        \\ \hline
\begin{tabular}[c]{@{}c@{}}MED\\ \cite{Potapov2014MEDSummaries}\end{tabular} & 160                 & 1 - 5                                                              & 15 categories of various genres                                                                                           & \begin{tabular}[c]{@{}c@{}}one set of \\ importance score\end{tabular}     & 1 - 4                                                                    \\ \hline
\begin{tabular}[c]{@{}c@{}}VTW \\ \cite{Zeng2016TitleWild}\end{tabular}      & 2000                & 1.5 (avg)                                                          & \begin{tabular}[c]{@{}c@{}}user-generated videos \\ that contain a highlight event\end{tabular}                           & \begin{tabular}[c]{@{}c@{}}sub-shot level \\ highlight scores\end{tabular} & -                                                                        \\ \hline
\begin{tabular}[c]{@{}c@{}}LoL \\ \cite{Fu2017VideoLoL}\end{tabular}         & 218                 & 30 - 50                                                            & \begin{tabular}[c]{@{}c@{}}matches from \\ LoL tournament\end{tabular}                                                    & \begin{tabular}[c]{@{}c@{}}one set of\\ key fragments\end{tabular}         & 1                                                                        \\ \hline
\begin{tabular}[c]{@{}c@{}}FPVSum \\ \cite{Ho2018FVPSum}\end{tabular}        & 98                  & 4.3 (avg)                                                          & first-person videos (14 categories)                                                                                       & \begin{tabular}[c]{@{}c@{}}multiple \\ frame level scores\end{tabular}     & 10                                                                       \\ \hline
\end{tabular}
  \label{table:dataset-characteristics}
\end{table}


\subsection{Evaluation Metrics}
\label{subsec:rel-evaluation}
    
    The evaluation of video summarization algorithms is a complex task due to the inherent difficulty in quantifying the quality of a summary. This section aims to explore the challenges encountered by previous research when assessing video summarization algorithms. Additionally, it provides an overview of commonly used evaluation metrics in video summarization, including those utilized in this study (as discussed in sections \ref{subsubsec:rel-evaluation-difficulties} and \ref{subsubsec:rel-evaluation-prior}). These details offer valuable insights into the current research progress on the evaluation process of video summarization algorithms. By understanding these challenges and metrics, researchers can better grasp the strengths and limitations of different evaluation approaches, thus contributing to the advancement of video summarization evaluation methodologies.
	
	\subsubsection{Difficulties in Evaluation}
	\label{subsubsec:rel-evaluation-difficulties}
		The evaluation of video summarization algorithms presents significant challenges, including the absence of high-quality ground-truth summaries, the subjective nature of human perception, and the absence of a consensus on what constitutes a good summary. These issues greatly impact the evaluation process and the ability to accurately assess the performance of video summarization algorithms. In-depth details regarding these challenges can be found in \cite{Apostolidis2021Video}, providing valuable insights into the complexities involved in evaluating video summarization algorithms. Addressing these challenges is crucial for advancing the field and developing more robust evaluation methodologies.

			\paragraph[short]{Lack of high-quality ground-truth summaries}
				The lack of high-quality ground-truth summaries is one of the main problems when evaluating video summarization algorithms. Ground-truth summaries are summaries that are created by humans and are used as a reference for the evaluation of automatic video summarization algorithms. However, the construction of such annotated summaries is a time-consuming and expensive process as it requires the involvement of human annotators which are inconsistent in nature. This inconsistency of human annotators means that the same evaluator may produce different summaries for the same video at different times, leading to unsure and possibly conflicting ground-truth summaries among the annotations from the same evaluator, left alone the annotations from different evaluators as provided in the datasets from previous Subsection \ref{subsec:rel-datasets}.

				Besides the inconsistency issue, the ground-truth summaries are also limited in quantity. This is because the creation of ground-truth summaries is a time-consuming and expensive process while only a small number of videos were annotated with a limited number of annotators in the previously published datasets.
			
			\paragraph[short]{Subjectivity of human perception}
				Different people may have different opinions on what constitutes a good summary for a given video. This subjectivity makes it difficult to evaluate the performance of an automatic video summarizer as it can lead to different ground-truth summaries for the same video which in turn creates distinct and possibly conflicting scores or opinions on the quality of a summary. Furthermore, perceptive subjectivity also possesses a problem in comparing the performance of different automatic video summarization algorithms due to several corner cases such as when an algorithm produces a summary that is judged as good by some of the human evaluators but not by others, while the other algorithm produces a summary that is judged vice versa. This problem is also known as the \textit{inter-annotator agreement} problem that is described by both \cite{measure-annotator-agreement} and \cite{inter-annotator-agreement} in detail.
			
			\paragraph[short]{Lack of consensus on the definition of a good summary}
				Different people may have different opinions on what constitutes a good summary. This lack of consensus can make it difficult to evaluate the performance of automatic video summarization algorithms.

		Other than the problems that are already described in the previous research, there are also other problems that are not yet addressed in the evaluation process of video summarization algorithms. A notable problem that our team found during the research for prior evaluation is the possibility of several semantically different summaries that can well represent the same video. This problem is only \textit{partially} addressed in the SumMe dataset with the use of specialized aggregation method on multiple ground-truth summaries but most of this problem still persists as the number of available ground-truths is still limited.

	\subsubsection{Prior Evaluation Methods}
	\label{subsubsec:rel-evaluation-prior}
		There are several methods that have been used in literature to evaluate video summarization algorithms while addressing some of the problems mentioned in the subsection above. The two most employed methods include the approach of user studies which is the most naive and original one, and the use of ground-truth summaries as references for computation of objective metrics. Details about these methods are presented in \cite{Apostolidis2021Video} and we provide a brief overview of them in the following paragraphs.

		\paragraph[short]{User studies}
			The most naive and original method for evaluating video summarization algorithms is to conduct user studies. In this method, the performance of an algorithm is evaluated by asking human evaluators to watch the video summaries produced by the algorithm and then rate the quality of the summaries. The quality of a summary is usually rated by the evaluators based on their subjective opinions. This method is the most naive and original one because it is the most straightforward way to evaluate the performance of an algorithm. Furthermore, it is also the most expensive and time-consuming method as it requires the involvement of human evaluators. Besides such disadvantages, this method is also the least accurate one as the human evaluators are inherently inconsistent in nature. This inconsistency of human evaluators means that the same evaluator may produce distinct scores for the same summary at different times, making such evaluation process not possible to be reproduced in the future. Therefore, the current literature has moved away from this method for easier reproducibility as well as consistency and low-cost evaluation of their methods. More details can be found in the survey by Apostolidis et al. \cite{Apostolidis2021Video}.

		\paragraph[short]{Objective metrics}
			Another method that has been increasingly used in literature to evaluate automatically generated summaries is the use of artificially annotated ground-truths as references for computation of objective metrics. In this method, the performance of an algorithm is evaluated by comparing the summary produced by the algorithm with the pre-defined ground-truth summaries created by human evalutors. The comparison is usually done by computing the similarity between the summary produced by the algorithm and the ground-truth summaries. The similarity between the summary produced by the algorithm and the ground-truth summaries is usually computed with objective metrics such as accuracy and error rates which were proposed by \cite{ejaz2012adaptive} and adopted by \cite{almeida2012vison,cahuina2013new,jacob2017video}, or the more well-known precision, recall, and f-measure that were published by \cite{mahmoud2013unsupervised} and used by \cite{gong2014diverse,guan2014top,mei2015video,demir2015video}. This method is less expensive and time-consuming than the user studies method as it does not require the involvement of human evaluators.

			As all of the previously mentioned objective metrics are computed based on a fundamental assumption that all summaries, either automatically generated or artificially annotated, comprise of keyframes selected from the video content, their resulting performance measures are not soft enough to finely rank the performane of different methods summarizing the same video. This is because the keyframes selected from the video content by human evaluators are singular in its nature, meaning that an automatically generated keyframe falling aside an annotated one would be viewed as false positive without concerning the distance between the two. Hence, the use of such metrics would \textit{not} result in a difference of measured performance for approaches with different discrepancies to the user-generated keyframes. In other words, the algorithm that produces a summary with keyframes that are close to the annotated ones but not exactly the same is assigned a similar score with another algorithm producing a summary with keyframes that are far from the annotated ones.
			
			To address such problem, a notable variation of this method was introduced in the SumMe dataset \cite{SumMe} and adopted by TVSum dataset \cite{TVSum}, which 
			
			However, the methods of using objective metrics to evaluate the performance of video summaries is also less accurate than the user studies as it is based on the assumption that the ground-truth summaries are accurate representations of the videos. This assumption is not always true as the ground-truth summaries are usually created by human evaluators which are inconsistent in nature. This inconsistency of human evaluators means that the same evaluator may produce different summaries for the same video at different times, leading to unsure and possibly conflicting ground-truth summaries among the annotations from the same evaluator, leave alone the annotations from different evaluators as provided in the datasets from previous subsection \ref{subsec:rel-datasets}.

		To the best of our knowledge, there is currently no prior work that has fully addressed this problem and hence, solving
\section{Supervised approaches}
\label{section:rel-supervised}
Supervised methods rely on datasets with human-labeled ground-truth annotations. For example, the SumMe dataset \cite{SumMe} utilizes video summaries as ground truth, while the TVSum dataset \cite{TVSum} employs frame-level importance scores. By leveraging this labeled data, supervised approaches aim to learn the criteria for selecting video frames or fragments to construct effective video summaries.

\subsection{Supervision on frame importance with inter-frame temporal dependency}
\label{subsec:rel-sup-temporal-dependency}

\begin{figure}[ht]
  \centering
  \includegraphics[width=0.73\paperwidth]{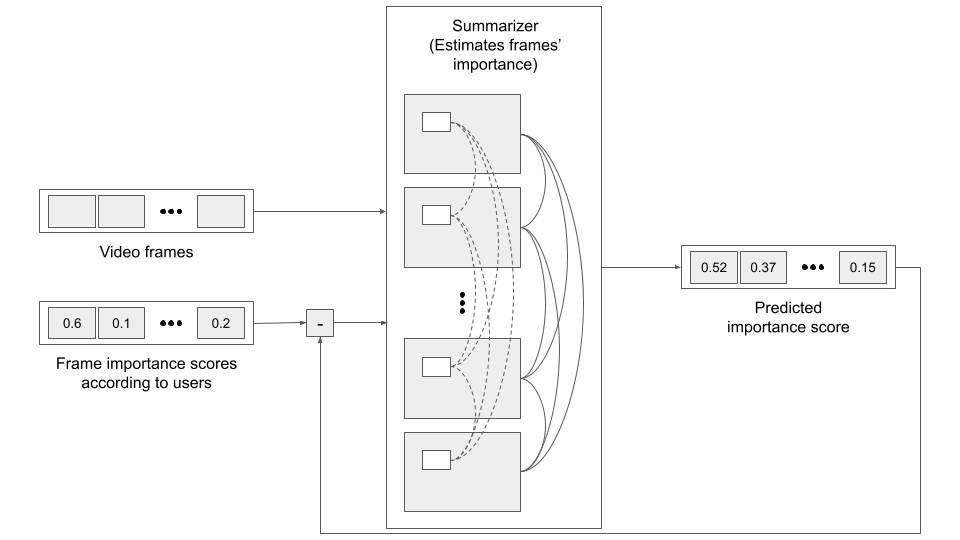}
  \caption{High-level representation of the analysis pipeline of supervised algorithms that perform summarization by learning the frames' importance after modeling their temporal or spatiotemporal dependency. For the latter class of methods (i.e., modeling the spatiotemporal dependency among frames), object bounding boxes and object relations in time shown with dashed rectangles and lines, are used to illustrate the extension that models both the temporal and spatial dependency among frames.}
  \label{figure:rel-sup-model}
\end{figure}

Early deep-learning-based approaches for video summarization treat the task as a structured prediction problem, aiming to estimate the importance of video frames by modeling their temporal dependencies. During the training phase, these approaches utilize ground-truth data indicating frame importance based on user preferences (see Figure \ref{figure:rel-sup-model}). The frames' temporal dependencies are modeled, and importance scores are predicted, which are then compared with the ground-truth data to guide the training of the summarization model. 

One of the initial approaches by Zhang \etal~\cite{zhang2016lstm} employed Long Short-Term Memory (LSTM) units to model variable-range temporal dependencies among video frames. Frame importance was estimated using a multi-layer perceptron (MLP), and diversity in the generated summary's visual content was enhanced using Determinantal Point Process (DPP). Zhao \etal~\cite{zhao2017hierarchical} introduced a two-layer LSTM architecture. The first layer extracted and encoded video structure data, while the second layer estimated fragment-level importance and selected key fragments. In their subsequent work, Zhao \etal~\cite{zhao2018hsa} incorporated a component that learned to identify shot-level temporal structure, enabling importance estimation at the shot level and producing a key-shot-based video summary. Extending their method, Zhao \etal~\cite{zhao2020tth} introduced a tensor-train embedding layer to address large feature-to-hidden mapping matrices. This layer, combined with a hierarchical structure of recurrent neural networks (RNNs), captured temporal dependencies within manually-defined video subshots and across different subshots, determining the probability of each subshot being selected for the video summary. Lebron Casas \etal~\cite{lebron2019attention} expanded on the work of Zhang \etal~\cite{zhang2016lstm} by incorporating an attention mechanism to model the temporal evolution of users' interest. This information was then used to estimate frame importance and select keyframes for constructing a video storyboard. Several other methods adopted sequence-to-sequence (seq2seq) architectures with attention mechanisms. Ji \etal~\cite{ji2019attentionEnDe} formulated video summarization as a seq2seq learning problem, using an LSTM-based encoder-decoder network with an intermediate attention layer. They extended their model in Ji \etal~\cite{ji2020deep} by integrating a semantic preserving embedding network and employing the Huber loss instead of Mean Square Error (MSE) loss for enhanced robustness. Fajtl \etal~\cite{fajtl2019summarizing} utilized a soft self-attention mechanism and a two-layer fully connected network for regression to estimate frame importance, avoiding computationally-demanding LSTMs. Liu \etal~\cite{liu2019learning} proposed a hierarchical approach combining a generator-discriminator architecture to estimate shot representativeness and select candidate keyframes, followed by a multi-head attention model for further importance assessment and final keyframe selection. Li \etal~\cite{li2021exploring} introduced a global diverse attention mechanism based on the self-attention mechanism of the Transformer Network, encoding temporal relations between frames and transforming diverse attention weights into importance scores. Another approach, presented by Rochan \etal~\cite{rochan2018sequence}, treated video summarization as a semantic segmentation task, treating the video as a 1D image and employing semantic segmentation models such as Fully Convolutional Networks (FCN) and DeepLab. They developed a network called Fully Convolutional Sequence Network that effectively modeled long-range dependencies among frames and learned frame importance by using convolutions with increasing effective context size. To address the limited capacity of LSTMs, some techniques incorporated additional memory. Feng \etal~\cite{feng2018memory} proposed a deep learning architecture that stored information about the entire video in an external memory, allowing each shot's importance to be predicted by learning an embedding space for matching shots with the memory information. Wang \etal~\cite{wang2019stacked} stacked multiple LSTM and memory layers hierarchically to capture long-term temporal context and estimate frame importance based on this information.

\subsection{Supervision on frame importance with video spatiotemporal structure}
\label{subsec:rel-sup-spatiotemporal}

In order to improve the estimation of video frame/fragment importance, certain techniques focus on capturing both the spatial and temporal structure of the video. These approaches not only take into account the input sequence of video frames and the available ground-truth data indicating frame importance but also model the spatiotemporal dependencies among frames. This additional analysis enhances the training process of the Summarizer, as shown by the dashed rectangles and lines in Figure \ref{figure:rel-sup-model}. Lal \etal~\cite{lal2019online} introduced an encoder-decoder architecture with convolutional LSTMs that effectively model the spatiotemporal relationships within the video. The algorithm not only estimates frame importance but also enhances visual diversity through next-frame prediction and shot detection mechanisms, leveraging the likelihood that the initial frames of a shot are often part of the summary. Yuan \etal~\cite{yuan2019spatiotemporal} employed a trainable 3D-CNN to extract deep and shallow features from the video content and fused them to create a new representation. This representation, combined with convolutional LSTMs, captures the spatial and temporal structure of the video. A novel loss function called Sobolev loss is then used to learn summarization by minimizing the distance between the series of frame-level importance scores and the ground-truth scores, effectively exploiting the temporal structure of the video. Chu \etal~\cite{chu2019spatiotemporal} leveraged CNNs to extract spatial and temporal information from raw frames and optical flow maps. Through a label distribution learning process, they learned to estimate frame importance based on human annotations.  Elfeki \etal~\cite{elfeki2019video} combined CNNs and Gated Recurrent Units (GRUs), a type of RNN, to form spatiotemporal feature vectors. These vectors were used to estimate the level of activity and importance for each frame. Huang \etal~\cite{huang2019novel} trained a neural network to extract spatiotemporal information from the video, specifically focusing on inter-frame motion. This information was used to create an inter-frame motion curve, which was then input into a transition effects detection method for shot segmentation. A self-attention model, guided by human-generated ground-truth data, was employed to estimate intra-shot importance and select keyframes/fragments for creating static/dynamic video summaries. By incorporating the spatial and temporal aspects of videos, these supervised approaches improve the accuracy of frame importance estimation and enable the generation of more informative video summaries.

\subsection{Supervision on summary authenticity with discriminative adversarial learning}
\label{subsec:rel-sup-discriminative}

\begin{figure}[ht]
  \caption{High-level representation of the analysis pipeline of supervised algorithms that learn summarization with the help of ground-truth data and adversarial learning.}
  \includegraphics[width=0.73\paperwidth]{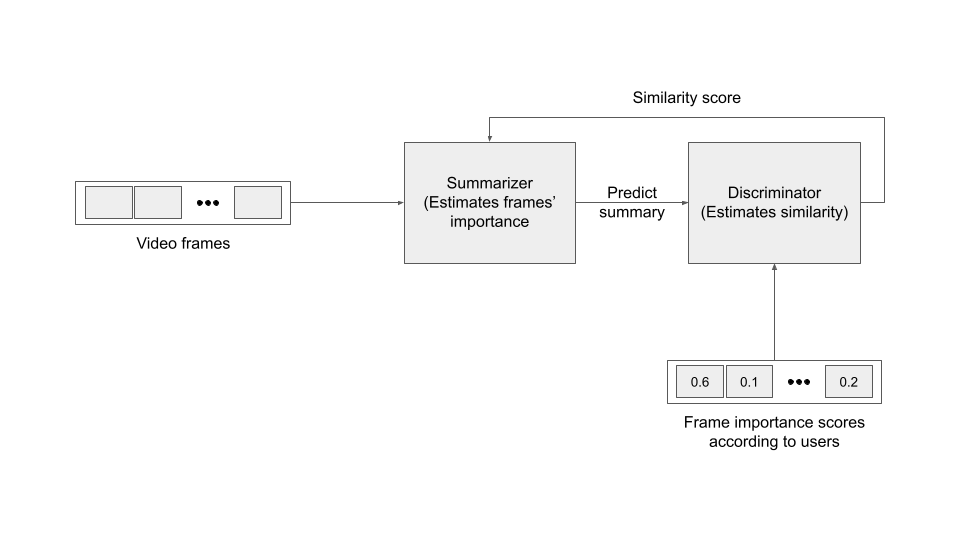}
  \label{figure:rel-sup-gan}
\end{figure}

Taking a distinct approach to bridge the gap between machine-generated and ground-truth summaries, certain methods leverage Generative Adversarial Networks (GANs). Illustrated in Figure \ref{figure:rel-sup-gan}, the Summarizer, acting as the GAN's Generator, takes the video frames as input and generates a summary by computing frame-level importance scores. These predicted scores, along with an optimal video summary based on user preferences, are fed to a trainable Discriminator, which evaluates their similarity and outputs a corresponding score. The training process encompasses an adversarial framework where the Summarizer aims to deceive the Discriminator by producing summaries that are indistinguishable from the user-generated ones, while the Discriminator learns to differentiate between them. When the Discriminator's confidence level becomes low, indicating an equal classification error for both machine- and user-generated summaries, the Summarizer successfully generates summaries that align closely with users' expectations. Zhang \etal~\cite{zhang2019dtr} introduced a method that employs LSTMs and Dilated Temporal Relational (DTR) units to capture temporal dependencies across different time windows. Their approach trains the Summarizer by attempting to mislead a trainable discriminator into distinguishing between machine-generated summaries, ground-truth summaries, and randomly created summaries. Fu \etal~\cite{fu2019attentive} proposed an adversarial learning technique for (semi-)supervised video summarization, where the Generator/Summarizer is an attention-based Pointer Network. It determines the start and end points of each video fragment used in the summary. The Discriminator, a 3D-CNN classifier, determines whether a fragment belongs to a ground-truth or machine-generated summary. Instead of using a conventional adversarial loss, their algorithm employs the output of the Discriminator as a reward for training the Generator/Summarizer through reinforcement learning. While the use of GANs in supervised video summarization is relatively limited, this machine learning framework has been extensively employed in unsupervised video summarization, which will be discussed in the subsequent section.
\section{Unsupervised approaches} 
\label{section:rel-unsupervised} 

Unsupervised methods eliminate the need for ground-truth data, which typically requires time-consuming and labor-intensive manual annotation. Instead, unsupervised approaches leverage large collections of original videos for training. Through learning mechanisms designed for unsupervised settings, these methods extract meaningful information from the video data to generate summaries.

\subsection{Fooling Discriminator to Discriminate Original Video from Summary-Based Reconstruction}
\label{section:rel-unsup-discriminative}

\begin{figure}[ht]
    \centering
    \includegraphics[width=0.73\paperwidth]{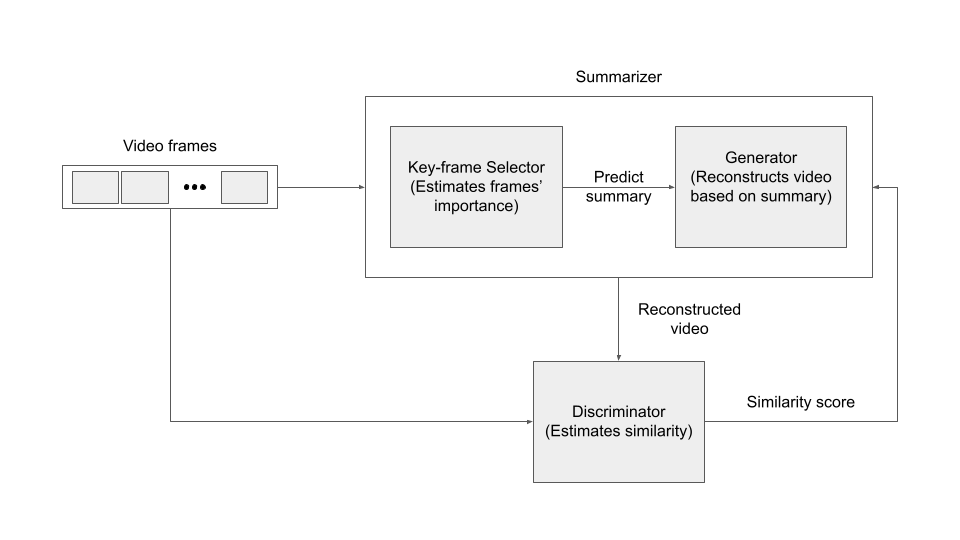}
    \caption{High-level representation of the analysis pipeline of unsupervised
    algorithms that learn summarization by increasing the similarity between the
    summary and the video.}
    \label{figure:rel-unsup-gan}
  \end{figure}

To address the absence of ground-truth data, unsupervised techniques leverage the principle that a representative summary should enable viewers to comprehend the original video content. To achieve this, Generative Adversarial Networks (GANs) are employed to learn the creation of video summaries that facilitate accurate reconstruction of the original video. The training process (see Figure \ref{figure:rel-unsup-gan}) involves a Summarizer, consisting of a Key-frame Selector and a Generator. The Key-frame Selector estimates frame importance and generates a summary, while the Generator reconstructs the video based on the generated summary. By inputting the video frames and predicting frame-level importance scores, the Summarizer reconstructs the original video. The reconstructed video, alongside the original one, is fed into a trainable Discriminator that evaluates their similarity. Similar to supervised GAN-based methods, the training of the entire summarization architecture follows an adversarial approach. In this case, the Summarizer's objective is to deceive the Discriminator by making it challenging to distinguish between the summary-based reconstructed video and the original video. Conversely, the Discriminator aims to improve its discrimination abilities. When the discrimination becomes indistinguishable (i.e., similar classification error for both videos), the Summarizer successfully constructs a highly representative video summary.  

Notably, Mahasseni \etal~\cite{mahasseni2017unsupervised} combined an LSTM-based key-frame selector, a Variational Auto-Encoder (VAE), and a trainable Discriminator, using adversarial learning to minimize the distance between the original video and the summary-based reconstructed version. Building upon this foundation, Apostolidis \etal~\cite{apostolidis2019stepwise} proposed a stepwise, label-based approach for training the adversarial part of the network, leading to enhanced summarization performance. Yuan \etal~\cite{yuan2019cycle} introduced an approach aiming to maximize the mutual information between the summary and the video, utilizing a pair of trainable discriminators and a cycle-consistent adversarial learning objective. Their frame selector, a bidirectional LSTM, constructs a video summary by modeling temporal dependencies among frames. The summary is then evaluated by two GANs—an encoder-decoder GAN for forward reconstruction and a backward GAN for reverse reconstruction. The consistency between these reconstructions quantifies information preservation, guiding the frame selector to identify the most informative frames for the video summary. In a subsequent work, Apostolidis \etal~cite{apostolidis2020ac} integrated an Actor-Critic model into a GAN, formulating the selection of important video fragments as a sequence generation task. The Actor and Critic engage in a game that incrementally selects video key fragments, with rewards from the Discriminator influencing their choices. This training workflow enables the Actor and Critic to learn a value function (Critic) and a policy for key-fragment selection (Actor). Other approaches extended the core VAE-GAN architecture by incorporating tailored attention mechanisms. For instance, Jung \etal~\cite{jung2019discriminative} proposed a VAE-GAN architecture extended with a chunk and stride network (CSNet) and a tailored difference attention mechanism, capturing frame dependencies at various temporal granularities during keyframe selection. In a subsequent work, Jung \etal~\cite{jung2020global} introduced a self-attention mechanism combined with relative position modeling, decomposing the frame sequence into non-overlapping groups to capture both local and global interdependencies. Apostolidis \etal~\cite{apostolidis2020unsupervised} presented a variation of their prior work \cite{apostolidis2019stepwise}, replacing the VAE with a deterministic Attention Auto-Encoder to improve attention-driven reconstruction and key-fragment selection. He \etal~\cite{he2019unsupervised} proposed a self-attention-based conditional GAN, utilizing a conditional feature selector and a multi-head self-attention mechanism to focus on important temporal regions and model long-range dependencies in the video sequence. Finally, Rochan \etal~\cite{rochan2019video} developed an approach for video summarization from unpaired data, employing an adversarial process with GANs and a Fully-Convolutional Sequence Network (FCSN) encoder-decoder. The model aimed to learn a mapping function from raw video to a human-like summary, aligning the summary distribution with human-created summaries while ensuring content diversity through an applied constraint on the learned mapping function.

These techniques aim to generate representative video summaries by fooling the Discriminator, making it difficult to distinguish between the summary-based reconstruction and the original video. By achieving similar classification errors for both, it indicates that the Summarizer has successfully created a summary that captures the overall video content.

\subsection{Focusing on Specific Desired Properties with Reinforcement Learning}
\label{subsec:rel-unsup-specific-properties}
\begin{figure}[ht]
    \centering
    \includegraphics[width=0.73\paperwidth]{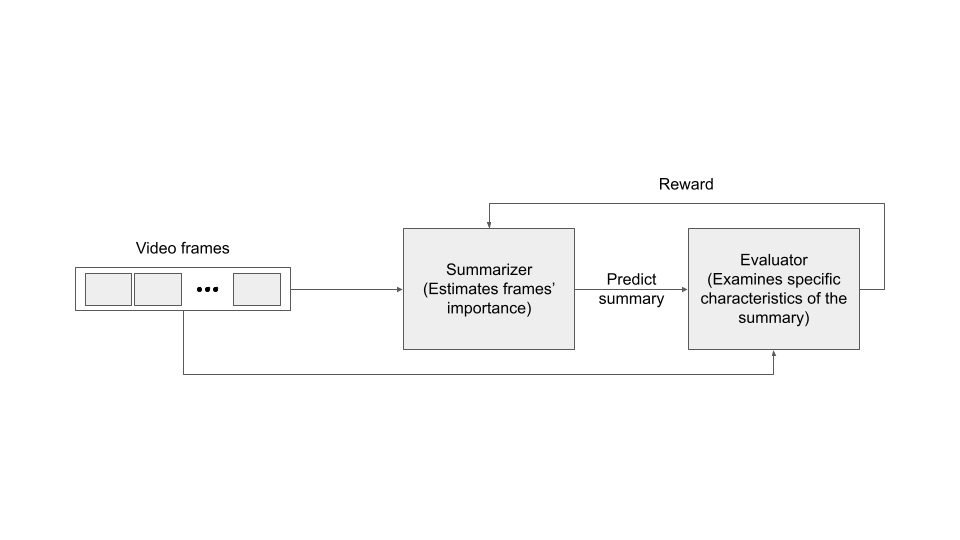}
    \caption{High-level representation of the analysis pipeline of supervised algorithms that learn summarization based on hand-crafted rewards and reinforcement learning.}
    \label{figure:rel-unsup-properties}
  \end{figure}

Addressing the challenges of unstable training and limited evaluation criteria in GAN-based methods, certain unsupervised approaches focus on specific properties of an optimal video summary. These approaches employ reinforcement learning principles in conjunction with hand-crafted reward functions that quantify desired characteristics in the generated summary. Illustrated in Figure \ref{figure:rel-unsup-properties}, the Summarizer takes the video frame sequence as input and generates a summary by predicting frame-level importance scores. The predicted summary is then evaluated by an Evaluator, which employs hand-crafted reward functions to measure the presence of specific desired characteristics. The computed scores are combined to form an overall reward value, guiding the training of the Summarizer.

The initial work in this direction, proposed by Zhou \etal~\cite{zhou2018deep}, formulates video summarization as a sequential decision-making process. They train a Summarizer to produce diverse and representative video summaries using a diversity-representativeness reward. The diversity reward quantifies the dissimilarity among selected keyframes, while the representativeness reward measures the visual resemblance of the selected keyframes to the remaining frames of the video. Expanding on this method, Yaliniz \etal~\cite{yaliniz2021using} present another reinforcement-learning-based approach that incorporates the uniformity of the generated summary. They employ Independently Recurrent Neural Networks (IndRNNs) activated by a Leaky ReLU function to model temporal dependencies among frames. This addresses issues related to decaying, vanishing, and exploding gradients in LSTM models and facilitates better learning of long-term dependencies. In addition to rewards associated with representativeness and diversity, Yaliniz et al. introduce a uniformity reward to enhance the coherence of the summary and prevent redundant jumps between selected video fragments. Gonuguntla \etal~\cite{gonuguntla2019enhanced} propose a method utilizing Temporal Segment Networks, originally designed for action recognition in videos, to extract spatial and temporal information from video frames. They train the Summarizer using a reward function that evaluates the preservation of the video's main spatiotemporal patterns in the generated summary. Lastly, Zhao \etal~\cite{zhao2019property}present a mechanism that combines video summarization and reconstruction. Video reconstruction aims to estimate how well the summary allows viewers to infer the original video, similar to some GAN-based methods. Video summarization is learned based on feedback from the reconstructor and the output of trained models that assess the representativeness and diversity of the visual content in the generated summary.

In conclusion, reinforcement learning has emerged as a promising alternative to GAN-based methods in the field of video summarization. By employing principles of sequential decision-making and custom reward functions, these techniques strive to produce video summaries that are diverse, representative, and coherent. This approach overcomes the challenges of training stability and limited evaluation criteria often associated with GAN-based approaches. By incorporating rewards for diversity, representativeness, uniformity, and preservation of spatiotemporal patterns, the Summarizer can effectively learn optimal summary generation. Although still a developing area, reinforcement learning in video summarization shows great potential for advancing the development of automated summarization algorithms that effectively capture key information while preserving the visual integrity of the original videos. Ongoing research and experimentation will undoubtedly refine and enhance the capabilities of reinforcement learning-based approaches in this domain.

\subsection{Building Object-Oriented Summaries through Key Object Motion}
\label{subsec:rel-unsup-object-oriented}

Zhang \etal~\cite{zhang2020unsupervised} devised a novel method that prioritizes the retention of fine-grained semantic and motion details within the video summary. Their approach involves an initial preprocessing step aimed at identifying significant objects and their key motions. Leveraging this information, the method represents the entire video by creating segmented object motion clips. Subsequently, these clips are fed into the Summarizer, which employs an online motion auto-encoder model known as Stacked Sparse LSTM Auto-Encoder. This model continually updates a customized recurrent auto-encoder network to encode and memorize previous states of object motions. The network's primary task is to reconstruct object-level motion clips, with the reconstruction loss computed between the input and output frames serving as a guide for training the Summarizer. Through this training process, the Summarizer becomes proficient in generating summaries that highlight the representative objects in the video and the key motions associated with each object.

\section{Weakly supervised approaches}
\label{section:rel-weakly}

Similar to unsupervised approaches, weakly-supervised methods aim to reduce the reliance on extensive sets of hand-labeled data. Instead of completely forgoing ground-truth data, these methods leverage less costly weak labels, such as video-level metadata or sparse annotations for a subset of frames. The underlying hypothesis is that while these labels are imperfect compared to comprehensive human annotations, they can still facilitate the training of effective summarization models.

This class of methods does not follow a typical analysis pipeline, as they diverge in their approach to learning the summarization task. One of the early approaches in this domain was introduced by Panda \etal~\cite{panda2017weakly}. Their method utilizes video-level metadata to categorize videos and extracts 3D-CNN features to learn a parametric model for categorizing unseen videos. The model is then employed to select video segments that maximize the relevance between the summary and the video category. Panda \etal~ addressed challenges related to limited dataset size by exploring cross-dataset training, incorporating web-crawled videos, and employing data augmentation techniques to increase the training data.

Cai \etal~\cite{cai2018weakly} extended the idea of learning summarization from semantically-similar videos in a weakly-supervised setting. They proposed an architecture combining a Variational AutoEncoder (VAE) to learn latent semantics from web videos and a sequence encoder-decoder with attention mechanism for summarization. The VAE's decoding part reconstructs input videos using samples from the learned latent semantics, while the attention mechanism of the encoder-decoder network identifies the most important video fragments. The attention vectors are obtained by integrating the learned latent semantics from collected web videos. The architecture is trained using a weakly-supervised semantic matching loss to learn topic-associated summaries.

Ho \etal~\cite{Ho2018FVPSum} presented a deep learning framework for summarizing first-person videos but are included here as their method was also evaluated on a dataset used for assessing generic video summarization methods. Recognizing the difficulty of collecting a large amount of fully-annotated first-person video data, they utilized transfer learning principles. Annotated third-person videos, which are more readily available, were used to train the model on how to summarize first-person videos. The algorithm employed cross-domain feature embedding and transfer learning for domain adaptation between third- and first-person videos in a semi-supervised manner.

Chen \etal~\cite{chen2019weakly} employed the principles of reinforcement learning to construct and train a summarization method using a limited set of human annotations and handcrafted rewards. The rewards encompassed similarity between machine- and human-selected fragments and specific characteristics of the generated summary, such as representativeness. Their method employed a hierarchical key-fragment selection process divided into sub-tasks. Each task was learned through sparse reinforcement learning, utilizing annotations only for a subset of frames rather than exhaustive annotations for the entire set. The final summary was formed based on rewards related to diversity and representativeness.

These weakly-supervised approaches demonstrate innovative strategies to overcome the limitations of fully-supervised learning while leveraging available weak labels and tailored reward functions to train effective video summarization models.

\chapter{Context-Aware Video Summarization}
\label{chapter:method}

\begin{ChapAbstract}
    This chapter introduces our approach for context-aware video summarization and human-centric evalution. In Section \ref{section:method-model}, we present the general architecture of our proposed method, which incorporates contextual information to generate more informative and relevant video summaries. Afterward, the Section \ref{section:method-details} is dedicated to describe the technical details related to the implementation of our method. Additionally, in Section \ref{section:method-evaluation}, we describe our human-centric evaluation pipeline that allows us to assess the effectiveness of our approach in capturing the key content and understanding of the original videos.
\end{ChapAbstract}

\section{Context-Aware Video Partitioning and Summarization}
\label{section:method-model}
    Our proposed approach involves a sequential process consisting of four stages to select an ordered subset $\mathbf{S}$ of $L$ frames from an original video $\mathbf{I} = \{I_t\}^{T}_{t = 1}$, where $T$ represents the total number of frames in the video. The summarized subset $\mathbf{S} = \{I_{t_i}\}^{L}_{i = 1}$ is obtained by selecting frames indexed by  $t_i$ from the original video, satisfying the requirements that $t_i \in [1, T]$ and $t_i < t_{i + 1}$ for all valid $i \in [1, L)$.

\begin{figure}[ht]
        \centering
        \includegraphics[width=0.73\paperwidth]{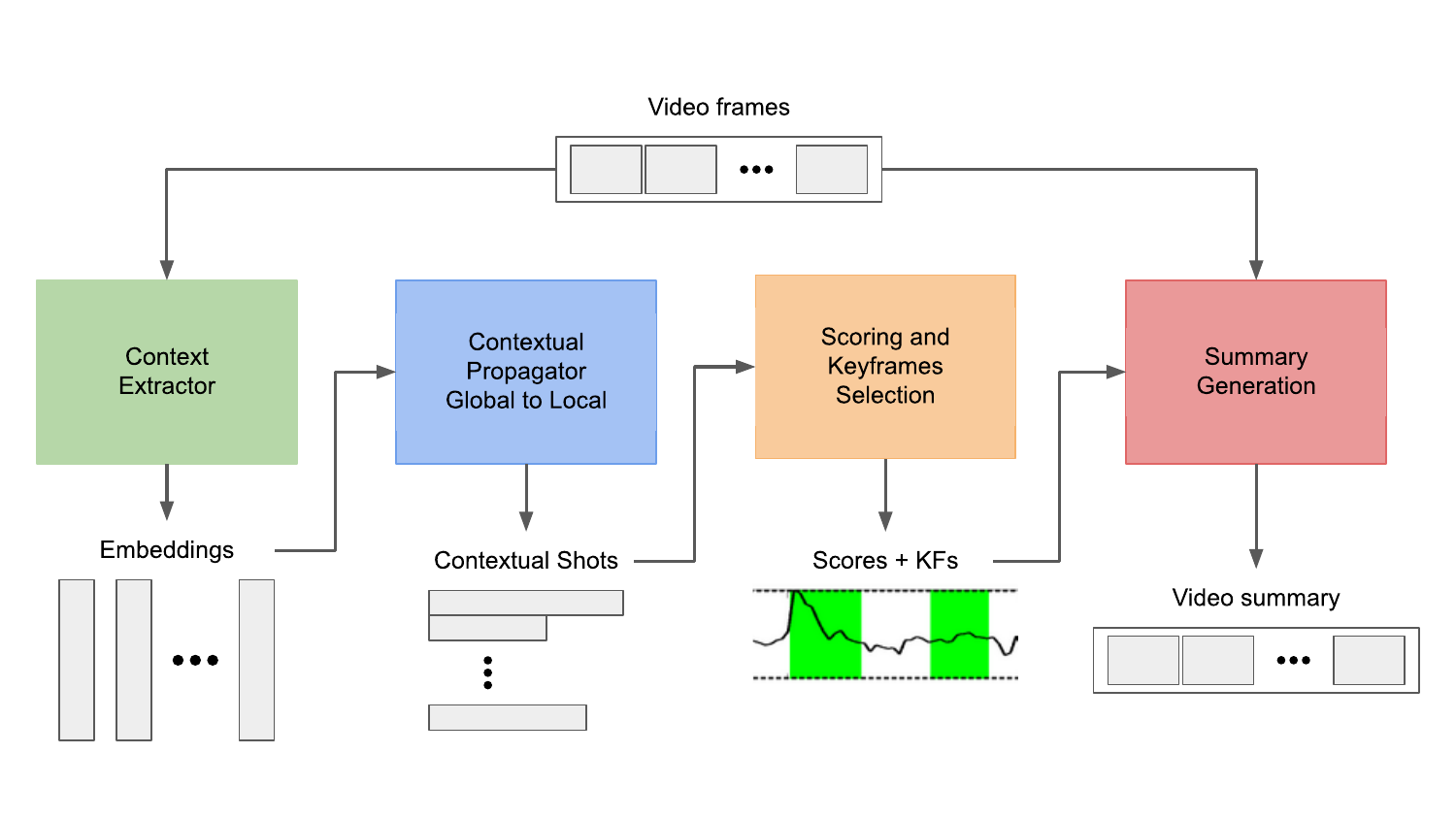}
        \caption{Pipeline of the proposed approach showcasing four modules and information flow across main stages. Contextual embeddings extracted from original video frames in the first stage and output summary generated in the final stage. Other information utilized in subsequent stages.}
        \label{fig:method-model-pipeline}
    \end{figure}

    In Figure \ref{fig:method-model-pipeline}, we illustrate the four main stages of our method as distinct modules, each associated with a particular stage in our pipeline. Each stage comprises several steps that are tailored to the specific role and algorithm implemented in that stage. Throughout the remaining text of this section, we provide a detailed explanation of each stage, discussing its purpose, the steps involved, and the algorithms utilized.
    
    To summarize, our pipeline consists of four stages that contribute to the generation of the video summary. Here is an overview of each stage and its corresponding subsection:

    \begin{enumerate}
        \item Context Extractor (Subsection \ref{subsec:method-model-generating})
            \begin{itemize}
                \item This stage extracts contextual information from the original input video.
                \item The contextual information is represented by the contextual embedding $\mathbf{E}$.
            \end{itemize}
        \item Global to Local Propagation (Subsection \ref{subsec:method-model-global-local})
            \begin{itemize}
                \item In this stage, the contextual information is propagated from the global level to the segment-like level.
                \item A global clustering step is performed to group frames with similar contextual information.
                \item Further partitioning is applied to create semantic partitions $\mathbf{P}$, which contain frames with related content.
            \end{itemize}
        \item Keyframes and Importance Scores (Subsection \ref{subsec:method-model-keyframes-importance})
            \begin{itemize}
                \item The created semantic partitions serve as the basis for scoring and selecting relevant frames.
                \item Frame-level information is used to assign importance scores to frames.
                \item Keyframes, representing the most important frames, are selected from each semantic partition.
            \end{itemize}
        \item Summary Generation (Subsection \ref{subsec:method-model-summary})
            \begin{itemize}
                \item In the final stage, the selected keyframes and their surrounding frames are used to construct the output video summary $\mathbf{S}$.
                \item The summary is generated based on the frame-level information and importance scores.
            \end{itemize}
    \end{enumerate}

    \subsection{Generating Contextual Embeddings}
    \label{subsec:method-model-generating}
        In this stage, we aim to extract the context of an input video $\mathbf{I}$ from its frames $I_t$. This process involves two main steps: sampling the video \ref{subsubsec:method-model-generating-sampling} and constructing embeddings \ref{subsubsec:method-model-generating-embeddings} for each sampled frame. We will discuss each step in detail.
    
        \subsubsection{Sampling}
        \label{subsubsec:method-model-generating-sampling}
            To reduce computational complexity of the embedding extraction step, we employ a sampling technique to extract frames from the original video $\mathbf{I}$ into a sequence of samples denoted as $\mathbf{\hat{I}}$ consists of $\{I_{t_i}\}_{i = 1}^{\hat{T}}$ with $\hat{T}$ is length of the sampled sequence satisfying $\hat{T} < T$. Hence, the complexity of the embedding extraction is squeezed from $\mathcal{O}(cT)$ to $\mathcal{O}(c\hat{T})$ with $\mathcal{O}(c)$ being the complexity of extracting the embedding of an individual frame.
            
            The frames are sampled so that the frame rate of the reduced sequence $\mathbf{\hat{I}}$ (which is then denoted as $r_S$) matches a pre-specified frame rate, noted as $R$. This is done to ensure that samples are stable and representative enough of the original content while being short enough to improve the complexity of the step in a significant way. The reason that this method can ensure the representativeness of sampled sequence stems from an observation that the information density through the temporal dimension of input videos is relatively constant. For example, the number of actions per second would usually be distributed in a normal way across multiple videos, guaranteeing that sampling the videos with frame rate slightly higher than the average action density would capture most of the original actions.
            
            Furthermore, the use of a pre-specified parameter $R$ also serves as a normalization for different inputs with several frame rates so that the sampled sequence $\mathbf{\hat{I}}$ would eventually have a nearly fixed frame rate (close to $R$) even though the original frame-rate $r_I$ is arbitrarily distributed. Therefore, this sampling method would pre-process the original video with high stability while retaining most of the necessary information, paving the way for easier processing in subsequent steps.
            
\begin{figure}[ht]
                \centering
                \includegraphics[width=0.73\paperwidth]{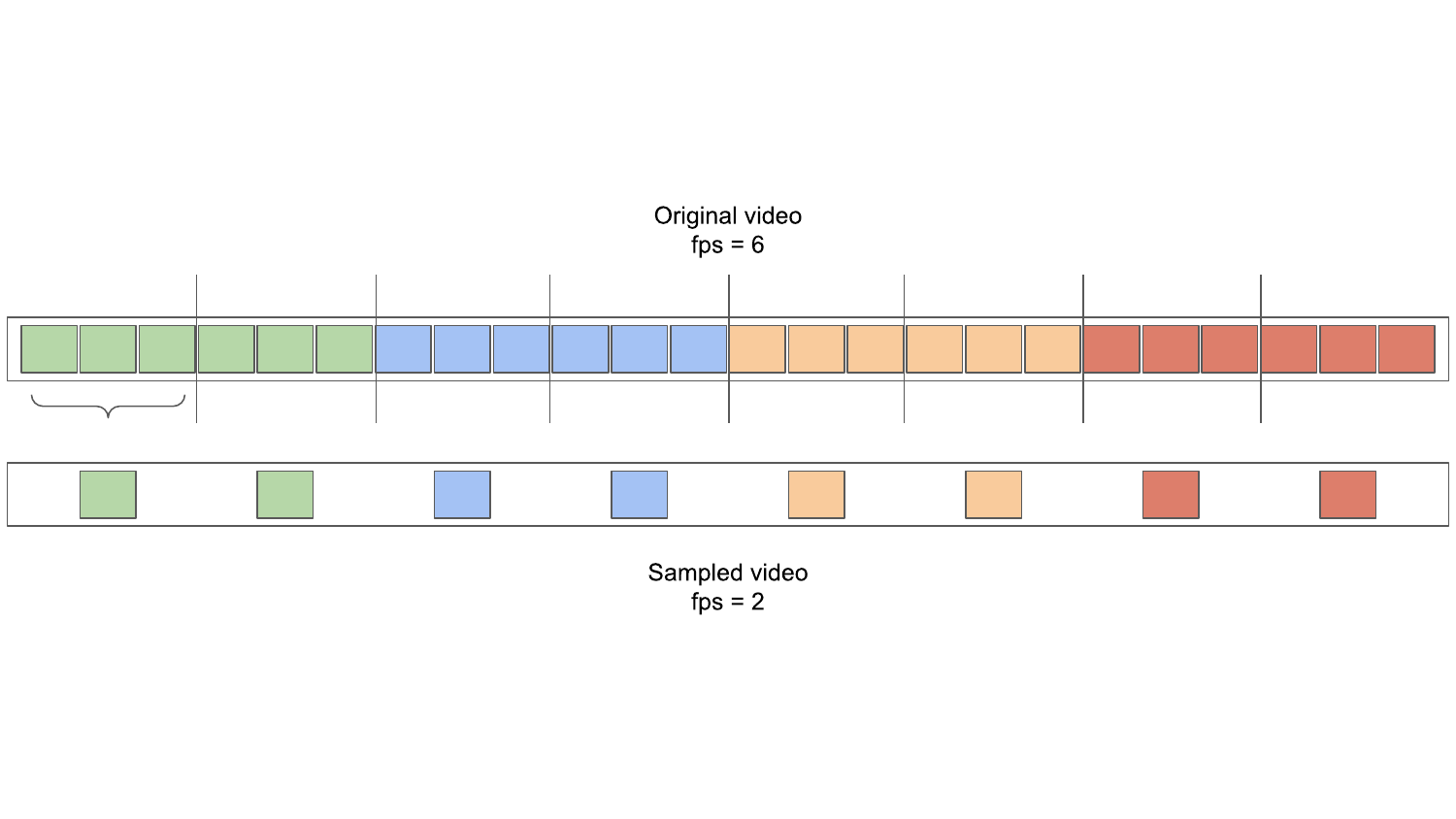}
                \caption{Original video ($r_I = 6$ fps) downsampled with a sampling rate of $r_S = 2$ fps. Frames inside the same clocking period of a second are denoted with similar colors and frames from different periods are illustrated with distinct colors. Frames inside a space created with black dashes belong to the same snippet.}
                \label{fig:method-model-sampling}
            \end{figure}

            This process involves dividing the original frames within a one-second period into several equal-length snippets. Afterward each of the periods is further separated into equal-length snippets whose lengths are determined based on the original frames per second (fps) and the desired sampling rate. Subsequently, the middle frame of each snippet is selected as the final sample. This process is depicted in Figure \ref{fig:method-model-sampling}.
    
            Realizing the above sampling process requires only simple operations on the indexes of the original videos: we need to recover the indexes of target samples as a vector $\mathbf{t} = \{t_i\}_{i = 1}^{\hat{T}}$ from the original indexing range $[1, T]$.

            We first identify the length of the sampling snippets from which samples are drawn, which is approximate $\hat{\ell} = \frac{r_I}{r_S} + \mathcal{O}(1)$ with the adjustment $\mathcal{O}(1)$ due to non-divisible relation between the two frame-rates. This variable gives a direct calculation for the first position of the sample as it lies in the middle of the first snippet: $t_1 = \frac{\hat{\ell}}{2} + \mathcal{O}(1)$. On the other hand, the spacing between samples equals the length of the snippets as all of the sampled indexes are the midpoints of these snippets. Thus, the indexes of our samples are of the following form:

            \begin{equation}
                t_i = t_1 + \hat{\ell} (i - 1), \forall i \in [1, \hat{T}]
            \end{equation}

        With the number of samples $\hat{T}$ computed by $\lceil \frac{T - t_1}{\hat{\ell}} \rceil$. And from the vector $\mathbf{t} = \{t_i\}_{i = 1}^{\hat{T}}$, one can map the original input into the sampled frames with $\mathbf{\hat{I}} = \{\hat{I}_i\}_{i = 1}^{\hat{T}}$ where $\hat{I}_i = I_{t_i}$.

        \subsubsection{Embeddings}
        \label{subsubsec:method-model-generating-embeddings}
\begin{figure}[ht]
                \centering
                \includegraphics[width=0.73\paperwidth]{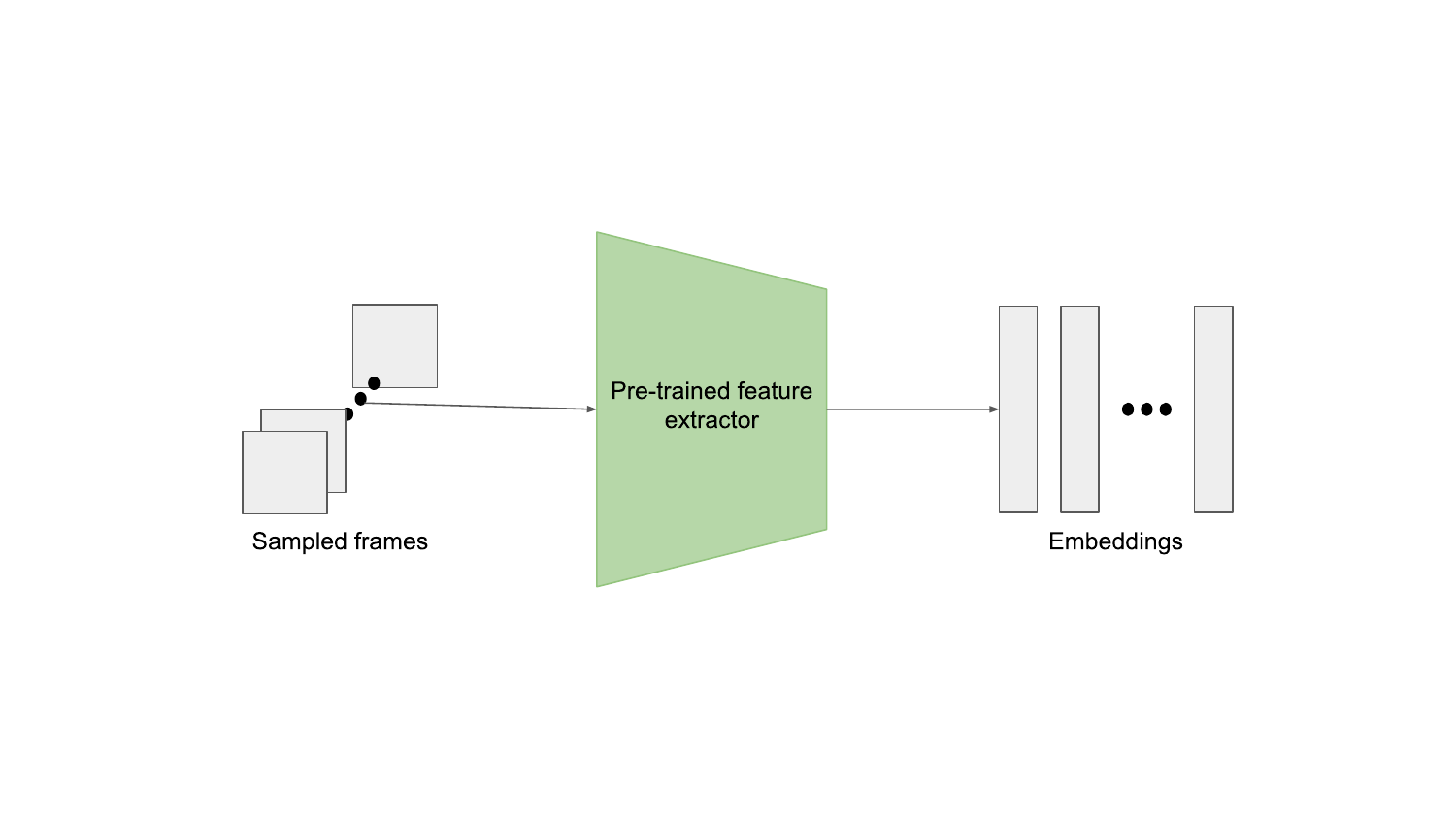}
                \caption{Frame-by-frame extraction of visual embeddings from the video.}
                \label{fig:method-model-embeddings}
            \end{figure}

            For each sampled frame $\hat{I}_i$ obtained from the previous step, we utilize a pre-trained model to extract its visual embedding $\mathbf{e}_i$ as illustrated in Figure \ref{fig:method-model-embeddings}). The pre-trained model is herein denoted as a function $g: \mathbb{R}^{W \times H \times C} \longrightarrow \mathbb{R}^{D}$ that converts the sample $\hat{I}_i$ as a multi-channel tensor into an embedding vector of size $D$ whose value depends on the pre-trained model used.
            
            The embedding $\mathbf{e}_i$ represents the visual information captured by the individual frame $\hat{I}_i$. All the embeddings $\mathbf{e}_i$ are then used to form the contextual embedding of the sampled video $\mathbf{E} = \{\mathbf{e}_i\}_{i = 1}^{\hat{T}}$.
            
\begin{figure}[ht]
                \centering
                \includegraphics[width=0.73\paperwidth]{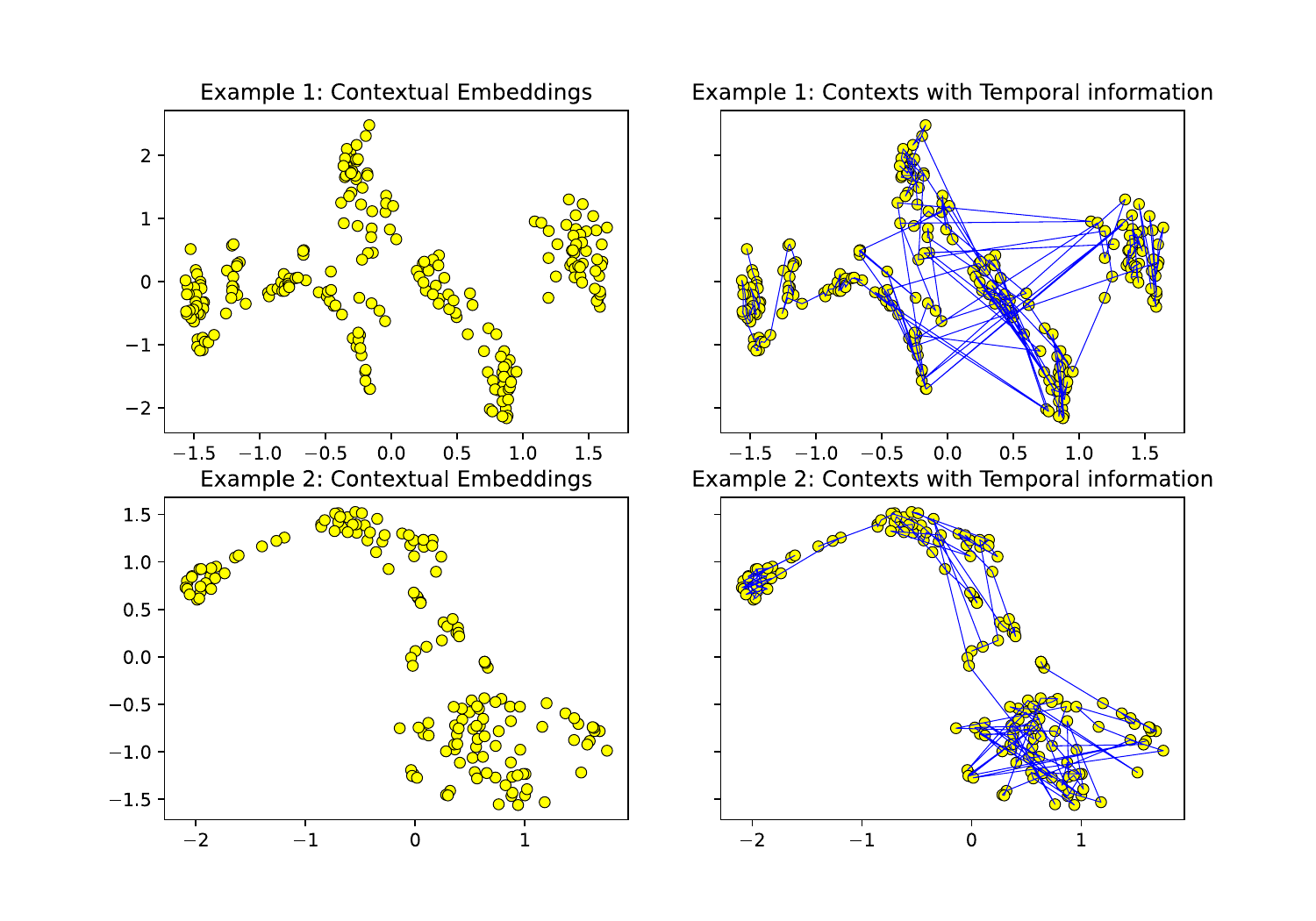}
                \caption{Visual illustration of the contextual information which is the result of the embedding step explained.}
                \label{fig:method-model-context}
            \end{figure}

            This contextual information, comprising the set of embeddings, serves as an important intermediate result for subsequent stages of the process. Two examples of the contextual embeddings are given in Figure \ref{fig:method-model-context} to visually illustrate the obtained contexts in the form of raw embeddings as well as temporally connected depictions. In that Figure, the first row depicts the first example while the second row demonstrates the second one, where left-hand side of each row shows the context's raw embeddings projected into 2-dimensional space while the right-side part adds temporal connections to that projected embeddings. The temporal connections are shown in blue segments while the sampled embeddings are given in yellow circles. Each of these scattered point represents an embedding $\mathbf{e}_i$ of a sampled frame $\hat{I}_i$ in the associate video, projected into $2$-dimensional space. Each temporal connection connecting two sampled embeddings $\mathbf{e}_i$ and $\mathbf{e}_j$ means that their associated frames $\hat{I}_i$ and $\hat{I}_j$ are adjacent to each other in the temporal dimension, in other words either $i = j + 1$ or $j = i + 1$.
            
            Note that these embeddings serve as a basis for further computations in the summarization process and improvement of their generation is not of this work's interest. Therefore, the work only uses an established methodology, a pre-trained model for example, for computing the visual features while leaving details of such methodology out of scope.
        
    \subsection{From Global Context to Local Semantics}
    \label{subsec:method-model-global-local}
        In this stage, the global information contained inside the contextual embedding $\mathbf{E}$ which is obtained from the previous stage is distilled to finer and more local levels, more specifically the partition-level and sample-level. Our proposed method comprises two steps that distill the global information into the respective levels. In the first one, we use traditional clustering to propagate the contextual information into clusters that represent partition-level information. Afterward the second step is applied so that the partition-level information is further distilled into sample-level by our proposed algorithm.

        \subsubsection{Contextual Clustering}
\begin{figure}[ht]
                \centering
                \includegraphics[width=0.73\paperwidth]{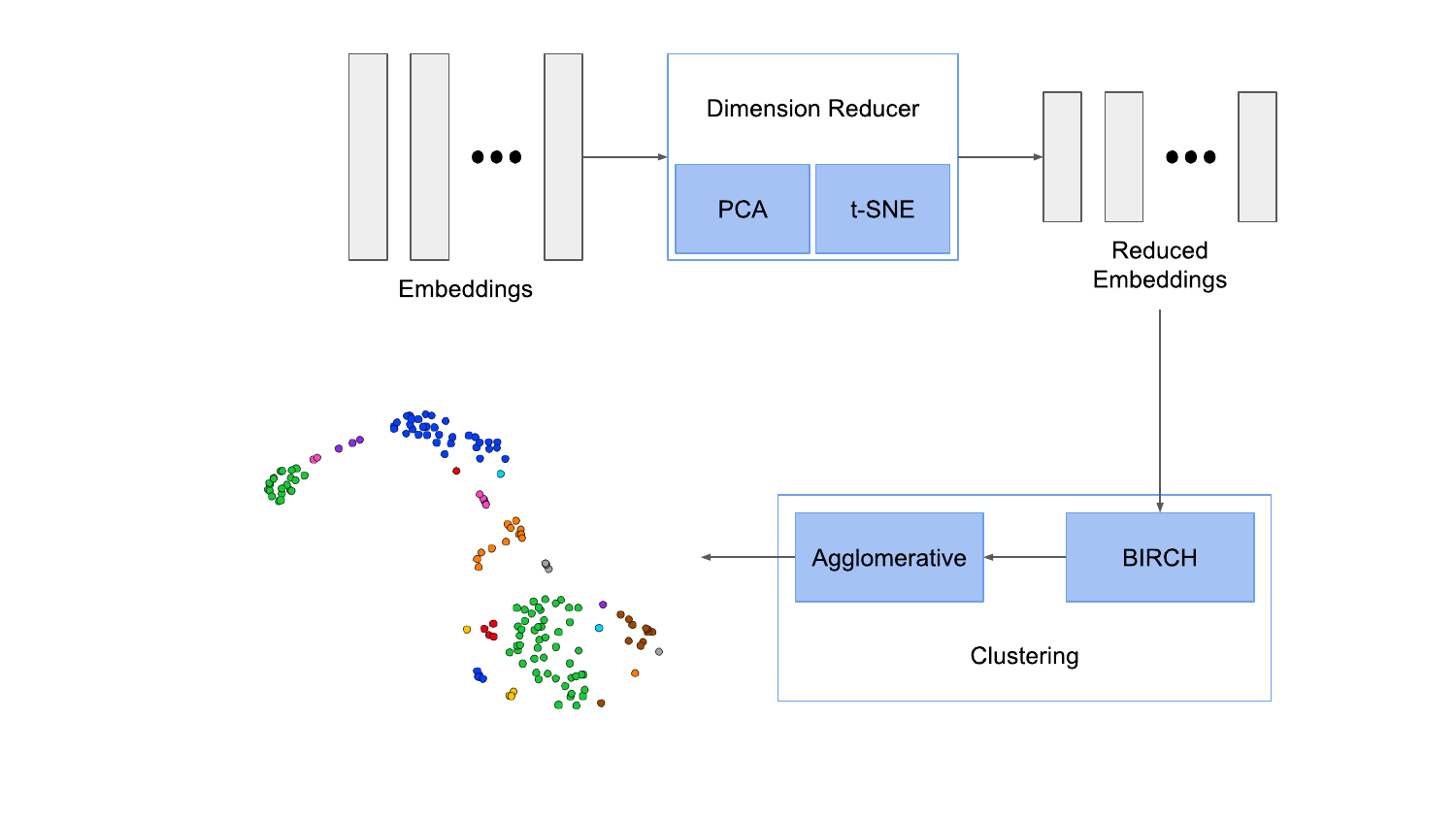}
                \caption{Overall pipeline for the Contextual Clustering step.}
                \label{fig:method-model-clustering}
            \end{figure}

            Clustering the aforementioned contextual embeddings $\mathbf{E}$ allows us to capture both the global relationships between different visual elements in the video which are inter-cluster relations (\textit{.i.e}, samples belonging to different clusters come from distinct parts of the video), and the local relationships representing intra-cluster relations (\textit{.i.e}, samples inside the same cluster share similar semantic meanings). To achieve this, we first reduce the dimension of the contextual embedding $\mathbf{E}$ to a reduced embedding $\mathbf{\hat{E}}$. After which a coarse-to-fine clustering approach is applied on this reduced embedding to divide the sampled frames into $K$ clusters, creating a label vector $\mathbf{c} \in \mathbf{N}^{\hat{T}}$. More details about this step can be found at the following texts as well as in the demonstration given by Figure \ref{fig:method-model-clustering}.

            \paragraph{Dimension reduction}
                Starting with the contextual embedding $\mathbf{E} \in \mathbb{R}^{\hat{T} \times D}$, a reduced embedding $\mathbf{\hat{E}} \in \mathbb{R}^{\hat{T} \times \hat{D}}$ is computed through several methods satisfying $\hat{D} \ll D$.

                In our method, two classic algorithms on dimensionality reduction are employed to perform this computation, with the first being Principal Components Analysis (PCA) \cite{pca} and the second one is t-Distributed Stochastic Neighbor Embedding (t-SNE) \cite{tsne}. The PCA has higher efficiency than t-SNE which makes it suitable for reducing large dimensions in shorter time. On the other hand, t-SNE has greater performance compared to that of PCA, meaning that it preserves the metric relationships between data points after its reduction, though with much longer runtime.
                
                Therefore, in order to utilize the best out of these algorithms we first use a PCA first to reduce the dimension of the context to a smaller size enough for t-SNE to further. Mathematically, the contextual information $\mathbf{E}$ is reduced by PCA first into a semi-reduced embedding denoted as $\mathbf{E'} = \mathbb{R}^{\hat{T} \times D'}$ where $D'$ is the dimension of the semi-reduced embedding with each vector reduced to $\mathbb{R}^{D'}$. Then the t-SNE is applied to perform final reduction that converts $\mathbf{E'}$ into $\mathbf{\hat{E}}$ of specified target dimension $\hat{D}$.

            \paragraph{Coarse clustering}
                With the reduced context $\mathbf{\hat{E}}$, a traditional clustering method called BIRCH (Balanced Iterative Reducing and Clustering using Hierarchies) algorithm \cite{birch} is applied to compute the coarse clusters of sampled frames. This method is a parameter-free clustering method, it calculates the coarse clusters based on the spatial characteristics of the reduced embedding $\mathbf{\hat{E}}$, hence each coarse cluster represents a set of sampled frames that share similar semantic properties.
                
                The sample-level notation for coarse clusters is a vector $\mathbf{\hat{c}} = \{{\hat{c}}_i\}_{i = 1}^{\hat{T}}$ with ${\hat{c}}_i$ is the label of the coarse cluster that $i$-th sample belongs to, satisfying that ${\hat{c}}_i \in [1, K']$ where $K'$ is the number of coarse clusters outputted by BIRCH algorithm. As BIRCH is a parameter-free algorithm, this number $K'$ is also automatically detected by the algorithm.

\begin{figure}[ht]
                    \centering
                    \includegraphics[width=0.73\paperwidth]{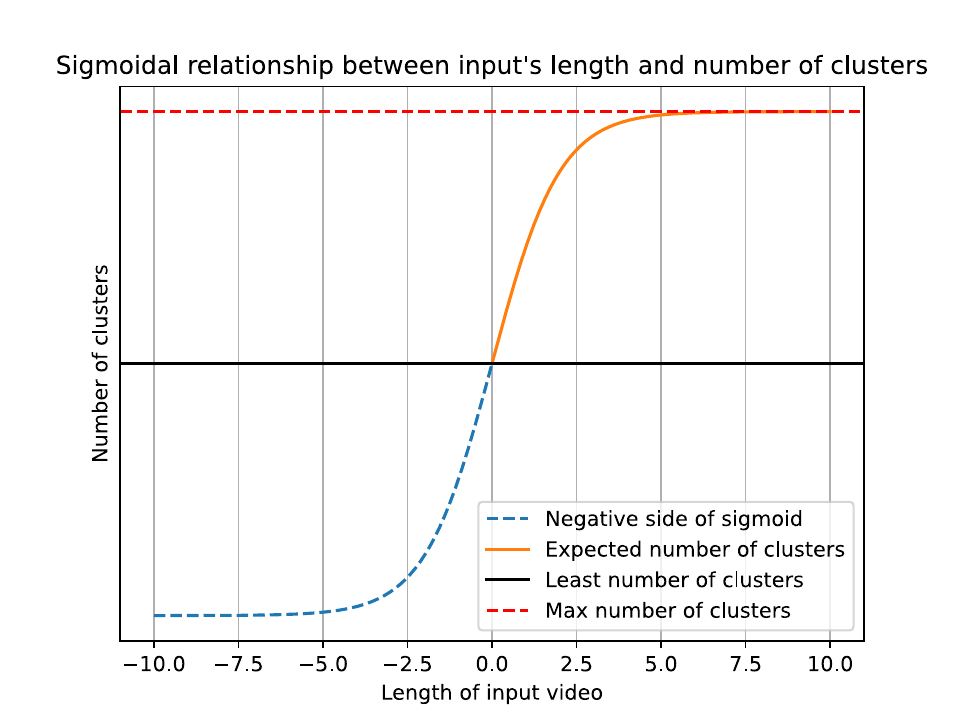}
                    \caption{The function illustrating the relationship between length of input video and number of final clusters.}
                    \label{fig:method-model-clusternum}
                \end{figure}

            \paragraph{Fine clustering}
                After the calculation of coarse clusters, a more sophisticated hierarchical clustering algorithm is employed to combine them into finer clusters. The number of these eventual clusters is pre-determined based on prior computations involving a positive side of a sigmoidal function and a maximum threshold on the number of possible clusters. Visual demonstration of this sigmoidal relationship is illustrated by Figure \ref{fig:method-model-clusternum} while the specific function is given in Equation \ref{eq:method-model-num-cluster} with $Z$ is a modulation parameter, $\hat{K}$ is the maximum number of clusters allowed, and $L'$ is the target number of frames in the final summary. The definition of $L'$ is provided in Subsection \ref{subsec:method-model-summary} under Equation \ref{eq:method-summary-length}.

                \begin{equation}
                \label{eq:method-model-num-cluster}
                    K = \frac{2 \cdot \hat{K}}{1 + \exp^{Z \cdot L'}} - \hat{K}
                \end{equation}
                
                To ensure that the local relationships between samples in the coarse clusters are preserved in the final clustering results, the fine cluster is formed as the union of at least one coarse cluster. This means that frames belonging to the same coarse cluster will be grouped into the same cluster in the final clustering $\mathbf{c}$. In other words, if ${\hat{c}}_i = {\hat{c}}_j$ then ${c}_i = {c}_j$.

                The rule for merging different clusters into one during the processing of this fine clustering is adopted based on the algorithm of Agglomerative Clustering that was originally proposed in \cite{agglo} with various types of rules available while supporting several forms of distance. In general, clusters are progressively and incrementally merged based on the affinity between them where a pair of remaining clusters with highest affinity is merged first. For each of the pair, this affinity value is calculated based on the spatial distances between elements from both clusters. Some rules about computing the affinity is outlined below without further clarification as it is out of this work's scope:

                \begin{itemize}
                    \item Rule \textit{single}: The affinity is the minimum distance between any pair of elements from two clusters with the affinity between two clusters $i$ and $j$ computed by Equation \ref{eq:agglo-single}.

                    \begin{equation}
                    \label{eq:agglo-single}
                        a(i, j) = \min_{{c}_k = i, {c}_l = j} d\left(\mathbf{\hat{e}}_k, \mathbf{\hat{e}}_l\right)
                    \end{equation}

                    \item Rule \textit{complete}: The affinity is the maximum distance between any pair of elements from two clusters with the affinity between two clusters $i$ and $j$ computed by Equation \ref{eq:agglo-complete}.

                    \begin{equation}
                    \label{eq:agglo-complete}
                        a(i, j) = \max_{{c}_k = i, {c}_l = j} d\left(\mathbf{\hat{e}}_k, \mathbf{\hat{e}}_l\right)
                    \end{equation}

                    \item Rule \textit{average}: The affinity is the average distance between any pair of elements from two clusters with the affinity between two clusters $i$ and $j$ computed by Equation \ref{eq:agglo-average}.

                    \begin{equation}
                    \label{eq:agglo-average}
                        a(i, j) = \frac{1}{|{c}_k = i| \times |{c}_l = j|} \sum_{{c}_k = i, {c}_l = j} d\left(\mathbf{\hat{e}}_k, \mathbf{\hat{e}}_l\right)
                    \end{equation}
                \end{itemize}

                In the above examples, $d\left(\mathbf{\hat{e}}_k, \mathbf{\hat{e}}_l\right)$ denotes the distance between the reduced embeddings of $k$-th and $l$-th samples, respectively. The particular type of distance used in the method is specified by another section of this chapter (Section \ref{section:method-details}).

\begin{figure}[ht]
                    \centering
                    \includegraphics[width=0.73\paperwidth]{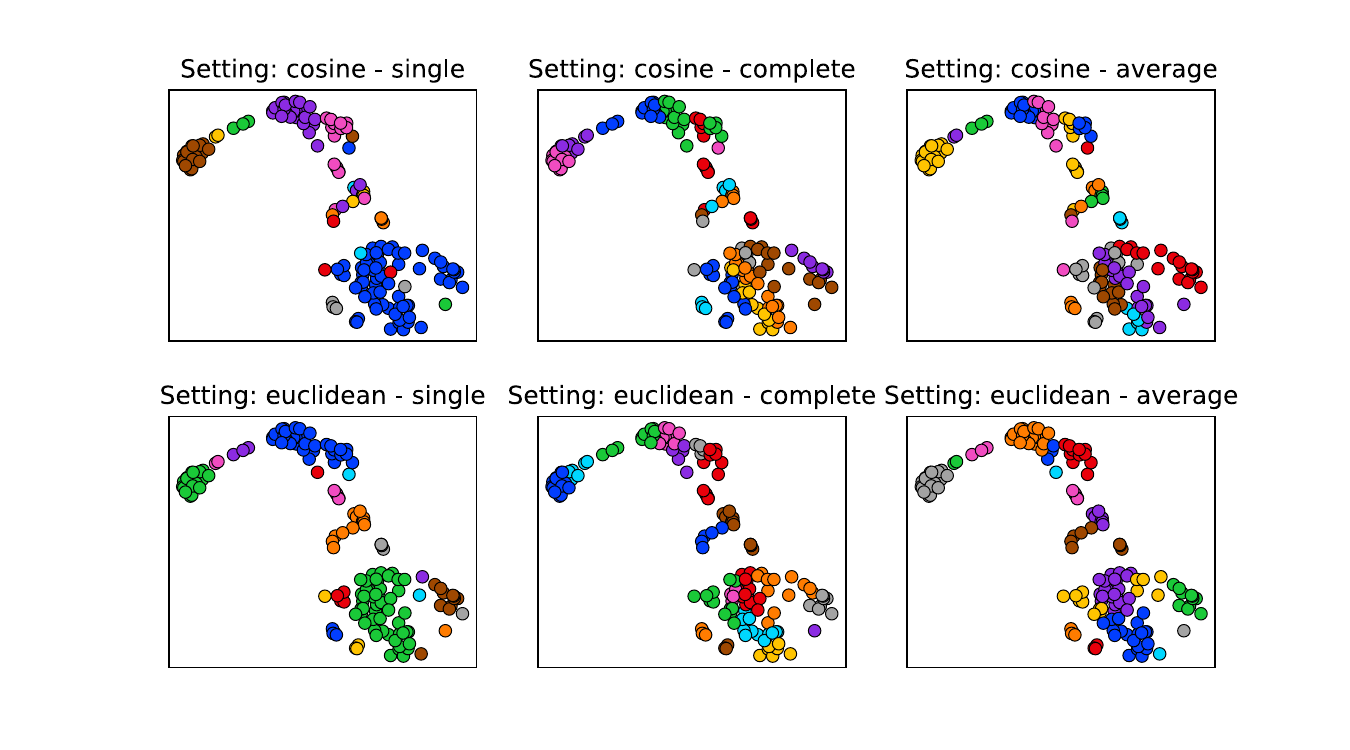}
                    \caption{The clusters illustrating different options for fine clustering. The first row contains the clustering result using the \texttt{cosine} distance while the second one is for \texttt{euclidean} $L2$ distance.}
                    \label{fig:method-model-cluster}
                \end{figure}

                An example illustrating the use of different types of distance together with various kinds of affinity rules is demonstrated in Figure \ref{fig:method-model-cluster}.
                
            By employing this approach, we achieve a hierarchical clustering that effectively propagates information from the global level $\mathbf{\hat{E}}$, capturing the relationships between different visual parts of the video, to the local level $\mathbf{c}$, which consists of the frames within each cluster. This global-to-local propagation enables us to extract semantically meaningful clusters that reflect both the overall video content and the finer details within specific clusters.

        \subsubsection{Semantic Partitioning}
            Following the contextual clustering step, each sampled frame $\hat{I}_i$ is assigned a label ${c}_i$ corresponding to its cluster index, which is used to create several \textbf{non-overlapping} partitions $\mathcal{P}$ in this step with multiple sub-steps to ensure the partitions' semantics are as meaningful as possible. These sub-steps include an outlier elimination that removes possible outliers from the initial partitions and a refinement step which consolidates smaller partitions into larger one with a specific threshold $\epsilon$. The later is supposed to compose the elemental pieces of information presented in small partitions into semantically meaningful knowledge under a larger partition.

            \paragraph{Outlier handling}
                Before partitioning the clustered samples into semantic partitions, we need to handle potential outliers where a frame may not perfectly align with its true cluster, a smoothing operation is applied to these labels. This smoothing process involves assigning the final label ${\hat{c}}_i$ of each frame by taking a majority vote $m(\cdot)$ among its consecutive neighboring frames $\left\{{c}_j \mid {|i - j| < \hat{W}}\right\}$ of window size $W = 2\hat{W} + 1$, with the mode value being used as the final label. This post-processing step helps ensure more accurate labeling and reduces the impact of isolated misclassifications. In its realization, the smoothing can be seen as a convolution $\mathcal{C} = \mathbf{c} \ast m$ between the vector $\mathbf{c}$ containing clustering result and the mode function denoted in Equation \ref{eq:mode-function}.
    
                \begin{equation}
                \label{eq:mode-function}
                    m(\mathbf{x}) = \arg\max_{{x}_i \in \mathbf{x}} \left\{|{x}_j = {x}_i| \mid {x}_j \in \mathbf{x} \right\}
                \end{equation}

            \paragraph{Initialization}
                Once the frames have been assigned their final labels $\mathcal{C}$, they are partitioned into several sections $\mathcal{P}$ based on these labels. The following definitions govern the initialization of this partitioning process:
            
                \begin{itemize}
                    \item A section $\mathcal{P}_i$ is defined as a vector containing indexes of samples forming a consecutive segment, meaning that $\mathcal{P}_i = [p_i, p_i + N_i)$ with $p_i$ is the start of $i$-th section and $N_i$ is the length of same section, satisfying that $p_i > 0$ and $p_i + N_i = \hat{T} + 1$.
                    \item The sections shall be consecutive without overlapping on each other, this means that for two sections next to each other $\mathcal{P}_i$ and $\mathcal{P}_{i + 1}$, the element that is immediately after the end of first section is the start of the next one $p_{i+1} = p_i + N_i$.
                    \item The sections are initialized with consecutive segment of samples that share similar labels of final clusters $\mathcal{C}$. Meaning that samples in a section $\mathcal{P}_i$ belong to the same clusters (the intra-cluster condition $\mathcal{C}_j = \mathcal{C}_k, \forall{j, k} \in \mathcal{P}_i$) and a section is bounded with samples from other clusters (the inter-cluster condition $\mathcal{C}_{p_i - 1} \neq \mathcal{C}_{p_i}$ and $\mathcal{C}_{p_i + N_i - 1} \neq \mathcal{C}_{p_i + N_i}$).
                \end{itemize}
            
                By partitioning the frames into sections based on their final labels, we create distinct segments that represent coherent subsets of the video content.

            \paragraph{Partition refinement}
                The semantic partitioning obtained from the initialization of this process $\mathcal{P} = \left\{\mathcal{P}_i\right\}_{i = 1}^{\hat{N}}$ contains $\hat{N}$ sections which would then be progressively refined with length condition as illustrated in the Algorithm \ref{alg:method-partitioning}.

                \begin{algorithm}
                \caption{Algorithm for refining partitions $\mathcal{P}$ with minimum length of $\epsilon$.}
                \label{alg:method-partitioning}
                \begin{algorithmic}[1]
                \State $N \leftarrow \hat{N}$
                \While{$\epsilon > \min_{i \in [1, N]} N_i$}
                    \State Find the index of the shortest partition $\hat{i} \leftarrow \arg\min_{i \in [1, N]} N_i$
                    \If{$\hat{i} = 1$}
                        \State $\mathcal{P}_2 \leftarrow [1, p_2 + N_2)$
                        \State $N_2 \leftarrow N_1 + N_2$
                    \ElsIf{$\hat{i} = N$}
                        \State $\mathcal{P}_{N - 1} \leftarrow [p_{N - 1}, L]$
                        \State $N_{N - 1} \leftarrow N_{N - 1} + N_N$
                    \Else
                        \State Merge left side $\mathcal{P}_{\hat{i} - 1} \leftarrow \left[p_{\hat{i} - 1}, p_{\hat{i}} + \left\lceil\frac{N_{\hat{i}}}{2}\right\rceil\right)$
                        \State Update left partition $N_{\hat{i} - 1} \leftarrow N_{\hat{i} - 1} + \left\lceil\frac{N_{\hat{i}}}{2}\right\rceil$
                        \State Merge right side $\mathcal{P}_{\hat{i} + 1} \leftarrow \left[p_{\hat{i}} + \left\lceil\frac{N_{\hat{i}}}{2}\right\rceil, p_{\hat{i} + 1} + N_{\hat{i} + 1}\right)$
                        \State Update right partition $N_{i + 1} \leftarrow N_{\hat{i} + 1} + \left\lfloor\frac{N_{\hat{i}}}{2}\right\rfloor$
                    \EndIf
                    \State Update the indexes of $\mathcal{P}$ accordingly
                    \State $N \leftarrow N - 1$
                \EndWhile
                \end{algorithmic}
                \end{algorithm}

                After the execution of the algorithm, all $N$ partitions in $\mathcal{P}$ have lengths of at least $\epsilon$ (.i.e, $N_i \geq \epsilon$). The naive version of algorithm has a runtime complexity of at most $\mathcal{O}(n^2)$ as each iteration requires $\mathcal{O}(\hat{N})$ operations to search for the appropriate partitions and there are $\mathcal{O}(\hat{N})$ iterations in the worst case where all partitions have to be merged together. A better optimization of approximately $\mathcal{O}(\hat{N} \log \hat{N})$ can be achieved with the support from an auxiliary data structure such as Binary Search Tree \cite{bst}.
                
            This partitioning result allows us to focus on individual semantic parts within the video and analyze their characteristics independently, enabling more detailed analysis and summary generation in subsequent stages.
    
    \subsection{Keyframes and Importance Scores}
    \label{subsec:method-model-keyframes-importance}
        After the partitioning step of the previous stage, the resulted partitions $\mathcal{P}$ is used to generate the keyframes $\mathbf{k}$ which carries important information of the original input that serves as bases for constructing the final summary. After which, an importance score ${v}_i$ is calculated for every sampled frame ${\hat{I}}_i$ which signifies the importance of this sample as it contains necessary parts to cover the input's information.

        \subsubsection{Keyframes}
        \label{subsubsec:method-model-keyframes}
            The set of keyframes $\mathbf{k}$ is a subset of the indexes of sampled frames $\mathbf{k} \subset \mathbf{t}$ which is originally denoted in the sampling step of of our approach at \ref{subsubsec:method-model-generating-sampling}. Furthermore, this set is a union of multiple smaller sets of partition-wise keyframes $\mathbf{k}^{(i)}$, which is extracted from $i$-th resulting semantic section $\mathcal{P}_i$. The number of keyframes $|\mathbf{k}^{(i)}|$ extracted depends on different settings. There are three different options for extracting keyframes from any individual partition which are listed below.
            
            \begin{itemize}
                \item \textit{Mean} setting: For each section $\mathcal{P}_i$, a keyframe $k \in \mathbf{k}^{(i)}$ is selected as the frame $k$ whose reduced embedding $\mathbf{\hat{e}}_k$ is closest to the mean embedding of that section $\mathbf{\hat{e}}^{(i)} = \frac{1}{N_i} \sum_{j = 1}^{N_i} \mathbf{\hat{e}}_j$. This setting results in one additional keyframe $k$ per section. Detailed information is given in Equation \ref{eq:method-keyframe-mean}.

                \begin{equation}
                \label{eq:method-keyframe-mean}
                    k = \arg\min_{j \in \mathbf{P}_i} d\left(\mathbf{\hat{e}}_j, \mathbf{\hat{e}}^{(i)}\right)
                \end{equation}
                
                \item \textit{Middle} setting: The keyframe $k \in \mathbf{k}^{(i)}$ is chosen as the frame located in the middle of each section $\mathcal{P}_i$. It means that the keyframe of this setting is formulated as $k = p_i + \left\lfloor\frac{N_i}{2}\right\rfloor$. This setting also yields one additional keyframe $k$ per section.
                
                \item \textit{Ends} setting: Frames at the beginning $p_i$ and end $p_i + N_i - 1$ of each section are selected as keyframes. This setting produces two additional keyframes per section.
            \end{itemize}

            Please note that the above options can be further combined into advanced settings such as \textit{Mean} + \textit{Ends} or \textit{Mean} + \textit{Middle}, leading to the set of keyframes $\mathbf{k}^{(i)}$ for each partition with different sizes (\textit{.i.e}, $2$ for \textit{Mean} + \textit{Middle} and $3$ for \textit{Middle} + \textit{Ends}).
        
            \begin{figure}[ht]
                \centering
                \includegraphics[width=0.73\paperwidth]{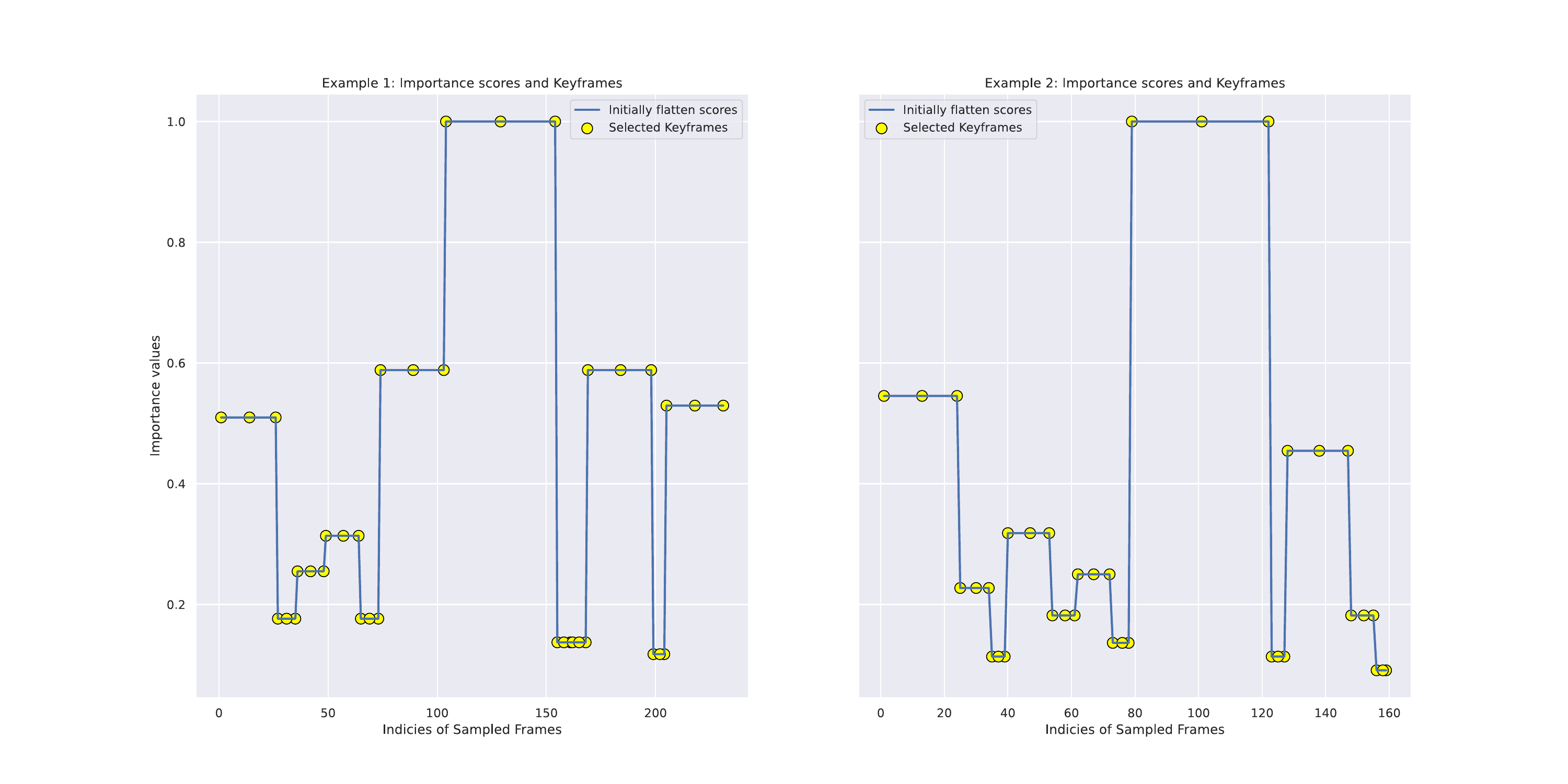}
                \caption{Two examples of our selected keyframes with the rule \textit{Middle} + \textit{Ends} where a horizontal segment denotes a single partition.}
                \label{fig:method-model-keyframe}
            \end{figure}

            Two examples demonstrating the selection of keyframes according to the rule \textit{Middle} + \textit{Ends} in this proposed work are given in Figure \ref{fig:method-model-keyframe}. In this figure, the sampled frames $\mathbf{\hat{I}}$ of each example $\mathbf{I}$ are decomposed into several semantic partitions $\mathcal{P}$ with the application of previous stages, which are shown as multiple horizontal segments at distinct vertical altitudes. These altitudes denote the lengths of such segments, meaning that a segment with higher altitude is longer than a lower one, so as to demonstrate the difference between those partitions. With the rule \textit{Middle} + \textit{Ends}, a total of $3$ keyframes are selected per each segment at the position of segment's start, midpoint, and end. This means that a segment $\mathcal{P}_i$ provides a set of keyframes $\mathbf{k}^{(i)} = \left\{p_i, p_i + \left\lceil\frac{N_i}{2}\right\rceil, p_i + N_i  -1\right\}$.
    
        \subsubsection{Importance Scores}
        \label{subsubsec:method-model-importance}
            The individual importance scores ${v}_i$ of all sampled frames ${\hat{I}}_i$ form a vector of importances $\mathbf{v} \in \mathbb{R}^{\hat{T}}$. This importance values would contribute to the decision of whether the frame may be included in the final summary or not.
        
            In the computation of each sampled frame's importance, we initialize the importance score $\mathbf{\hat{v}}$ to be the length of the section it belongs to. The formulation of such scores is detailed under Equation \ref{eq:method-importance}. The assumption behind this initialization is that segments of the video focusing on similar visual information for a longer duration provide necessary information for the summary.

            \begin{equation}
            \label{eq:method-importance}
                {\hat{v}}_j = N_i, \forall{j} \in [p_i, N_i)
            \end{equation}
        
            The final importance score of each sample ${v}_i$ is computed by scaling the initialized value ${\hat{v}}_i$ using a keyframe-biasing method. This method takes into account the proximity of frames to the keyframes $\mathbf{k}$ and assigns higher importance to sampled frames closer to the keyframes compared to normal samples.

            Various schemes for biasing the importance scores are implemented for all settings of keyframe selection except \textit{Mean}. In general, for each partition $\mathcal{P}_i$, there are several \textit{keypoints} $\mathcal{K}$ representing important points in the calculation of importances for all other samples inside that partition. These keypoints can be categorized into two types as follow:
            
            \begin{itemize}
                \item \textit{High} keypoints: The positions of all keyframes $\mathbf{k}^{(i)}$ in the partition are set as high keypoints whose importances are highest among all frames in the partition $\mathcal{P}$. In other words, $\mathcal{K}_H = \mathbf{k}^{(i)}$.
                \item \textit{Low} keypoints: The positions of frames whose locations are furthest from any of the partition's keyframes $\mathbf{k}^{(i)}$. Detailed formulation of such low keypoints is illustrated under Equation \ref{eq:method-model-keyframe-low}.

                \begin{equation}
                \label{eq:method-model-keyframe-low}
                    \mathcal{K}_L = \left\{\arg\max_{j \in \mathcal{P}_i} \left(\min_{k \in \mathbf{k}^{(i)}} |j - k|\right)\right\}
                \end{equation}
            \end{itemize}

            Several biasing options are given to either increase the importance of keyframes compared to the flat scores $\mathbf{\hat{v}}$ or decrease the scores of others compared to that filling importances. Details are outlined in the following list with $B$ parameterize this biasing scheme:
        
            \begin{itemize}
                \item Increase the importances of keyframes: The importance scores of high keypoints $\mathcal{K}_H$ are finalized with value $v_i = \hat{v}_i(1 + B), \forall{i} \in \mathcal{K}_H$ while those of low keypoints are set to be the flat scores $v_i = \hat{v}_i, \forall{i} \in \mathcal{K}_L$. In this setting, the parameter $B \in \mathbb{R}^+$.
                \item Decrease the scores of other sampled frames: The importance scores of high keypoints are assigned with the filling scores $\mathbf{\hat{v}}$ that are previously computed $v_i = \hat{v}_i, \forall{i} \in \mathcal{K}_H$ while the low keypoints' importances are biased toward zero by the parameter $v_i = \hat{v}_i(1 - B), \forall{i} \in \mathcal{K}_L$. For this option, $B \in [0, 1]$.
            \end{itemize}
    
            Base on the found keypoints $\mathcal{K}$ as well as the initialized importances at positions of those keypoints, different interpolating methods are then used to fill the importance scores of samples between key positions that are listed in the below list. In particular, the importance of a sampled frame $v_i$ is computed based on its nearest keypoints $j \in \mathcal{K}$ and $k \in \mathcal{K}$ satisfying that $i \in (j, k)$ as well as $\nexists l \in \mathcal{K}: l \in (j, k)$, together with their final scores $v_j$ and $v_k$. These methods help assign relative importance scores to frames based on their positions between keyframes.

            \begin{itemize}
                \item \textit{Cosine} interpolation: The importances of samples in between the keyframes are interpolated based on a \verb|cosine| scheme described in Equation \ref{eq:method-model-interpolate-cosine}.
                
                \begin{equation}
                \label{eq:method-model-interpolate-cosine}
                    {v}_i = \frac{{v}_j - {v}_k}{2} \cdot \cos\left(\frac{\pi (i-j)}{k-j}\right) + \frac{{v}_j + {v}_k}{2}
                \end{equation}

                \item \textit{Linear} interpolation: The importances of samples in between the keyframes are interpolated based on a \verb|linear| scheme described in Equation \ref{eq:method-model-interpolate-linear}.
                
                \begin{equation}
                \label{eq:method-model-interpolate-linear}
                    {v}_i = {v}_j + \frac{{v}_k - {v}_j}{k-j} \cdot (i-j)
                \end{equation}

            \end{itemize}

\begin{figure}[ht]
                \centering
                \includegraphics[width=0.73\paperwidth]{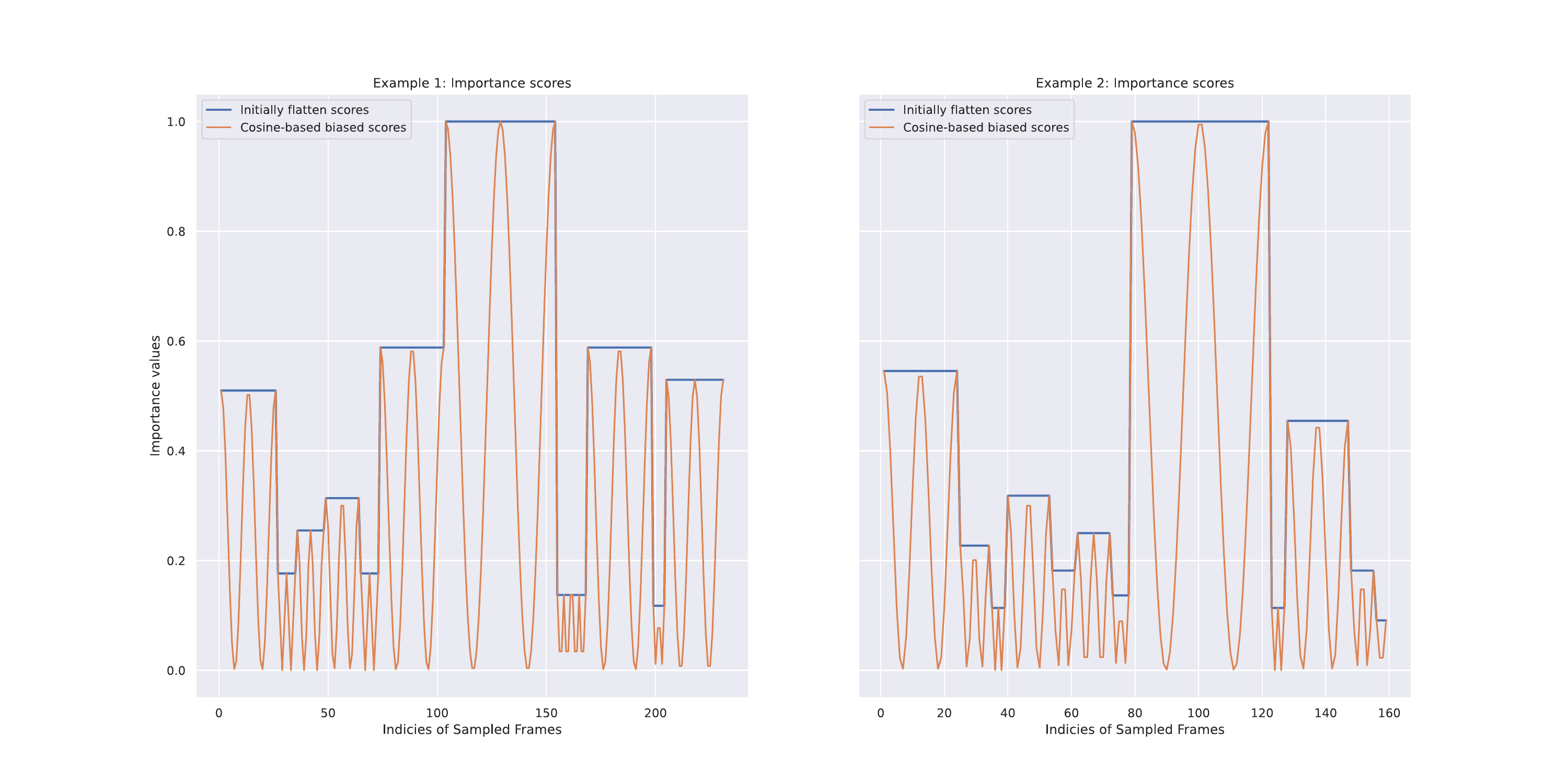}
                \caption{Comparison between cosine-interpolated scores and flat scores are demonstrated for two examples.}
                \label{fig:method-model-importance}
            \end{figure}

            Examples illustrating the difference between cosine-interpolated importances and flat scores are given in Figure \ref{fig:method-model-importance} in which the fluctuating yellow lines represent cosine-interpolation while the blue ones show the associated flat scores.
            
            By determining the importance scores of frames, we can prioritize and select the most significant frames for inclusion in the video summary, ensuring that keyframes and important frames are appropriately represented.
    
    \subsection{Summary Generation}
    \label{subsec:method-model-summary}
        The final summary $\mathbf{S} = \{{S}_i\}_{i = 1}^{L}$ is eventually created in this stage with the process depending on the generation's purpose. According to the convention of evaluation adopted by prior works \cite{apostolidis2021pglsum, apostolidis2020unsupervised}, we provide a specific algorithm to generate summary based on prior segmentations. Besides that, we also describe a general way to construct the summary that is more straightforward and efficient.
    
        \subsubsection{For usable results}
            To generate a usable summary, the target number of frames in the final summary $L'$ is determined based on a specified proportion of the original video's length $\mathcal{R} \in (0, 1)$. However, the length of the final summary is also constrained by a maximum limit in time $\mathcal{L}$. The detailed formulation is given in Equation \ref{eq:method-summary-length} where $r_O$ is the frame-rate of the output summary.

            \begin{equation}
            \label{eq:method-summary-length}
                L' = \min\left(T \cdot \mathcal{R}, \frac{\mathcal{L}}{r_O}\right)
            \end{equation}

            The number of keyframes that are considered to generate the final summary is $\hat{L} = \min(L', |\mathbf{k}|)$, this means that some of the selected keyframes may not be used for summary generation in case of target length $L'$ requires a number of frames smaller than the keyframes being selected. For such case, $\hat{L}$ keyframes whose importances are highest would be joined together to create the indexes of summary samples $\mathbf{\hat{s}} \subset \mathbf{k}$ from which the sample summary $\mathbf{\hat{S}}$ is constructed with ${\hat{S}}_i = {\hat{I}}_{{\hat{s}}_i}$. As there is not more positions for other frames in this summary, the sample summary is used as final result in this case $\mathbf{S} = \mathbf{\hat{S}}$.

            In the other case where $L' > |\mathbf{k}|$, all the keyframes selected in the previous steps are included in the final summary as summarizing frames. This makes the indexes of sample summary equal to the set of keyframes $\mathbf{\hat{s}} = \mathbf{k}$. Besides the keyframes which are directly used in the summary ${t}_k \in \mathbf{s}, \forall{k} \in \mathbf{k}$, the original frames $I_t$ surrounding them are also used with a fixed number of frames following each keyframe. This fixed number $\ell$ is calculated based on the desired size of the target summary $L'$, ensuring that the final summary meets the requirements in length. Such calculation is given in the Equation \ref{eq:usable-summary-length} below for further references.
    
            \begin{equation}
                \label{eq:usable-summary-length}
                \ell = \left\lfloor\frac{L'}{|\mathbf{k}|}\right\rfloor
            \end{equation}

            The additional frames are selected from the two-sided consecutive segment $[t_k - \ell_H, t_k + \ell_H]$ around each keyframe $k$ with the distance of at most $\ell_H = \left\lfloor\frac{\ell - 1}{2}\right\rfloor$. The indexes of these selected keyframes and frames are then mapped back onto the original input domain $[1, T]$ through the set $\mathbf{t}$, creating a binary vector $\mathbf{y} \in \{0, 1\}^{T}$ that describes the final selection of summary frames. In the end, the selection process can be detailed as in Equation \ref{eq:method-model-summary-usable}, providing overall understanding of the final summary's generation.

            \begin{equation}
            \label{eq:method-model-summary-usable}
                \mathbf{y}_i =
                \begin{cases}
                    1 & \text{if } \ell_H \geq \min_{k \in \mathbf{k}} \left|i - {t}_k\right| \\
                    0 & \text{otherwise }
                \end{cases}
            \end{equation}
            
            Finally, the indexes of frames existing in the final summary $\mathbf{s}$ is extracted from the selection vector $\mathbf{y}$ as $\mathbf{s} = \left\{i \mid {y}_i = 1\right\}$. The summary $\mathbf{S}$ is then computed following that ${S}_i = {I}_{{s}_i}$.
    
            By including the keyframes $\mathbf{k}$ and their surrounding frames, the generated summary $\mathbf{S}$ captures the essential content of the video in continuity while maintaining the desired length and coherence.

        \subsubsection{For evaluation purposes}
            For evaluating the performance of the generated summaries, we adopt a method used in previous works for automatic evaluation. The importance scores $\mathbf{v}$ obtained from the previous stage, along with the available segmentations $\mathcal{S}$ from the dataset, are used to generate the summary $\mathbf{S}$ required for evaluation procedure.

            Firstly, the importances $\mathbf{v}$ is extrapolated from samples domain $[1, \hat{T}]$ to the original input domain $[1, T]$, denoted as $\mathcal{I} \in \mathbb{R}^{T}$. There are two approaches in this extrapolation that are listed below:

            \begin{itemize}
                \item \textit{Linear} approach: The importance score of $i$-th frame is calculated by the importances of two samples which are closest to it in both directions. Formally, $j$ and $j + 1$ are sample indexes of two samples that are nearest to $i$ satisfying that ${t}_{j} \leq i \leq {t}_{j + 1}$. The importances of these keyframes ${v}_{j}$ and ${v}_{j + 1}$ are used to compute the final importance of $i$-th frame. This computation employs a weighted average demonstrated by Equation \ref{eq:method-model-linear}.

                \begin{equation}
                \label{eq:method-model-linear}
                    \mathcal{I}_i = \frac{|{t}_{j} - i|{v}_{j} + |i - {t}_{j + 1}|{v}_{j + 1}}{|{t}_{j} - i| + |i - {t}_{j + 1}|}
                \end{equation}
                
                \item \textit{Nearest} approach: The importance score of $i$-th frame is filled with that of the nearest sample $\mathcal{I}_i = {v}_k$ where $k$ is the sample index of the nearest sample, determined in the Equation \ref{eq:method-model-nearest}.
                
                \begin{equation}
                \label{eq:method-model-nearest}
                    k = \arg\min_{j \in [1, \hat{T}]} |i - {t}_j|
                \end{equation}
                
            \end{itemize}
        
            After the extrapolation of $\mathcal{I}$, the machine-generated summary is generated by selecting a subset of segmentations $\mathcal{S}$ from the input video. The result of such selection is denoted with a vector $\mathbf{y}$ where ${y}_i = 1$ means that the segment $\mathcal{S}_i$ is selected, and ${y}_i = 0$ means otherwise. This selection aims to keep the total length of the selected segments, or the number of summarizing frames in other words, below the thresholding length $L'$ while maximizing the sum of importances associated with these segments. This optimization problem drives us to the Knapsack requirements specified by the Equation \ref{eq:method-model-knapsack}.

            \begin{equation}
            \label{eq:method-model-knapsack}
                \begin{aligned}
                    & \text{maximize} && \sum_{i=1}^{S} \mathcal{J}_i {y}_i \\
                    & \text{subject to} && \sum_{i=1}^{S} \left|\mathcal{S}_i\right| {y}_i \leq L_T, \\
                    & && {y}_i \in \{0, 1\}, \quad i \in \left[1, S\right].
                \end{aligned}
            \end{equation}

            The importance value of each segmentation $\mathcal{J}_i$ is calculated as the sum of the importances of its constituent frames $j \in \mathcal{S}_i$ following Equation \ref{eq:method-model-segment-importance} with $S = |\mathcal{S}|$ is the number of segments presented for the input video $\mathbf{I}$.
            
            \begin{equation}
            \label{eq:method-model-segment-importance}
                \mathcal{J}_i = \sum_{j \in \mathcal{S}_i} \mathcal{I}_j
            \end{equation}

            Similar to the generation of usable summary, the final summary $\mathbf{S}$ is then computed from the selection vector $\mathbf{y}$.

        

\section{Implementation Details}
\label{section:method-details}

    In this section, we provide comprehensive details on the techniques and pre-trained models utilized in our approach.

    \subsection{Embedding Model}
    \label{subsec:method-details-emb}
        In our implementation, we employ the pre-trained model DINO \cite{dino} provided by Hugging Face with the model path \verb|facebook/dino-vitb16|\footnote{Docs available at \url{https://huggingface.co/facebook/dino-vitb16}} to generate embeddings for each frame. DINO has been pre-trained on a large-scale dataset using the distillation approach without relying on labeled data. It captures valuable visual information by contrasting image patches within an individual frame, allowing it to learn meaningful representations in an unsupervised manner. DINO-ViT-B16 is a variant of the DINO model that combines the power of self-supervised learning and vision transformers. It leverages the Vision Transformer (ViT) architecture as its backbone and uses a patch size of \(16 \times 16\) pixels.
    
        First, the input frame $\mathbf{\hat{I}_t} \in \mathbb{R}^{W\times H\times C}$ is processed using the pre-trained image processor associated with the DINO model. The output is an image $\mathbf{\hat{I}'}_t \in \mathbb{R}^{224\times 224\times 3}$. We then feed $\mathbf{\hat{I}'}_t$ into DINO, which returns the embeddings $\mathbf{E'}_t \in \mathbb{R}^{197\times 768}$. To extract the contextual information of this sample, we select the first vector in its output embedding $\mathbf{E'}_t[0]$ as the semantic embedding of the sample, which corresponds to the \textit{classifier} token in the input. This token represents the learned context by the pre-trained model. We concatenate the vector $\mathbf{E}_t = \mathbf{E'}_t[0]$ from all frames in $\mathbf{\hat{I}}$ to obtain the contextual embedding $\mathbf{E} \in \mathbb{R}^{\hat{T} \times D}$ of the input where $D = 768$ for DINO pre-trained model.

    \subsection{Clustering Model}
    \label{subsec:method-details-clustering}
        For dimension reduction, we utilize models from \verb|scikit-learn|\footnote{Docs available at \url{https://scikit-learn.org/stable/}}. Specifically, we use the PCA implementation provided by \verb|sklearn.decomposition.PCA|\footnote{Docs available at \url{https://scikit-learn.org/stable/modules/generated/sklearn.decomposition.PCA.html}}and the t-SNE implementation provided by \verb|sklearn.manifold.TSNE|\footnote{Docs available at \url{https://scikit-learn.org/stable/modules/generated/sklearn.manifold.TSNE.html}}. We first apply PCA to reduce the dimensionality to 6. This helps in capturing the most important features of the data. Subsequently, we apply t-SNE to further reduce the dimensionality to 2. The \verb|euclidean| metric is used for distance calculations during the t-SNE process. This combination of PCA and t-SNE allows us to visualize the high-dimensional data in a lower-dimensional space, facilitating better understanding and analysis of the embeddings.
    
\begin{figure}[ht]
            \centering
            \includegraphics[width=0.73\paperwidth]{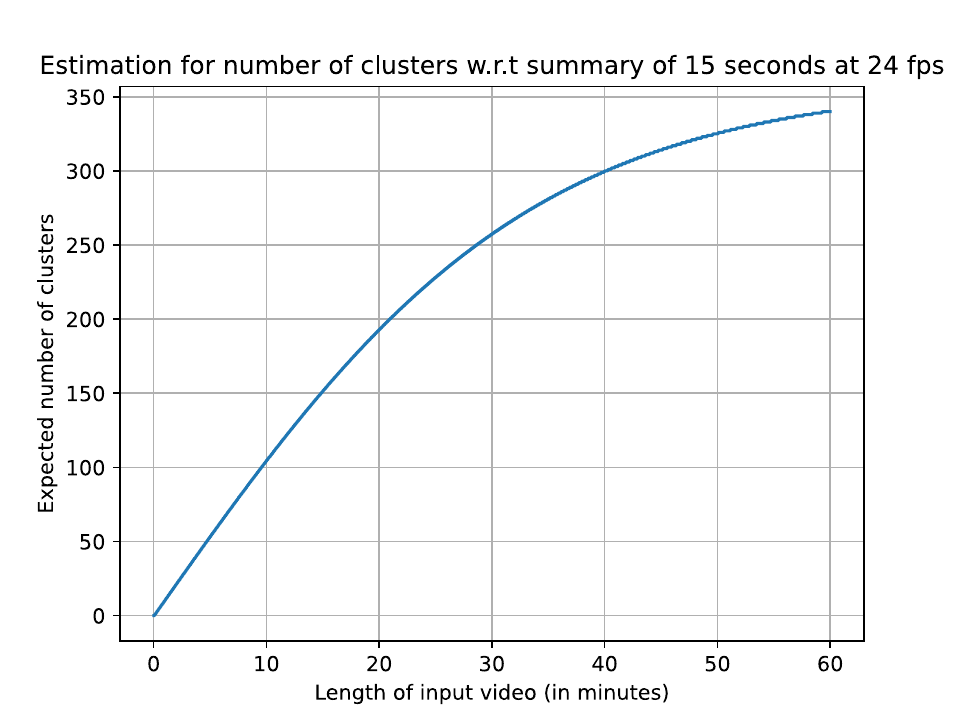}
            \caption{Actual number of clusters calculated in the contextual clustering step, with maximum summary length of $\mathcal{L} = 15$ seconds and output frame-rate $r_O = 24$ fps.}
            \label{fig:exp-implementation-num-cluster}
        \end{figure}

        In the contextual clustering step, we utilize the BIRCH implementation provided by \verb|sklearn.cluster.Birch|\footnote{Docs available at \url{https://scikit-learn.org/stable/modules/generated/sklearn.cluster.Birch.html}} to form coarse clusters. This is followed by applying Agglomerative Clustering using \verb|sklearn.cluster.AgglomerativeClustering|\footnote{Docs available at \url{https://scikit-learn.org/stable/modules/generated/sklearn.cluster.AgglomerativeClustering.html}} to generate fine clusters. The number of clusters, denoted as $K$, is computed using Equation \ref{eq:method-model-num-cluster} with a modulation parameter $Z=10^{-3}$. Figure \ref{fig:exp-implementation-num-cluster} illustrates the actual number of clusters computed in our implementation with the aforementioned configurations.
    
    \subsection{Relevant Hyper-Parameters}
    \label{subsec:method-details-param}
        In the sampling step before feature extraction, the video is sampled with a target frame rate of $R = 4$. In the semantic partition step, we set the window size $W$ for mode convolution to $5$, and the minimum length $\epsilon$ for each segment is assigned a value of $4$.
    
        For keyframe selection, we employ the setting \textit{Middle} + \textit{End}, which involves selecting keyframes from the middle and end portions of each partition. In the importance scoring step, we use \verb|cosine| interpolation to assign scores to samples that are not keyframes. The keyframe biasing scheme is parameterized with $B=0.5$ and set to \textit{Increase the importances of keyframes}.
    
        To obtain a usable summary, we set the summarization rate $\mathcal{R}$ to $0.2$ (or $20\%$) and the maximum summary length $\mathcal{L}$ to $120$ seconds (or $2$ minutes). Finally, the output frame rate $r_O$ is set equal to the input frame rate $r_I$. These settings help control the length and speed of the generated video summary, ensuring that it aligns with the desired requirements and constraints.

\section{Human-Centric Evaluation of Video Summaries}
\label{section:method-evaluation}

    To address the limitations of conventional evaluation metrics for video summaries, we have proposed a novel evaluation pipeline that incorporates human evaluation. The objective of this approach is to replicate the way humans summarize long videos into concise content. By integrating human judgment, we aim to capture the subjective aspects of video summarization and provide a more comprehensive evaluation framework.
    
    The inclusion of human evaluation in our pipeline allows us to consider factors such as semantic relevance, coherence, and overall subjective quality, which are often challenging to quantify objectively. By leveraging human expertise and perception, we can gain valuable insights into the effectiveness of video summarization algorithms and their ability to capture the essence of the original video content.
    
    \subsection{Video Set}
    \label{subsec:method-eval-video}
        In our evaluation process, we present evaluators with different types of video sets and ask them a variety of questions. There are three types of video sets that are randomly shown to evaluators:
        
        \begin{itemize}
            \item \textbf{Original video~} These videos are not shortened in any way and serve as a basis for evaluating the shortened summaries. The purpose of putting the original videos into our evaluation is to collect ground truth-like information in the form of answers to the prescribed questions. These labels would be referenced in the process of computing metric values for any of the summaries shown below. Note that the answers to questions for these videos are directly used as inputs in the computation of a summary's metric values.
            \item \textbf{Summary video~} This type of video set includes two subtypes:
                \begin{itemize}
                    \item \textbf{User summary~} These summaries are provided with the dataset and are generated by users. These summaries are evaluated in order to create baseline scoring values for each of the given videos so that every automatic method can be compared to get experimenters a direct understanding of the automation's performance. The answers from these summaries are not used to calculate the metric scores of the automated summaries but to be compared with them as human-level baselines.
                    
                    \item \textbf{Machine-generated summary~} These summaries are extracted using an automated mechanism, our proposed method for example.
                
                \end{itemize}
            
            \item \textbf{A pair of videos~} In this case, evaluators are shown a pair consisting of one original video and one summary video. For this type, the evaluators are aware of which video is the original and which one is the summary.
        
        \end{itemize}
    
    \subsection{Questionnaire}
    \label{subsec:method-eval-questionnaire}
        Along with each video set, evaluators are presented with different types of questions. There are three question types in which two comprise nominal selections while the other presents ordinal scoring.

        \paragraph{Nominal questions}
            All questions in this type providing evaluators with a question of multiple options $\mathbf{O} = \{O_i\}_{i = 1}^{C}$ and they are asked to generate an answer $\mathbf{A}$.
        
            \begin{itemize}
                \item \textit{Multiple choice~} Evaluators are asked to choose one answer, so that $\mathbf{A} = \{A\}, A \in \mathbf{O}$. For instance, they might be asked whether the scene in the video is mainly outside or inside. We use answers for the original video as the ground-truth, and compare answers for the summary with ground-truth to produce the accuracy. More specifically, given an answer $\mathbf{\hat{A}}$ for a summary and another answer $\smash{A}$ for the original video, the accuracy score for this pair is calculated with Equation \ref{eq:method-eval-mcq}.
                
                \begin{equation}
                \label{eq:method-eval-mcq}
                    h\left(\mathbf{\hat{A}}, \mathbf{A}\right) = 
                    \begin{cases}
                        1 & \text{if } \hat{A} = A \\
                        0 & \text{otherwise }
                    \end{cases}
                \end{equation}
                
                \item \textit{Checkbox~} This question type is similar to multiple choice, but evaluators can select more than one option for their answer $\mathbf{A} = \{A_i\}_{i = 1}^{\hat{C}}$ where $A_i \in \mathbf{O}, \forall{i} \in [1, \hat{C}]$ and $\hat{C} \in [1, C]$ is the number of selections by user. For example, they might be asked about the main subjects of the video and can choose multiple options such as humans, animals, objects, buildings, or none of the above. Then, we use Intersection over Union (IoU) to compare the summary distribution with the original distribution. In particular, given an answer $\mathbf{\hat{A}}$ for a summary and another answer $\smash{A}$ for the original video, the IoU score for this pair is calculated with Equation \ref{eq:method-eval-checkbox}.
                
                \begin{equation}
                \label{eq:method-eval-checkbox}
                    h\left(\mathbf{\hat{A}}, \mathbf{A}\right) = \frac{\left|\mathbf{\hat{A}} \cap \mathbf{A}\right|}{\left|\mathbf{\hat{A}} \cup \mathbf{A}\right|}
                \end{equation}
            \end{itemize}

            In a human centric survey, it is usual that there are multiple users answering the same questions on the summary while several others providing answers to that questions on the original video. In this evaluation, we denote the answers for a summary as $\mathbf{\hat{A}}_i$ where $i \in [1, \hat{U}]$ with $\hat{U}$ is number of users answered the question, and $\mathbf{A}_j$ as the answers for that summary's origin, for $j \in [1, U]$ in which $U$ is number of users giving answers to this question.

            To compute the score of a summary with respect to a question of this type $\mathcal{A}$, we first compute the scores for each answer of this summary $\mathcal{A}_i$ against the original video. That score is assigned with the maximum score such answer $\mathbf{\hat{A}}_i$ could achieve in a pair-wise scoring $h(\cdot, \cdot)$ with any answer $\mathbf{A}_j$ on the original video.

            \begin{equation}
            \label{eq:method-eval-nominal}
                \mathcal{A}_i = \max_{j \in [1, U]} h\left(\mathbf{\hat{A}}_i, \mathbf{A}_j\right)
            \end{equation}

            Afterward, the general score of that summary with respect to this specific question $\mathcal{A}$ is computed with the average of scores obtained by all of its answers $\mathcal{A} = \frac{1}{\hat{U}} \sum_{i = 1}^{\hat{U}} \mathcal{A}_i$.
            
        \paragraph{Linear}
            In this type of questions, evaluators are asked to provide a grade $a \in [1, C]$ on a scale from $0$ to $C$ for a specific aspect. Specifically, they will be asked to rate how much of the original video they would understand if they only watched the summary video.
            
            Similar to previous type of questions, a linear question can be scored by $U$ different users with scores denoted as $a_i, \forall i \in [1, U]$. For evaluation on this type of question, we construct a mean across the answers $\hat{a} = \frac{1}{U} \sum_{i = 1}^{U} a_i$ which would represents the general informativeness of this summary. Distinct summaries will then be compared based on these means to find out which one performs better.

            In order to map the scores of questions in this type to a scale similar to other types, the average score is normalized back to $[0, 1]$ by a simple division $\mathcal{A} = \frac{\hat{a}}{C}$.

        \paragraph{Summary-level and Method-level}
            As different summaries would have distinct numbers of questions associated with them, the evaluated score of $i$-th summary $\mathcal{U}_i$ is computed as an average of all of its questions' scores $\mathcal{A}^{(j)}$, meaning that $\mathcal{U}_i = \frac{1}{Q_{i}} \sum_{j = 1}^{Q_{i}} \mathcal{A}^{(j)}$ where $\mathcal{A}^{(j)}$ denotes the general score of $i$-th summary at $j$-th question and $Q_{i}$ is the number of questions that $i$-th summary has.
        
            In order to calculate the eventual metric $\mathcal{U}$ representing performance of a method that generate the summaries, an avarage of its summaries' scores $\mathcal{U}_i$ is used $\mathcal{U} = \frac{1}{|\mathbf{X}|} \sum_{i = 1}^{|\mathbf{X}|} \mathcal{U}_i$ where $\mathbf{X}$ is the given dataset and $|\mathbf{X}|$ is the size of that dataset.
    
    \subsection{Implementation of the Survey}
    \label{subsec:method-eval-implementation}
        While preparing and executing a survey implementing the aforementioned human-centric evaluation, we employ the PyWebIO package\footnote{Docs available at \url{https://pywebio.readthedocs.io/en/latest/}} to develop a website for conducting the human-centric evaluation. 
        
        \begin{figure}[ht]
            \centering
            \includegraphics[width=0.73\paperwidth]{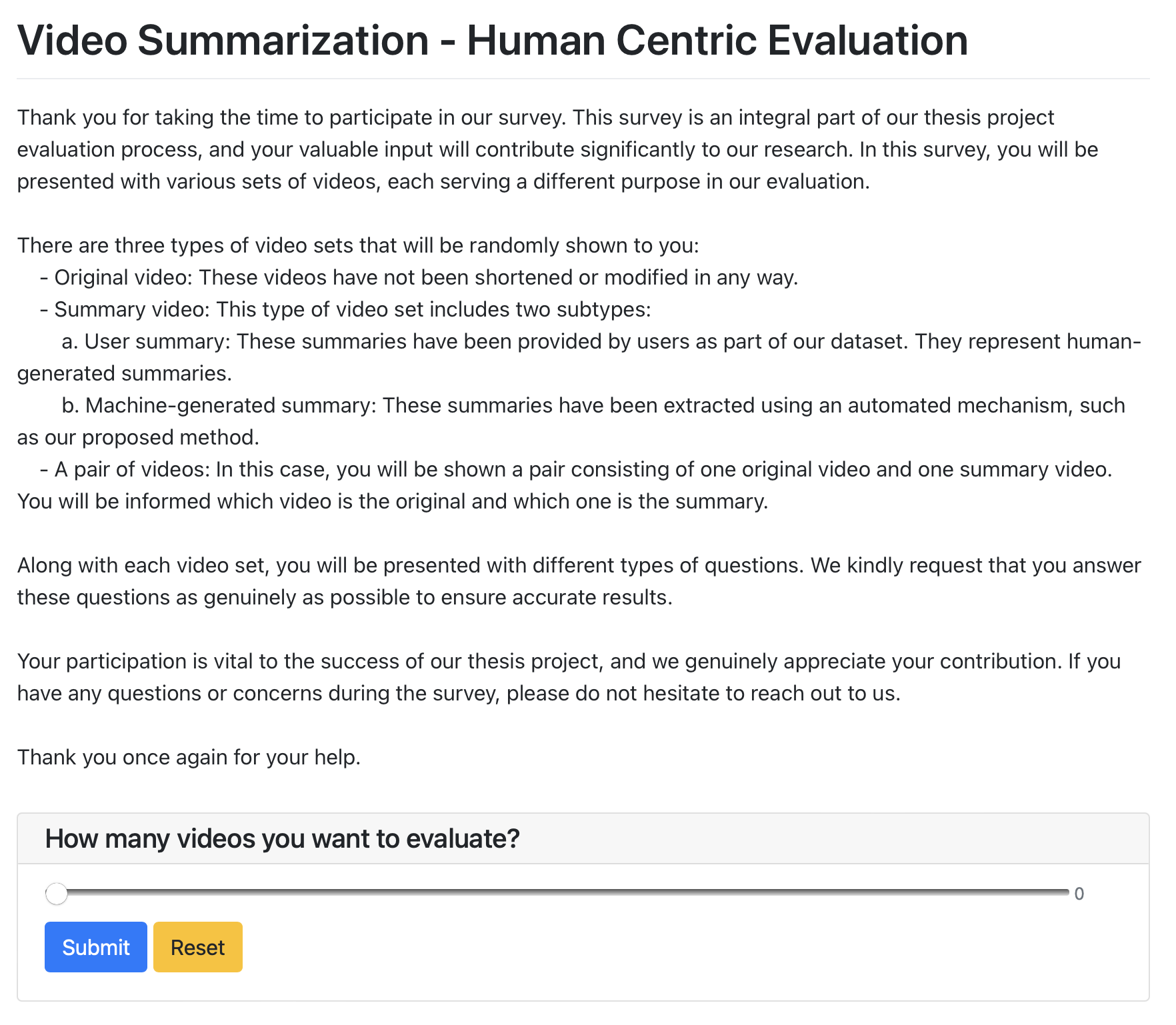}
            \caption{Introduction to the evaluation process, featuring a slider for evaluators to select the number of video sets for assessment}
            \label{fig:exp-imple-intro}
        \end{figure}
        
        As described in \ref{subsec:method-eval-questionnaire}, we design three types of questions for the evaluation: multiple choice, checkbox, and linear. To implement these question types, we utilize specific functions from the \verb|pywebio.input| module. The \verb|radio| function is employed for multiple choice questions, the \verb|checkbox| function for checkbox questions, and the \verb|slider| function for linear questions.
        
        To begin the evaluation process, participants are provided with a brief introduction that includes information about the video sets being evaluated, as well as an explanation of the accompanying questions. This introductory section aims to familiarize participants with the evaluation context and ensure they understood the task at hand.
        
        Once introduced, participants are prompted to select the number of video sets they wish to annotate. The video sets are then presented one by one, with each subsequent set being displayed only after participants have completed the questions for the previous set. 
        
        To provide a visual representation of the interface used for the introduction section, Figure \ref{fig:exp-imple-intro} is included. This figure showcases the user interface that participants interact with during the introduction phase of the evaluation process.
        
        Within each video set, the questions are presented to participants individually. To ensure accurate recording of participants' answers, we implement a mechanism where the answers are stored only when participants press the "Submit" button. This allows participants to review their responses and make any necessary adjustments before finalizing their answers.
        
\begin{figure}[h]
            \centering
            \includegraphics[width=0.73\paperwidth]{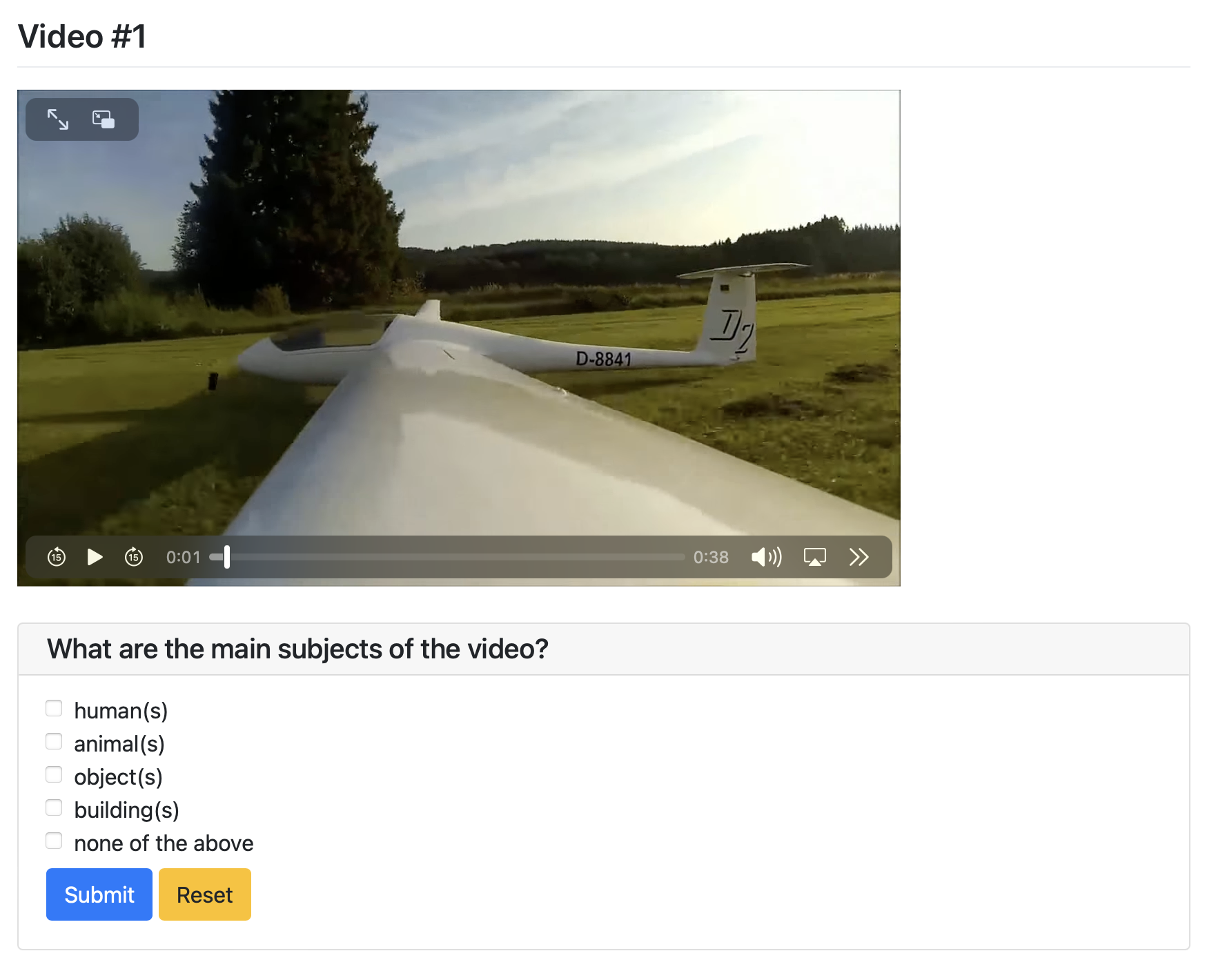}
            \caption{An example of a question in the evaluation process}
            \label{fig:exp-imple-example}
        \end{figure}

        To provide a visual example of a question in the evaluation process, please refer to Figure \ref{fig:exp-imple-example}. This figure illustrates an example question that participants encounter during the evaluation. The interface design may vary based on the specific question types and the overall user interface of the evaluation website.
        
        After completing all the video sets and associated questions, the answers provided by participants are stored anonymously. This ensures confidentiality and privacy in the evaluation process. Besides that, the scale of linear questions employed in our survey is set to $C = 10$.

\chapter{Experiments}
\label{chapter:experiments}

\begin{ChapAbstract}
    This chapter presents a comprehensive overview of the experiments conducted to evaluate the effectiveness of the proposed method. It includes details of the dataset used for the experiments (Section \ref{section:exp-dataset}), the implementation details of the proposed method and the evaluation process (Section \ref{section:exp-evaluation}), and the results obtained from the experiments (Section \ref{section:exp-results}). Furthermore, Section \ref{section:exp-discussion} provides a discussion on the significance of the results and their implications for the proposed method.
\end{ChapAbstract}

\section{Dataset}
\label{section:exp-dataset}

For our experimental analysis, we selected the SumMe dataset as our primary dataset, as it offers a comprehensive collection of 25 videos covering diverse topics and scenarios \cite{SumMe}. This dataset encompasses a wide range of subjects, including aerial footage, sports activities, natural scenery, and urban environments, ensuring the inclusion of various content types.

\begin{figure}[h]
    \centering
    \includegraphics[width=0.73\paperwidth]{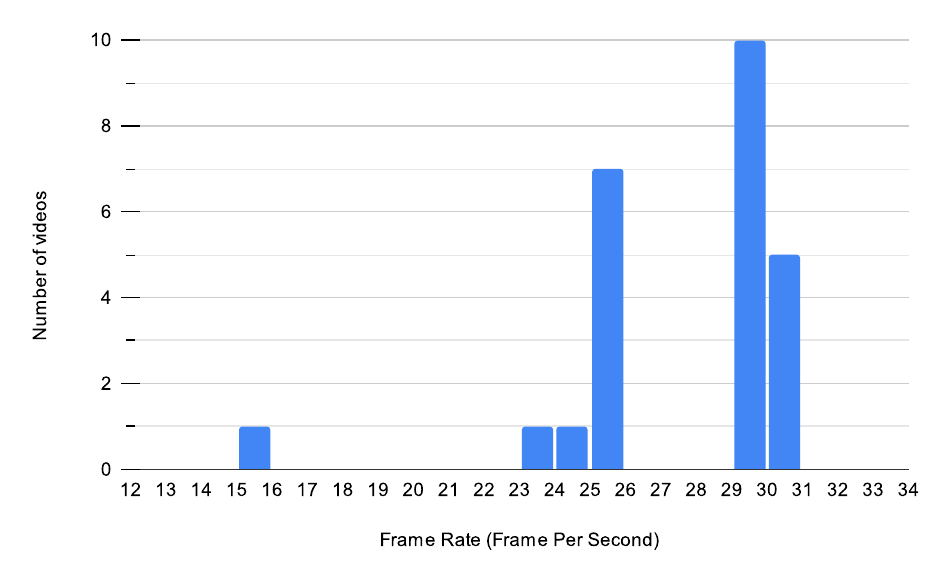}
    \caption{Distribution of frame rates across original videos}
    \label{fig:exp-data-fps}
\end{figure}

The videos in the SumMe dataset exhibit different frame rates, ranging from 15 to 30 frames per second (fps). The distribution of frame rates (Figure \ref{fig:exp-data-fps}) reveals that the most prevalent framerates are 26, 29, and 30 fps, collectively accounting for the majority of the videos in the dataset. 

While higher frame rates generally result in smoother video playback to the human eye, we adopt a different approach in our experiments. Considering that adjacent frames in a video are often very similar, we take advantage of this redundancy to reduce computational overhead. Rather than processing every frame, we choose to subsample the videos at a rate of 4 fps. This subsampling strategy allows us to select informative frames while reducing the overall number of frames to be processed.

\begin{figure}[h]
    \centering
    \includegraphics[width=0.73\paperwidth]{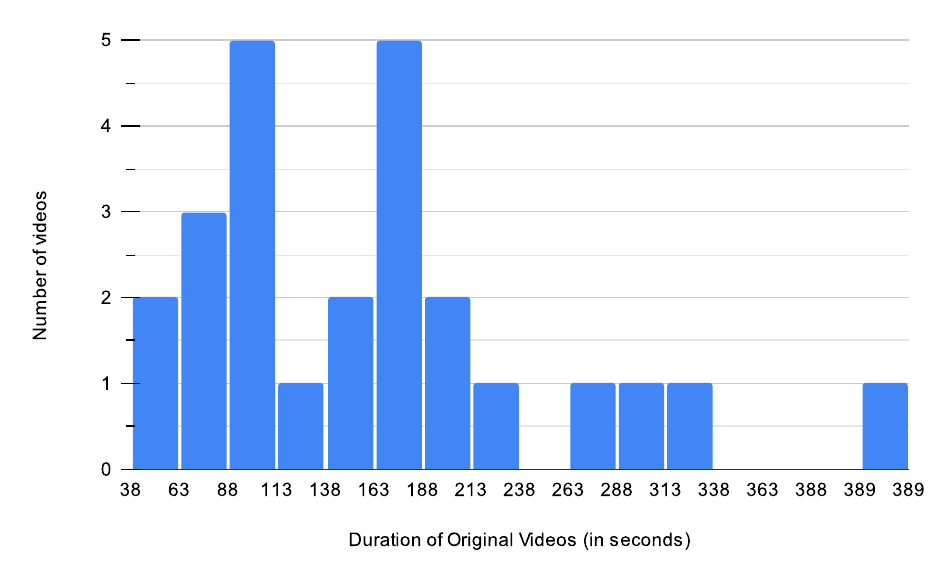}
    \caption{Distribution of original video durations}
    \label{fig:exp-data-duration}
\end{figure}

In terms of duration, the dataset consists of relatively short videos, with an average length of approximately 3 minutes. The shortest video spans 38 seconds, while the longest video extends up to 6.5 minutes. Figure \ref{fig:exp-data-duration} illustrates the distribution of video durations, emphasizing the range of temporal coverage within the dataset.

When performing video summarization, it is essential to consider the temporal length of the original video. Longer videos may require longer summaries to capture the essential content adequately. However, it is important to recognize that human attention spans are limited, and excessively long summaries may not effectively engage viewers. To address this challenge, we introduced Equation \ref{eq:usable-summary-length} to define a limit on the summary length based on the video duration. This ensures that the generated summaries strike a balance between capturing the key information and maintaining viewer engagement within a reasonable time frame.

The ground truth annotations in the SumMe dataset comprise three components: segments, user summaries, and ground truth summaries. The segments are obtained through temporal segmentation approaches, effectively dividing each video into disjoint shots. These segments play a crucial role during the testing and evaluation phases.

Each video in the dataset is annotated by 15 to 18 users, resulting in the generation of 15 to 18 user summaries per video. These user summaries serve as the basis for creating the ground truth summary. Each summary, including the ground truth summary, is represented as a binary array with a length equal to the number of frames in the annotated video.

\begin{figure}[h]
    \centering
    \includegraphics[width=0.73\paperwidth]{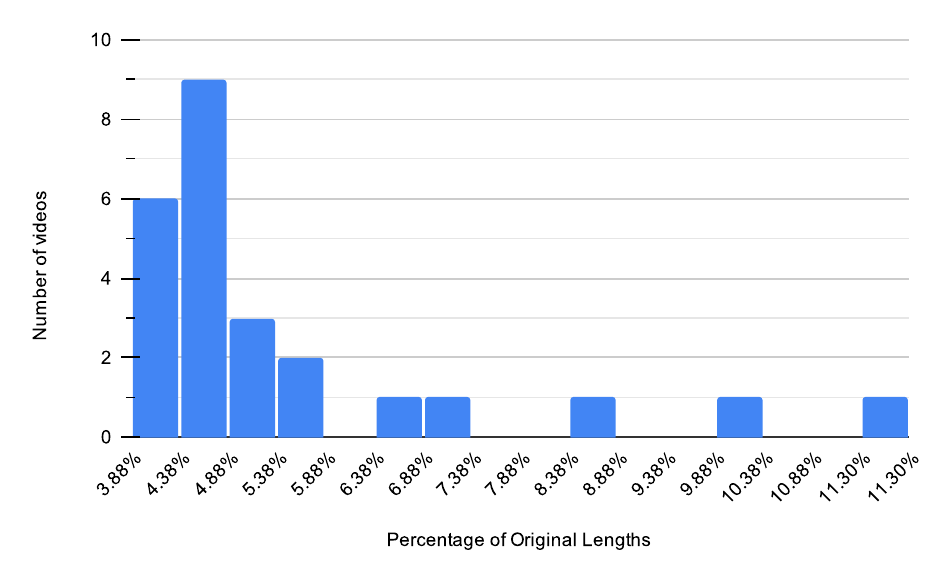}
    \caption{Distribution of the percentage of the original length captured by the ground truth summaries}
    \label{fig:exp-data-summary-length-percent}
\end{figure}

Analyzing the length distribution of the ground truth summaries reveals that the majority of summaries span approximately 4 to 5 percent of the original video length. Figure \ref{fig:exp-data-summary-length-percent} provides a visual depiction of the distribution of the percentage of the original length captured by the ground truth summaries.

The comprehensive nature of the SumMe dataset, coupled with its diverse content, variable frame rates, and detailed ground truth annotations, makes it an ideal choice for evaluating and comparing video summarization algorithms.

In the subsequent sections, we will elaborate on our experimental methodology, leveraging the SumMe dataset to assess the performance and effectiveness of our proposed video summarization approach.
\section{Evaluation Metrics}
\label{section:exp-evaluation}

    The evaluation process for our experiments consists of two components: a \textit{quantitative} measurement and a \textit{qualitative} evaluation. These two parts aim to assess the performance of our proposed method and compare it with baselines from previous works, as well as gather feedback on the quality of the generated summaries from a human perspective.
    
    For the qualitative evaluation, we specifically design a dedicated survey to collect human feedback on the quality of the generated summaries. This survey is constructed to capture the subjective perspective of human evaluators, considering factors such as relevance, coherence, and overall satisfaction with the summaries. The qualitative evaluation allows us to understand how well our proposed method aligns with human expectations and preferences in terms of summarization quality. A previous part in this work (Subsection \ref{subsec:method-eval-implementation}) already delved into the implementation of the survey, outlining the its design and the specific aspects considered for the qualitative assessment.

    With this section, the quantitative evaluation is focused in which we employ the widely adopted metric of f-measure to measure and compare the performance of our proposed method against selected baselines. The f-measure is a commonly used metric in video summarization evaluations, providing a quantitative assessment of the algorithm's effectiveness. This part of the evaluation allows for an objective comparison of our proposed method with existing approaches, providing insights into its raw performance and its ability to generate high-quality summaries. The section will provide detailed information about the automatic evaluation process, including the use of the f-measure metric and the selection of baselines for comparison. 
    
    In previous evaluations of video summarization approaches, the f-measure has been widely adopted as the main metric for performance comparison. In this section, we will revisit the definition of the f-measure to provide a better understanding of this metric and its nature.

    \paragraph[long]{f-measure nature}
        The f-measure metric is used to evaluate a set of predictions on the binary labels of a dataset consisting of multiple samples. To compute the f-measure value for a prediction, each sample in the dataset is assigned to one of four categories: true positive (TP), false positive (FP), true negative (TN), and false negative (FN). The assignment is done by comparing the binary label and the prediction according to the following rules:
        
        \begin{itemize}
            \item If both the label and the prediction are positive, the sample is assigned to TP.
            \item If the label is positive but the prediction is negative, the sample is assigned to FN.
            \item If the label is negative but the prediction is positive, the sample is assigned to FP.
            \item If both the label and the prediction are negative, the sample is assigned to TN.
        \end{itemize}

        Once the assignments are made, the precision value $P$ is computed as the ratio of the number of true positives to the sum of true positives and false positives, i.e., $P = \frac{|TP|}{|TP| + |FP|}$. The recall value $R$ is computed as the ratio of the number of true positives to the sum of true positives and false negatives, i.e., $R = \frac{|TP|}{|TP| + |FN|}$. The f-measure is then defined as the harmonic mean of precision and recall using the formula:

        \begin{equation}
            \label{eq:f-measure}
            F = 2 \times \frac{P \times R}{P + R}
        \end{equation}

        The f-measure combines precision and recall in the evaluation process to ensure that the metric captures the performance of predictions in both positive and negative cases. It provides a balanced evaluation that considers both the ability to correctly identify positive cases (precision) and the ability to capture all positive cases (recall). By using the harmonic mean, the f-measure rewards models that have a good balance between precision and recall.
    
    \paragraph[long]{f-measure in video summarization}
        In the context of video summarization, the f-measure is calculated for each video in the dataset. The prediction corresponds to the summary generated by the method being evaluated, while the label represents the ground truth summary of the video.

        The ground-truth summaries created by users can take different forms, depending on the type of annotations used in the dataset. Here are several forms of ground-truth summaries commonly encountered:

        \begin{itemize}
            \item \textbf{Sets of keyframes~~} The video frames are used as the main units of information for selection. The ground truth summary of a user is a set of keyframes that are chosen as the most important frames in the video. In this case, the ground truth $u_i$ for the $i$-th user is a set of keyframes $\mathbf{u}^{(j)} = \{u_{i}^{(j)}\}^{L}_{i = 1}$, where $u_{i}^{(j)}$ represents the index of the $i$-th keyframe in the video sequence selected by the $j$-th user, and $L$ denotes the number of keyframes selected by the user.
            
            \item \textbf{Sets of key-fragments~~} The video sequence is divided into non-overlapping fragments, with each fragment containing meaningful information. The partitioning method used depends on the dataset's nature and is typically specified along with the dataset. The ground truth summary of a user is a set of key-fragments that are selected as the most important fragments in the video. This form of ground truth is denoted as $u_i$, where $\mathbf{u}^{(j)} = \{u_{i}^{(j)}\}^{L}_{i = 1}$ represents the set of key-fragments selected by the $j$-th user, and $L$ is the number of key-fragments chosen by the user.
            
            \item \textbf{Fragment-level scores~~} Similar to the key-fragment form, the video sequence is partitioned into non-overlapping fragments containing meaningful information. However, in this case, the ground truth summary of a user is a set of scores assigned to each fragment in the video. This form of ground truth is represented as $u_i$, where $\mathbf{u}^{(j)} = \{u_{i}^{(j)}\}^{L}_{i = 1}$ denotes the set of scores assigned by the $j$-th user to each fragment in the video, and $L$ is the number of fragments in the video.
        \end{itemize}

        To evaluate a method using the f-measure on a dataset with one of the above forms of ground truth summaries, the generated summaries of the method are usually converted to the same form as the ground truth summary. For example, if the ground truth summary consists of keyframes, the generated summary is also converted into a set of keyframes. This conversion process involves selecting the most important frames or fragments in the video based on the scores assigned to them by the method. As a result, the pre-evaluation summary of a method for a video is typically in the form $\mathbf{s} = \{t_i\}^{L}_{i = 1}$, where $t_i$ represents the index of the $i$-th frame or fragment in the video sequence selected by the method as key information, and $L$ is the number of frames or fragments selected by the method.

        With the set-theoretic formulation of the generated summary and the user's summary, the four categories required for f-measure calculation can be defined as follows:

        \begin{itemize}
            \item \textbf{True positives~~} Frames or fragments that are selected by both the method and the $i$-th user, denoted as $TP_i = | \mathbf{s} \cap \mathbf{u}^{(i)} |$.
            \item \textbf{False positives~~} Frames or fragments that are selected by the method but not by the $i$-th user, denoted as $FP_i = | \mathbf{s} \setminus \mathbf{u}^{(i)} |$.
            \item \textbf{False negatives~~} Frames or fragments that are selected by the $i$-th user but not by the method, denoted as $FN_i = | \mathbf{u}^{(i)} \setminus \mathbf{s} |$.
            \item \textbf{True negatives~~} Frames or fragments that are not selected by both the method and the $i$-th user, denoted as $TN_i = n - | (\mathbf{s} \cup \mathbf{u}^{(i)})| $ where $n$ is the total number of frames or fragments in the video.
        \end{itemize}

        Based on the above formulation, the f-measure for the $i$-th user is calculated using Equation \ref{eq:f-measure} with $pre_i = \frac{TP_i}{TP_i + FP_i}$ and $rec_i = \frac{TP_i}{TP_i + FN_i}$. This leads to $f_i = 2 \times \frac{pre_i \times rec_i}{pre_i + rec_i}$. Finally, the f-measure for the video is calculated as the aggregation of the f-measure values for all users associating with such video with aggregating operators being either \textit{average} or \textit{max}. The respective formulas are $f = \frac{1}{u} \sum_{i=1}^{u} f_i$) and $f = \max_{i \in [1, u]} f_i$, where $u$ represents the number of users for the given input video in the dataset.

        For convenient of notation in this work, we denote $f(\mathbf{I}, \mathbf{U})$ as the f-measure of a given method for a video $\mathbf{I}$ with user summaries $\mathbf{U}$ in the dataset.
        

    In the computation of f-measure evaluation for our proposed approach, we adopted the multi-split scheme that was previously used in prior works for their evaluation \cite{almeida2012vison, apostolidis2020ac, apostolidis2020unsupervised}. This scheme was originally formulated around the idea of multi-fold cross-validation that randomly splits a dataset with its labels $(\mathbf{X}, \mathbf{Y})$ into a set of M different partitions where each consists of a training set $(\mathbf{X}_t^{(i)}, \mathbf{Y}_t^{(i)})$ and a validation or testing set $(\mathbf{X}_v^{(i)}, \mathbf{Y}_v^{(i)})$ for $i \in [1, M]$ denoting the index of the partition or a split. In the cross-validation practice that are used in the field, a metric is used to evaluate any given method on each and every partition splitted by the above randomness. Thus, f-measure is computed for all splits in this work, meaning that we obtain the f-measure with following formula of \ref{eq:f-measure-split} for the $i$-th split with $\mathbf{X}^{(i)}_j$ and $\mathbf{Y}^{(i)}_j$ are $j$-th video and user summaries in the $i$ split, respectively.

    \begin{equation}
        \label{eq:f-measure-split}
        f^{(i)} = \frac{1}{|\mathbf{X}^{(i)}|} \sum_{j=1}^{|\mathbf{X}^{(i)}|} f(\mathbf{X}^{(i)}_j, \mathbf{Y}^{(i)}_j)
    \end{equation}
    
    Combining the scores from different splits into one single score has been a variety across prior studies. Most of the previous studies have used the average aggregation to compute the final scores for their approaches $f = \frac{1}{M} \sum_{i=1}^{M} f^{(i)}$. Besides that, there are several works which also used the maximum of their splits as an alternative to the average $f = \max_{i \in [1, M]} f^{(i)}$. In this work, we also provide f-measure that is averaged among top 5 videos whose scores are highest in the dataset. In other words, a set $\mathbf{\hat{f}}$ of 5 videos with highest f-measures are selected from the f-measures of all videos, after which the score is computed as that set's average $f = \frac{1}{5} \sum_{i=1}^{5} \mathbf{\hat{f}}_i$
    
\section{Experimental results}
\label{section:exp-results}
    In this section, we present the experimental results of our study, evaluating the performance of our proposed method for video summarization.
    
    \subsection{f-Measure Evaluation}
    \label{subsec:exp-results-fmeasure}
        In the f-measure evaluation, we conduct two comparisons based on our results. First, we compare our best model with prior studies to assess the performance of our proposed method in relation to existing approaches. Second, we compare different settings of our model to investigate the impact of varying parameters on the summarization results.

        
        \subsubsection{Comparison with Prior Studies}
        \label{subsubsec:exp-results-fmeasure-prior}
            
            \begin{table}[ht]
            \centering
            \caption{Comparison of performance in f-Measure (\%) among previous approaches and our method together with their ranking.}
        \begin{tabular}{|cl|c|c|c|}
            \hline
            \multicolumn{2}{|c|}{\textbf{Methods}}                                                                                 & \textbf{F-Score}      & \begin{tabular}[c]{@{}c@{}}\textbf{Rank}\\ \textbf{(Unsupervised)}\end{tabular} & \begin{tabular}[c]{@{}c@{}}\textbf{Rank}\\ \textbf{(General)}\end{tabular} \\ \hline
            \multicolumn{2}{|c|}{Random summary}                                                                 & 40.2           & 7                                                             & 13                                                       \\ \hline
            \multicolumn{1}{|c|}{\multirow{5}{*}{Supervised}}   & SMN \cite{wang2019stacked}                     & \textbf{58.3}  & -                                                             & \textbf{1}                                               \\
            \multicolumn{1}{|c|}{}                              & VASNet \cite{fajtl2019summarizing}             & 49.7           & -                                                             & 10                                                        \\
            \multicolumn{1}{|c|}{}                              & PGL-SUM \cite{apostolidis2021pglsum}             & 57.1           & -                                                             & 2                                                       \\
            
            \multicolumn{1}{|c|}{}                              & H-MAN \cite{liu2019learning}                   & 51.8           & -                                                             & 5                                                        \\
            \multicolumn{1}{|c|}{}                              & SUM-GDA \cite{li2021exploring}                 & 52.8           & -                                                             & 4                                                        \\
            \multicolumn{1}{|c|}{}                              & SUM-DeepLab \cite{rochan2018sequence}          & 48.8           & -                                                             & 12                                                       \\ \hline
            \multicolumn{1}{|c|}{\multirow{6}{*}{Unsupervised}} & CSNet \cite{jung2019discriminative}            & 51.3           & 2                                                             & 6                                                        \\
            \multicolumn{1}{|c|}{}                              & AC-SUM-GAN \cite{apostolidis2020ac}            & 50.8           & 3                                                             & 7                                                        \\
            \multicolumn{1}{|c|}{}                              & CSNet+GL+RPE \cite{jung2020global}             & 50.2           & 4                                                             & 8                                                        \\
            \multicolumn{1}{|c|}{}                              & SUM-GAN-AAE \cite{apostolidis2020unsupervised} & 48.9           & 6                                                             & 11                                                       \\
            \multicolumn{1}{|c|}{}                              & SUM-GDA$_{unsup}$ \cite{li2021exploring}       & 50.0           & 5                                                             & 9                                                        \\
            \rowcolor{lightgray} \multicolumn{1}{|c|}{}                              & The proposed approach                        & \textbf{54.48} & \textbf{1}                                                    & 3                                                        \\ \hline
        \end{tabular}
            \label{table:exp-compare}
            \end{table}
            
            Table \ref{table:exp-compare} presents a comprehensive performance comparison of previous approaches, both supervised and unsupervised, along with the performance of our proposed method. These algorithms have been assessed via the established evaluation approach mentioned in \ref{section:exp-evaluation}. Additionally, the table includes the performance of a random summarizer for reference. This approach serves as a baseline for comparison and helps evaluate the effectiveness of the other methods. The random summarizer assigns importance scores to each frame of a given video using a uniform distribution of probabilities. These fragment-level scores are then utilized to generate video summaries using the Knapsack algorithm, adhering to a length budget of maximum 15\% of the original video's duration. The random summarization process is repeated 100 times for each video, and the overall average score is presented. For more detailed information about this algorithm, please refer to the relevant publication \cite{apostolidis2020performance}.

            The results shown in this table indicate that our unsupervised approach performs remarkably well, despite not incorporating any learnable aspect. In fact, it outperforms existing unsupervised models by at least 3.18\%, showcasing its effectiveness in generating high-quality summaries. Furthermore, compared to state-of-the-art supervised methods, our approach demonstrates competitive performance, surpassing several existing approaches' results.

            It is worth noting that the pre-trained models used in our architecture are originally trained on general datasets about image classification, which might not perfectly align with the distribution of the specific dataset used for this evaluation. Despite this potential mismatch in data distribution, the proposed method exhibits strong performance on the evaluated dataset, providing evidence of the generalizability and adaptability of this training-free framework.

        
        \subsubsection{Ablation Study}
        \label{subsubsec:exp-results-fmeasure-ablation}
            Table \ref{table:exp-ablation-dino} and Table \ref{table:exp-ablation-clip} provide comparisons of the three f-measures, namely \verb|max-f|, \verb|avg-f|, and \verb|top-5|, for different settings employed in our approach. Each table contains two main sections: "Settings" and the score columns. 
            In the "Settings" section, the first column represents the distance measurement used in the clustering module. This distance measurement determines how similarity is calculated between embeddings. The second column represents the methods used for embedding dimension reduction. Specifically, "PCA" and "t-SNE" denote the use of only PCA or t-SNE, respectively, for reducing the dimensionality of the embeddings. On the other hand, "PCA + t-SNE (2)" and "PCA + t-SNE (3)" indicate the combination of PCA and t-SNE in the dimension reduction process, with "2" and "3" referring to the final dimensions of the embeddings obtained from t-SNE.
            The score columns present the evaluation results for each setting. There are three groups of scores associated with three different modes of score aggregation which are mentioned in Section \ref{section:exp-evaluation}. Each group is shown with two columns named as "F-Score" and "Config". The "F-Score" columns display the best scores achieved by the corresponding settings, indicating their performance in video summarization. The "Config" column provides additional information about the intermediate embedding dimensions obtained by PCA in the PCA + t-SNE settings, as well as the final embedding dimensions obtained by PCA in the PCA settings.
            
            \begin{table}[!ht]
            \caption{Comparison of f-measure (\%) across different settings of our approach using DINO pre-trained model}
            \begin{adjustbox}{width=\columnwidth,center}
                \begin{tabular}{|cc|cc|cc|cc|}
                \hline
                \multicolumn{2}{|c|}{\textbf{Settings}}           & \multicolumn{2}{c|}{\textbf{max-f}}          & \multicolumn{2}{c|}{\textbf{avg-f}}          & \multicolumn{2}{c|}{\textbf{top-5}}          \\ \hline
                \multicolumn{1}{|c|}{Distance}                   & Reducer       & \multicolumn{1}{c|}{F-Score}       & Config & \multicolumn{1}{c|}{F-Score}       & Config & \multicolumn{1}{c|}{F-Score}       & Config \\ \hline
                \multicolumn{1}{|c|}{\multirow{5}{*}{Cosine}}    & PCA           & \multicolumn{1}{c|}{58.71}          & 22     & \multicolumn{1}{c|}{46.41}          & 33     & \multicolumn{1}{c|}{67.48}          & 33     \\
                \multicolumn{1}{|c|}{}                           & t-SNE (2)       & \multicolumn{1}{c|}{53.12}          & -      & \multicolumn{1}{c|}{42.76}          & -      & \multicolumn{1}{c|}{59.79}          & -      \\
                \multicolumn{1}{|c|}{}                           & t-SNE (3)       & \multicolumn{1}{c|}{55.46}          & -      & \multicolumn{1}{c|}{42.99}          & -      & \multicolumn{1}{c|}{64.78}          & -      \\
                \multicolumn{1}{|c|}{}                           & PCA + t-SNE (2) & \multicolumn{1}{c|}{56.63}          & 27     & \multicolumn{1}{c|}{50.56}          & 27     & \multicolumn{1}{c|}{69.99}          & 29     \\
                \multicolumn{1}{|c|}{}                           & PCA + t-SNE (3) & \multicolumn{1}{c|}{57.10}          & 14     & \multicolumn{1}{c|}{49.26}          & 5      & \multicolumn{1}{c|}{68.15}          & 5      \\ \hline
                \multicolumn{1}{|c|}{\multirow{5}{*}{Euclidean}} & PCA           & \multicolumn{1}{c|}{54.19}          & 47     & \multicolumn{1}{c|}{44.29}          & 47     & \multicolumn{1}{c|}{61.40}          & 9      \\
                \multicolumn{1}{|c|}{}                           & t-SNE (2)       & \multicolumn{1}{c|}{52.86}          & -      & \multicolumn{1}{c|}{45.09}          & -      & \multicolumn{1}{c|}{67.98}          & -      \\
                \multicolumn{1}{|c|}{}                           & t-SNE (3)       & \multicolumn{1}{c|}{\textbf{59.82}} & -      & \multicolumn{1}{c|}{47.54}          & -      & \multicolumn{1}{c|}{67.91}          & -      \\
                \multicolumn{1}{|c|}{}                           & PCA + t-SNE (2) & \multicolumn{1}{c|}{58.98}          & 8      & \multicolumn{1}{c|}{\textbf{54.48}} & 34     & \multicolumn{1}{c|}{\textbf{72.81}} & 54     \\
                \multicolumn{1}{|c|}{}                           & PCA + t-SNE (3) & \multicolumn{1}{c|}{57.26}          & 41     & \multicolumn{1}{c|}{50.46}          & 31     & \multicolumn{1}{c|}{68.69}          & 39     \\ \hline
                \end{tabular}
            \end{adjustbox}
            \label{table:exp-ablation-dino}
            \end{table}

            Utilizing the DINO pre-trained model for visual representation extraction, our model achieves the highest average F-Score of 54.48\%. This achievement is attributed to the specific settings employed: employing Euclidean distance measurement for clustering and implementing a combination of PCA and t-SNE for dimension reduction. In this configuration, PCA initially reduces the embedding dimension to 34, followed by t-SNE which further reduces the embedding dimension to 2.

            \begin{table}[!ht]
            \caption{Comparison of f-measure (\%) across different settings of our approach using CLIP pre-trained model}
            \begin{adjustbox}{width=\columnwidth,center}
                \begin{tabular}{|cc|cc|cc|cc|}
                \hline
                \multicolumn{2}{|c|}{\textbf{Settings}}           & \multicolumn{2}{c|}{\textbf{max-f}}          & \multicolumn{2}{c|}{\textbf{avg-f}}          & \multicolumn{2}{c|}{\textbf{top-5}}          \\ \hline
                \multicolumn{1}{|c|}{Distance}                   & Reducer       & \multicolumn{1}{c|}{F-Score}       & Config & \multicolumn{1}{c|}{F-Score}       & Config & \multicolumn{1}{c|}{F-Score}       & Config \\ \hline
                \multicolumn{1}{|c|}{\multirow{5}{*}{Cosine}}    & PCA           & \multicolumn{1}{c|}{59.23}          & 30     & \multicolumn{1}{c|}{45.86}          & 46     & \multicolumn{1}{c|}{64.32}          & 30     \\
                \multicolumn{1}{|c|}{}                           & t-SNE (2)       & \multicolumn{1}{c|}{47.79}          & -      & \multicolumn{1}{c|}{44.24}          & -      & \multicolumn{1}{c|}{57.82}          & -      \\
                \multicolumn{1}{|c|}{}                           & t-SNE (3)       & \multicolumn{1}{c|}{49.69}          & -      & \multicolumn{1}{c|}{42.06}          & -      & \multicolumn{1}{c|}{59.00}          & -      \\
                \multicolumn{1}{|c|}{}                           & PCA + t-SNE (2) & \multicolumn{1}{c|}{59.01}          & 13     & \multicolumn{1}{c|}{51.99}          & 9      & \multicolumn{1}{c|}{\textbf{73.50}} & 9      \\
                \multicolumn{1}{|c|}{}                           & PCA + t-SNE (3) & \multicolumn{1}{c|}{\textbf{63.55}} & 51     & \multicolumn{1}{c|}{49.84}          & 51     & \multicolumn{1}{c|}{70.19}          & 23     \\ \hline
                \multicolumn{1}{|c|}{\multirow{5}{*}{Euclidean}} & PCA           & \multicolumn{1}{c|}{54.35}          & 49     & \multicolumn{1}{c|}{46.95}          & 49     & \multicolumn{1}{c|}{66.04}          & 10     \\
                \multicolumn{1}{|c|}{}                           & t-SNE (2)       & \multicolumn{1}{c|}{49.68}          & -      & \multicolumn{1}{c|}{45.78}          & -      & \multicolumn{1}{c|}{66.09}          & -      \\
                \multicolumn{1}{|c|}{}                           & t-SNE (3)       & \multicolumn{1}{c|}{58.93}          & -      & \multicolumn{1}{c|}{45.79}          & -      & \multicolumn{1}{c|}{63.95}          & -      \\
                \multicolumn{1}{|c|}{}                           & PCA + t-SNE (2) & \multicolumn{1}{c|}{57.50}          & 59     & \multicolumn{1}{c|}{50.31}          & 3      & \multicolumn{1}{c|}{70.48}          & 48     \\
                \multicolumn{1}{|c|}{}                           & PCA + t-SNE (3) & \multicolumn{1}{c|}{61.73}          & 32     & \multicolumn{1}{c|}{\textbf{52.33}} & 44     & \multicolumn{1}{c|}{70.77}          & 20     \\ \hline
                \end{tabular}
            \end{adjustbox}
            \label{table:exp-ablation-clip}
            \end{table}

           Conversely, when employing the CLIP pre-trained model, our approach attains its highest average F-Score of 52.33\%. This accomplishment is attributed to the following settings: utilizing Euclidean distance measurement for clustering and implementing a combination of PCA and t-SNE for dimension reduction. Specifically, PCA reduces the embedding dimension to 44 initially, and subsequently, t-SNE further reduces the embedding dimension to 3.

            The results make it clear that the combination of PCA and t-SNE consistently outperforms the use of either PCA or t-SNE alone. Additionally, it's apparent that the Euclidean distance measurement is better suited for our problem, as it consistently yields superior results compared to the cosine similarity measurement.

            As mentioned in \ref{subsec:method-details-emb}, we conducted experiments using different pre-trained models for visual embedding extraction. Table \ref{table:exp-dino-clip} compares the result of the framework using different pre-trained models: dino-b16 and clip-base-16. Both models are base models with a patch size of 16. The "Best Config" column shows the configuration that achieved the best result, including the distance used in the clustering step (Euclidean), the algorithms used for embedding size reduction (PCA and t-SNE), and the dimension of the reduced embeddings represented by the number next to the reducer.

        
            The results presented in Table \ref{table:exp-dino-clip} demonstrate that our proposed framework performs relatively well with various pre-trained models, showcasing its flexibility and efficiency.
        
            The ability to work effectively with different pre-trained models indicates that our approach can leverage a wide range of visual embeddings, making it adaptable to various video summarization scenarios. This flexibility allows practitioners to choose the most suitable pre-trained model based on their specific requirements and available resources.
        
            \begin{table}[!t]
            \centering
            \caption{Comparison of performance (F-Score(\%)) with different embedding pre-trained models}
                \begin{tabular}{|cc|c|}
                \hline
                \multicolumn{2}{|c|}{\textbf{Setting}}                                       & \multirow{2}{*}{\textbf{F-Score}} \\ \cline{1-2}
                \multicolumn{1}{|c|}{\textbf{Embedding Model }} & \textbf{Best Config}       &                                   \\ \hline
                \multicolumn{1}{|c|}{dino-b16}                  & Euclidean PCA (34) + t-SNE (2) & \textbf{54.48}                    \\ \hline
                \multicolumn{1}{|c|}{clip-base-16}              & Euclidean PCA (44) + t-SNE (3) & 52.33                             \\ \hline
                \end{tabular}
            \label{table:exp-dino-clip}
            \end{table}      

        \begin{table}[!t]
        \caption{Ablation study based on the performance (F-Score(\%)) of two variants of the proposed approach on SumMe}
        \centering
            \begin{tabular}{|l|c|}
            \hline
            \multicolumn{1}{|c|}{\textbf{Settings}} & \textbf{F-Score} \\ \hline
            Our method w/o TC            & 46.00                   \\
            Our proposed method           & \textbf{54.48}                 \\ \hline
            \end{tabular}
        \label{table:exp-ablation}
        \end{table}

        To assess the contribution of each core component in our model, we conduct an ablation study, evaluating the following variants of the proposed architecture: variant \textbf{Our method w/o TC}: which is not aware of temporal context by skipping the semantic partitioning stage (Section \ref{subsec:method-model-global-local}), and the full algorithm \textbf{Our proposed method}.


        The results presented in Table \ref{table:exp-ablation} demonstrate that removing the temporal context significantly impacts the summarization performance, thus confirming the effectiveness of our proposed techniques. The inclusion of temporal context enhances the quality of the generated summaries, supporting the superiority of our proposed model.
        
    
    \subsection{Human-centric Evaluation}
        Our evaluation panel comprises 10 assessors, aged between 22 and 25, with a distribution of 4 females and 6 males. All participants are engaged in technology-related occupations.

        The human-centric evaluation results are centered on two primary dimensions. Firstly, we investigate the degree to which evaluators can grasp the original video's content solely through the presented summary video. Secondly, we assess the evaluators' performance in answering questions pertaining to significant video details.
            \begin{table}[h!]
            \centering
            \caption{Average scores (0-10) indicating the extent of original video comprehension if evaluators are presented with the summary video alone}
            \begin{tabular}{|c|c|c|}
            \hline
            Video name                & User summary    & Our summary \\ \hline
            Air\_Force\_One           & 9.50               & 8.00         \\ 
            Base jumping              & 9.50               & 9.00         \\ 
            Bearpark\_climbing        & 8.50               & 10.00        \\ 
            Bike Polo                 & 10.00              & 8.00         \\ 
            Bus\_in\_Rock\_Tunnel     & 9.00              & 9.00        \\ 
            car\_over\_camera         & 7.00               & 8.00         \\ 
            Car\_railcrossing         & 7.00               & 7.00         \\ 
            Cockpit\_Landing          & 8.00               & 9.00         \\ 
            Cooking                   & 7.50               & 8.00         \\ 
            Eiffel Tower              & 5.00               & 7.00         \\ 
            Excavators river crossing & 6.50               & 9.00         \\ 
            Fire Domino               & 7.67               & 6.00         \\ 
            Jumps                     & 7.00               & 9.00         \\ 
            Kids\_playing\_in\_leaves & 7.00               & 9.00         \\ 
            Notre\_Dame               & 7.00               & 9.00         \\ 
            Paintball                 & 7.00               & 9.00         \\ 
            paluma\_jump              & 8.00               & 9.00         \\ 
            playing\_ball             & 5.00               & 10.00        \\ 
            Playing\_on\_water\_slide & 7.00               & 9.00         \\ 
            Saving dolphines          & 6.00               & 9.50         \\ 
            Scuba                     & 8.00               & 9.00         \\ 
            St Maarten Landing        & 9.50               & 9.00         \\ 
            Statue of Liberty         & 8.00               & 8.00         \\ 
            Uncut\_Evening\_Flight    & 10.00              & 8.00         \\ 
            Valparaiso\_Downhill      & 8.00               & 9.50         \\ \hline
            \end{tabular}
            \label{table:exp-human-centric-average-score}
            \end{table}
        \subsubsection{Video Understanding}
            When comparing user-generated summaries with the summaries generated by our method, evaluators are presented with a pair of videos: an original video and its associated summary video. The evaluators are then asked to rate the extent to which they can comprehend the original video based solely on the summary video. The average scores given by the evaluators for each video in the dataset are presented in Table \ref{table:exp-human-centric-average-score}.

            The table clearly demonstrates that in the majority of cases, our method achieves scores that are either higher or comparable to human performance. This suggests that the video summaries generated by our approach are effective in capturing the essential content and facilitating understanding of the original videos.

        \subsubsection{Question Answering}
            \paragraph{Multiple choice questions}
            
            \begin{table}[h!]
            \centering
            \caption{Average accuracy (0-1) comparing answers of multiple choice questions for summary video to answers of the same questions for original video}
            \begin{tabular}{|c|cc|}
            \hline
            \multirow{2}{*}{Video name} & \multicolumn{2}{c|}{Average accuracy}                   \\ \cline{2-3} 
                                        & \multicolumn{1}{c|}{User summary} & Our summary \\ \hline
            Eiffel Tower                & \multicolumn{1}{c|}{0.90}          & 1.00         \\
            Kids\_playing\_in\_leaves   & \multicolumn{1}{c|}{0.80}          & 1.00        \\ \hline
            \end{tabular}
            \label{table:exp-multiple-original-vs-summary}
            \end{table}

            \begin{table}[!h]
            \centering
            \caption{Average IoU score (0-1) comparing answers of multiple choice questions for user summary to answers of the same questions for our summary}
            \begin{tabular}{|c|c|}
            \hline
            Video name                & Average IoU \\ \hline
            Air\_Force\_One           & 1.00        \\
            Base jumping              & 0.87        \\
            Bearpark\_climbing        & 0.93        \\
            Bike Polo                 & 1.00        \\
            Bus\_in\_Rock\_Tunnel     & 0.80        \\
            Car\_railcrossing         & 0.92        \\
            Cockpit\_Landing          & 0.67        \\
            Cooking                   & 0.90         \\
            Jumps                     & 0.90         \\
            playing\_ball             & 0.90         \\
            Playing\_on\_water\_slide & 0.90         \\
            Scuba                     & 0.80         \\
            Uncut\_Evening\_Flight    & 0.90         \\
            Valparaiso\_Downhill      & 0.80         \\ \hline
            \end{tabular}
            \label{table:exp-multiple-user-vs-our}
            \end{table}
            Table \ref{table:exp-multiple-original-vs-summary} and Table \ref{table:exp-multiple-user-vs-our} offer a comparison of the performance between user-generated summaries and our generated summaries on multiple-choice questions.
            Table \ref{table:exp-multiple-original-vs-summary} focuses on the average accuracy of the answers provided in the multiple-choice questions. The answers are compared between the summary videos, including both user-generated summaries and our generated summaries, and the associated original videos. Each question has a set of answer options, and the accuracy is determined by comparing the chosen answer to the ground truth answer from the original video. For example, if the original video's answers consistently indicate "moving" as the correct answer for a question like "Is the point of view standing still or moving?", then "moving" becomes the ground truth for that question. If the answers collected from the user summaries include both "moving" and "standing still", the accuracy for that question in that video would be 0.5, representing 50\% accuracy. The average accuracy for a video is then calculated by averaging the accuracy scores of all the questions for that specific video. It is apparent that our generated summaries show better performance with a small margin when compared to user-generated summaries.
            Table \ref{table:exp-multiple-user-vs-our} provides a comparison of the performance between user-generated summaries and our generated summaries on multiple-choice questions using the Intersection over Union (IoU) metric.
            This metric provides insight into the level of similarity and agreement between the answers obtained from user-generated summaries and our generated summaries for a given set of questions. The higher average IoU scores demonstrate that, in the majority of cases, the answers for user summaries and our generated summaries are very similar.
            
            \paragraph{Checkbox questions}
            
            \begin{table}[h!]
            \centering
            \caption{Average IoU score (0-1) comparing answers of checkbox questions for summary video to answers of the same questions for original video}
            \begin{tabular}{|c|cc|}
            \hline
            \multirow{2}{*}{Video name} & \multicolumn{2}{c|}{Average IoU}                \\ \cline{2-3} 
                                        & \multicolumn{1}{c|}{User Summary} & Our Summary \\ \hline
            Base jumping                & \multicolumn{1}{c|}{0.75}         & 1.00           \\
            Bus\_in\_Rock\_Tunnel       & \multicolumn{1}{c|}{0.5}          & 1.00        \\ \hline
            \end{tabular}
            \label{table:exp-checkbox-orginal-vs-summary}
            \end{table}

            \begin{table}[h!]
            \centering
            \caption{Average IoU score (0-1) comparing answers of checkbox questions for user summary to answers of the same questions for our summary}
            \begin{tabular}{|c|c|}
            \hline
            Video name                & Average IoU \\ \hline
            Air\_Force\_One           & 1.00        \\
            Bearpark\_climbing        & 0.50        \\
            Bike Polo                 & 0.75        \\
            Cockpit\_Landing          & 1.00        \\
            Eiffel Tower              & 1.00        \\
            Jumps                     & 1.00        \\
            Kids\_playing\_in\_leaves & 0.67        \\
            Notre\_Dame               & 1.00        \\
            Paintball                 & 0.50        \\
            paluma\_jump              & 1.00        \\
            playing\_ball             & 1.00        \\
            Scuba                     & 0.00        \\
            Statue of Liberty         & 0.75        \\
            Uncut\_Evening\_Flight    & 1.00        \\
            Valparaiso\_Downhill      & 0.50        \\ \hline
            \end{tabular}
            \label{table:exp-checkbox-user-vs-our}
            \end{table}


            Table \ref{table:exp-checkbox-orginal-vs-summary} and Table \ref{table:exp-checkbox-user-vs-our} provide a comparison of the performance between user-generated summaries and our generated summaries on checkbox questions.

            Table \ref{table:exp-checkbox-orginal-vs-summary} focuses on the average Intersection over Union (IoU) calculated between the original videos and the summary videos, which include both user-generated summaries and our generated summaries.
            Table \ref{table:exp-checkbox-user-vs-our} directly compares the answers provided by the user-generated summaries and our generated summaries for checkbox questions. This comparison allows for a direct evaluation of the similarity between the answers provided by the two types of summaries.

            Based on the results, it is clear that when compared to the ground truth (i.e., the answers for the original videos), our generated summaries demonstrate slightly higher IoU figures compared to those of user-generated summaries. This indicates that our generated summaries exhibit a stronger agreement with the ground truth checkboxes. Moreover, when directly compared to each other, in the majority of cases, the answers provided by user summaries and our generated summaries are very similar. This suggests a high level of agreement and similarity between the selected checkboxes in both types of summaries.

        \subsection{Qualitative Assessment}
                
\begin{figure}[h!]
                \centering
                \includegraphics[width=0.73\paperwidth]{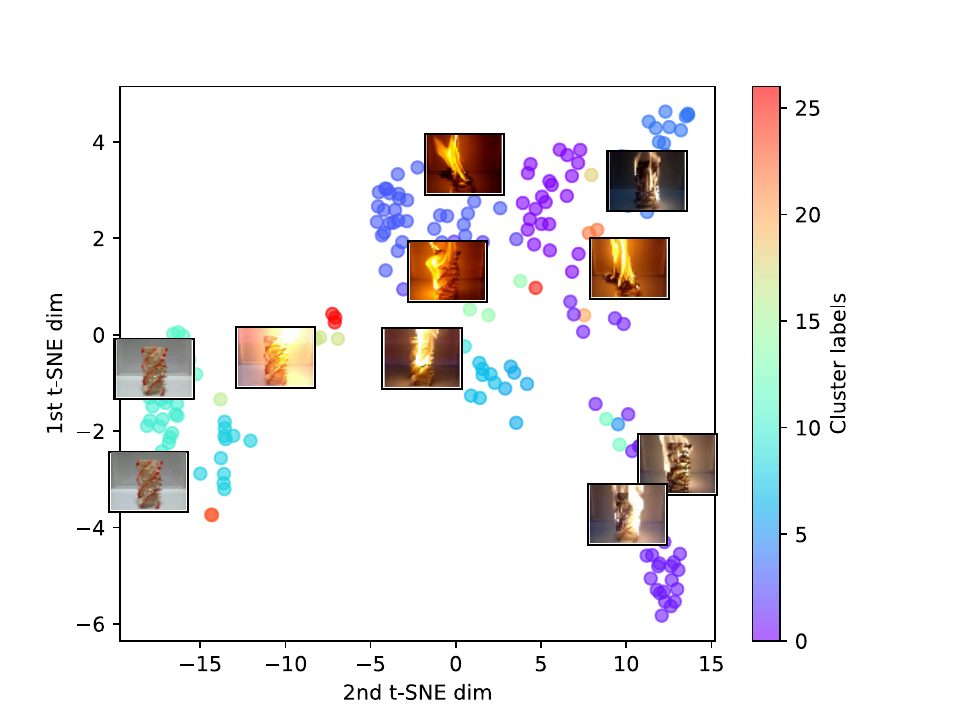}
                \caption{Visualization of a clustering result}
                \label{fig:exp-qualitative-cluster}
            \end{figure}

            Figure \ref{fig:exp-qualitative-cluster} provides a visualization of a clustering result obtained from our proposed method. Each dot represents the embeddings of a frame in the video, while the color of the dot indicates the cluster to which the frame belongs. We have also included some example frames for explanatory purposes.
            
            By examining the plotted frames, it is evident that frames with similar content are positioned close to each other, indicating the effectiveness of the embedding step and validating the performance of the initial stage, the context generation stage.
                
            Furthermore, the clusters with similar color labels exhibit proximity, indicating that the contextual clustering has successfully grouped samples into semantic clusters with high intra-cluster similarity. Visual images belonging to the same cluster share similar semantics.
                
            Additionally, it is noticeable that the dots (representing frames) follow a path in the plotted space, suggesting that frames that are temporally close to each other exhibit similar visual semantics.
            
\begin{figure}[h!]
                \centering
                \includegraphics[width=0.73\paperwidth]{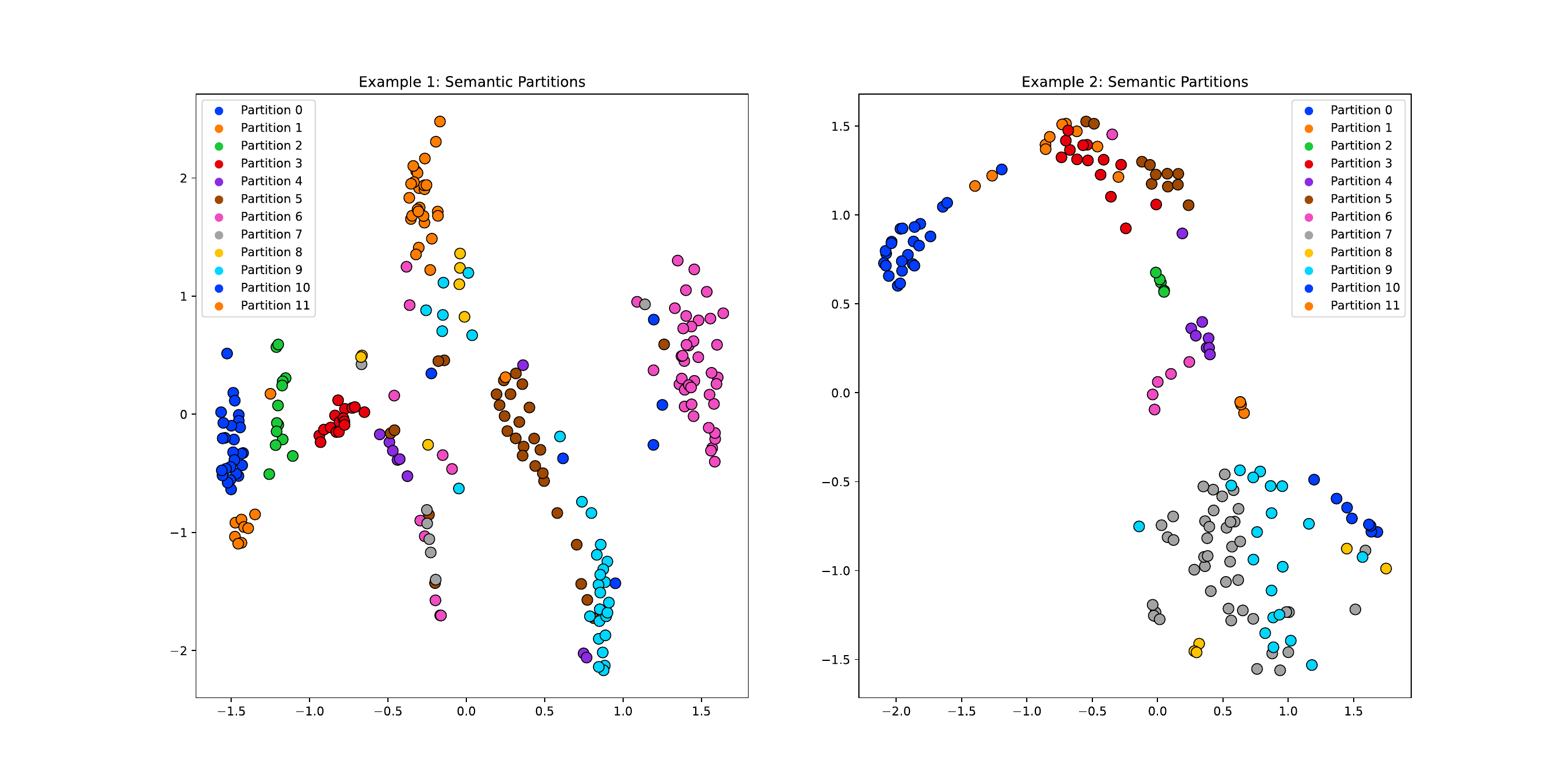}
                \caption{Visualization of partitioning results}
                \label{fig:exp-qualitative-partition}
            \end{figure}

            Figure \ref{fig:exp-qualitative-partition} showcases visualizations of two partitioning results obtained from our proposed method. Each dot represents the embeddings of a frame in the video, and the color of the dot indicates the partition to which the frame belongs.
    
            The partitions consist of frames that belong to the same clusters in consecutive positions along the temporal dimension of the video. This implies that each cluster can contribute to multiple partitions if the video revisits the same cluster after transitioning away from it.
            
            One important observation is that previous works on clustering-based summarization methods often focus on constructing representative keyframes from each cluster without considering the video's temporal dimension. As a result, the resulting keyframes are fragmented and concatenated summaries derived from such keyframes are challenging to comprehend. This limitation arises from the lack of awareness of temporal partitions by prior studies. They primarily view clustering-based methods as baselines for deep network approaches, neglecting the temporal aspect of video summarization.
            
            In contrast, our proposed method addresses this deficiency and incorporates temporal partitioning to improve the overall performance of the summarization process. By considering the temporal dimension and identifying multiple clusters associated with several partitions, we achieve more coherent and comprehensive video summaries.
            
\begin{figure}[h!]
                \centering
                \includegraphics[width=0.73\paperwidth]{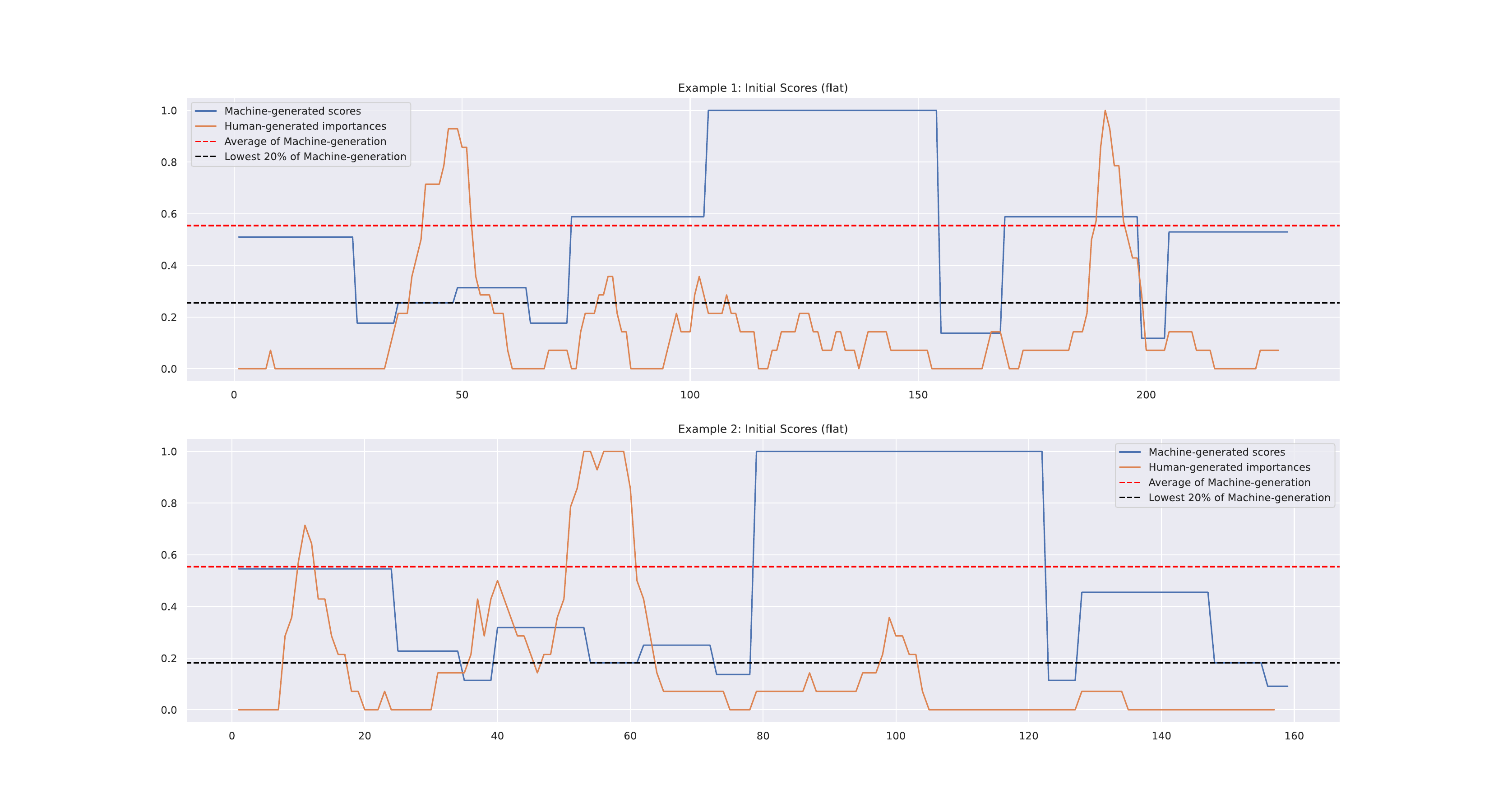}
                
                \caption{Comparison of importance scores between user-annotated scores and flat scores generated by the proposed method}
                \label{fig:exp-qualitative-flat-cmp}
            \end{figure}

            Figure \ref{fig:exp-qualitative-flat-cmp} depicts a comparison between the importance scores obtained through user annotation and the flat scores generated by our proposed method.
    
            The flat results showcase that each computed partition may be associated with one or several peaks in the user summaries, located at different positions within the partition. It is noticeable that most of these peaks tend to occur at the beginning, end, or middle positions of a partition. Furthermore, longer partitions, which have higher flat scores according to our definition, tend to provide more stable estimation of users' peaks. Specifically, all partitions with lengths approximately greater than the average possess at least two medium-sized peaks or one significant peak.
    
            This experimental result also provide insights into the keyframe-biasing method employed in our proposed method, wherein higher importance is assigned to frames that are closer to keyframes. This biasing mechanism is further explained in subsection \ref{subsubsec:method-model-importance}, providing additional details on how the importance scores are computed.
        
\begin{figure}[h!]
                \centering
                \includegraphics[width=0.73\paperwidth]{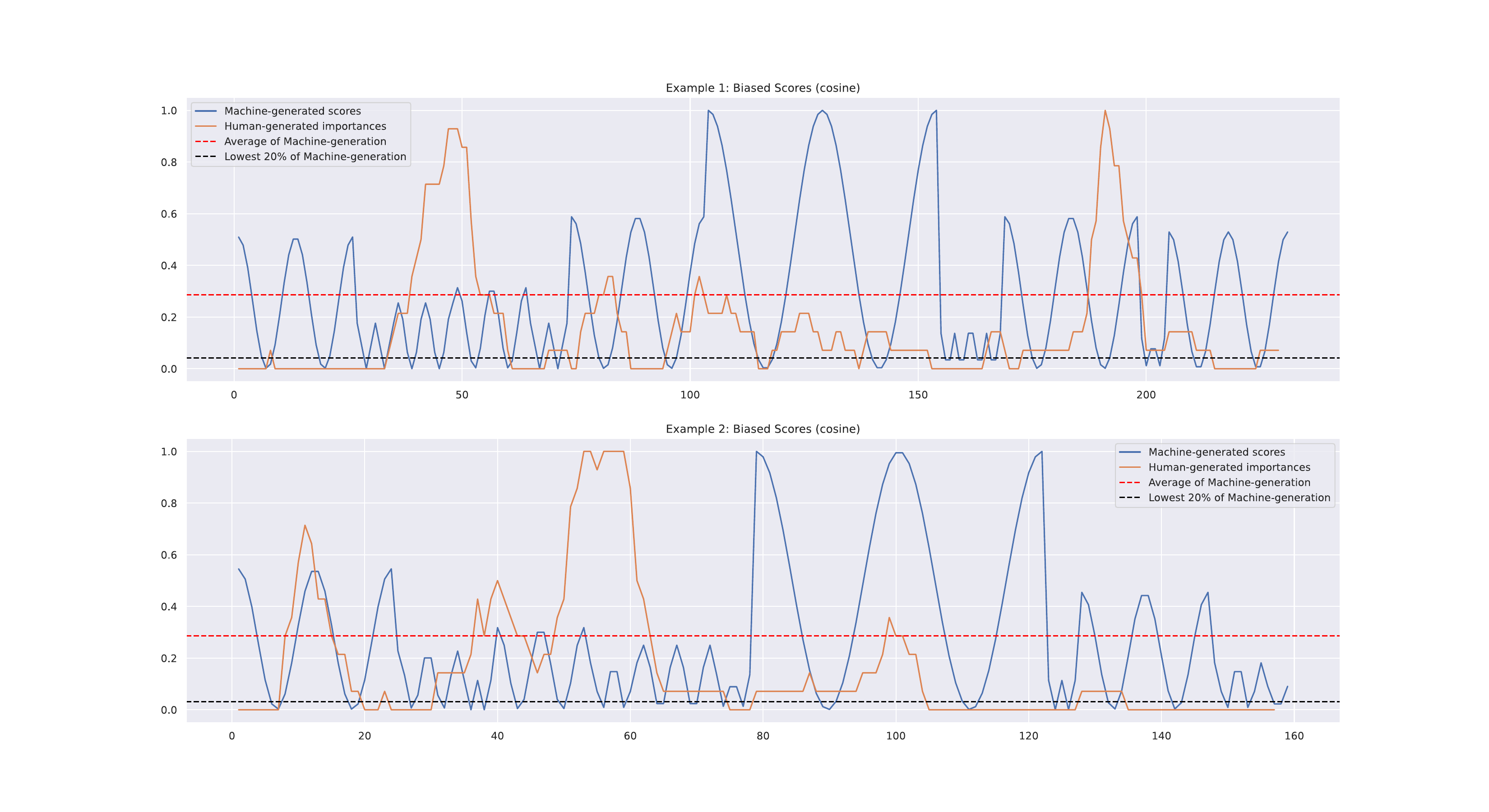}
                \caption{Comparison of importance scores between user-annotated scores and cosine scores generated by the proposed method}
                \label{fig:exp-qualitative-cosine-cmp}
            \end{figure}

            The previous results obtained from the flat scores lead us to the realization that frames with high proximity are more important. To validate this insight, Figure \ref{fig:exp-qualitative-cosine-cmp} is presented to compare the importance scores obtained through user annotation with the cosine scores generated by our proposed method.
            
            The figure reveals that the majority of the peaks in the cosine scores align with the peaks in the scores annotated by users. This observation confirms our hypothesis that frames with high proximity indeed carry greater importance. However, it is important to note that there are some peaks in the user-annotated scores that are not captured by the cosine scores. This discrepancy highlights a limitation of our current approach, indicating that there are certain important frames that are missed by solely considering the proximity-based cosine scores.
            
            This incorrect part of the insight leads us to an interesting idea for future work, which involves the utilization of learnable structures. By incorporating learnable structures into our method, we can potentially improve the accuracy and coverage of the importance scores, allowing us to capture important frames that may not be solely determined by proximity.
    
\begin{figure}[h!]
                \centering
                \includegraphics[width=0.81\paperwidth]{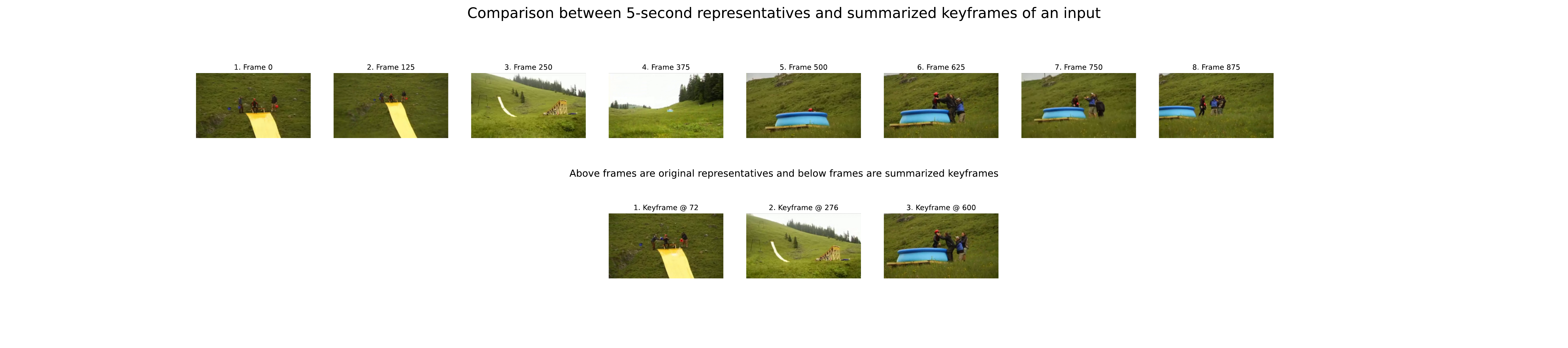}
                \caption{Comparison between the representatives sampled from original video with its summarization as a set of keyframes}
                \label{fig:exp-qualitative-visual}
            \end{figure}

            One of the ways to evaluate the performance of our proposed video summarization method is to compare the keyframes generated by our method with the original frames extracted from the video. To this end, we present a figure that shows a visual comparison between the original frames and the keyframes of a sample video, as shown in Figure \ref{fig:exp-qualitative-visual}. The representative of the original video are extracted every 5 seconds in that video, which demonstrates a man playing a game of sliding down a slope to gain momentum for jumping far into a pool of water and then he is surrounded and asked by his friends and family.
        
            The keyframes used in this comparison are selected as middle frames (using rule \textit{Middle}) of all segments whose importance scores are higher than the average on video-level, which means $v_k > \frac{1}{\hat{T}} \sum_{i = 1}^{\hat{T}} v_i$ for every selected keyframe $k$. This figure consists of two rows of frames, with the first one containing $8$ frames sampled from the video as representatives while the second have $3$ keyframes for demonstration of completeness. The frames on the first row is labeled with their temporal indexes in the video while keyframes are entitled with additional term of \textit{Keyframe}.

            The figure shows that our method is effective in summarizing the given video, as it captures all the information portrayed in the original frames under the keyframes’ visual information. For instance, the first keyframe at index $72$ shows the scene of a man preparing for the play, which corresponds to the first two original frames. In the second keyframe which position is $276$, that man is sliding down the hill and gaining momentum for jumping into a pool of water, which matches the content displayed in the next two representatives. The final keyframe at index $600$ illustrates the jumped man standing on the children swimming pool talking to other people, which relates to the last three sampled frames. The $5$-th representative frame at index $500$ shows the man standing up from that swimming pool which contains no significant information and hence are not required to be covered by the keyframes generated by our method. Thus, the proposed approach produces keyframes that maintain most information conveyed in the original video, demonstrating high preservation of the main content as well as remarkable events. Furthermore, the method also selects keyframes that are diverse and representative, as they show different aspects of the video, such as people (the man and his friends or relatives), objects (the swimming pool and jumping bridge), and actions (preparing or talking). Therefore, our proposal has chosen informative and expressive keyframes that convey the main theme, message, or story of the given video.
    
\section{Discussion}
\label{section:exp-discussion}
    In this section, we will discuss the advantages and disadvantages of our proposed method as well as the proposed human-centric evaluation.

    \subsection{Proposed Method}
    \label{subsec:exp-discussion-method}

        Firstly, let us discuss the advantages and disadvantages associated with our proposed method.

        \subsubsection{Advatanges}
        \label{subsubsec:exp-discussion-method-advantage}

            The proposed method offers several advantages over the prior studies in the published literature, including but not limited to the state-of-the-art approaches. These advantages shall be discussed with the remaining part of this text.

                \paragraph{Improved Performance~} Our method outperforms generative approaches by addressing their limitation of not being able to extrapolate original frames from summarized frames. By accurately representing the original video content, our method provides more informative and comprehensive video summaries.
                
                \paragraph{Minimized Cost of Training~} Our method  does not require any training and hence, it is more applicable to a wider range of realistic applications due to the lack of labeled summary in reality as well as computational resources for re-training or adaptation.
                
                \paragraph{Simplicity in Computation~} The proposed method offers summarization within simpler operations compared to prior state-of-the-art methods, which applies advanced techniques such as Long-Short Term Memory networks \cite{wang2019stacked} or Attention-based Transformer architecture \cite{apostolidis2021pglsum} that complicate the summarizing process.
                
                \paragraph{High Interpretability~} Our approach is constructed with the use of several interpretable components such as the classical clustering algorithms and naive rules for keyframe selection as well as importance scoring. The interpretability provided by these parts help the overall algorithm be easily understood and interpreted by the human.
                
                
                \paragraph{Potential for Further Improvements~} Our method serves as a solid foundation for future enhancements. For example, the current simplistic scoring and keyframe selection rules can be replaced with more sophisticated methods such as PGL-SUM \cite{apostolidis2021pglsum} or attention mechanisms. This opens up opportunities to refine and optimize the summarization process based on data-driven approaches.

            



        \subsubsection{Disadvantages}
        \label{subsubsec:exp-discussion-method-disadvantages}

            Besides multiple advantages provided by our method, it also has certain limitations that are outlined in the text below.
    
                \paragraph{Naive Scoring and Keyframes Rules~} The current scoring and keyframes rules employed in our method may not always accurately predict the importance of frames. This limitation suggests the need for more sophisticated scoring models that can capture the intricate nuances and context of the video content. By incorporating more complex scoring mechanisms, we can potentially enhance the summarization process and generate more accurate and informative summaries.
    
                \paragraph{Limited Data-Driven Improvement~} Our proposed method does not possess the ability to improve in a data-driven way. It relies on predefined rules and lacks the capacity to adapt and learn from data during the summarization process. To overcome this limitation, future research could explore the integration of data-driven approaches, such as machine learning algorithms or attention mechanisms, to enhance the performance and adaptability of the summarization model.
    
    
                
    

    \subsection{Protocol of Human-centric Evaluation}
    \label{subsec:exp-discussion-eval}
         Secondly, let us discuss the benefits and drawbacks of the proposed human-centric evaluation pipeline.
        
        \subsubsection{Benefits}
        \label{subsubsec:exp-discussion-method-benefits}
            The protocol of human-centric evaluation employed in our study gives advantageous benefits compared to prior approaches of evaluating video summarization. Such beneficial aspects are listed in the following paragraphs.
    
    
                \paragraph{Avoid Leaking of Segmentation~} Our evaluation protocol ensures that there is no leaking of segmentation information. In previous evaluations, the algorithm used to generate summaries from frames selected by users is based on a provided set of segments. These segments are also utilized to construct the final summary of any automatic method. This could potentially bias the evaluation results. By avoiding this leakage of segmentation information, our evaluation protocol maintains fairness and integrity in the assessment of automatic summarization methods.
    
                \paragraph{Focus on Informativeness~} Our evaluation protocol places emphasis on evaluating the informativeness of the summaries rather than their interestingness. By assessing users' ability to extract specific facts and information from the summaries, we gain insights into the extent to which the summaries effectively convey the relevant content of the original videos. This approach provides a more meaningful and informative evaluation of the summarization methods.
    
    
    
    
        \subsubsection{Drawbacks}
        \label{subsubsec:exp-discussion-method-drawbacks}
            Despite several benefits demonstrated by the proposed human-centric protocol, it still possesses some drawbacks that need to be acknowledged in the following texts.
    
                \paragraph{High Cost of Evaluation~} Our human-centric evaluation process has a disadvantage related to the labor-intensive nature of creating questions for each video in the dataset. This requirement adds a significant amount of manual effort and time, making the process less scalable, particularly compared to traditional automatic evaluations.
                
    
                \paragraph{Scalability~} The aforementioned requirement for intensive workload poses challenges in terms of scalability, as it becomes increasingly challenging and impractical to replicate the same level of manual effort for larger datasets. This limitation may restrict the evaluation process to a smaller subset of videos or require additional resources to address the scalability issue.
    
    
\chapter{Conclusion}
\label{chapter:conclusion}

\begin{ChapAbstract}
    In this concluding chapter, we provide a comprehensive summary of our work and discuss the limitations of the current approach, thereby setting the stage for future research in the field. We reflect on the key contributions and findings of our work, highlighting the advancements made in the domain of video summarization. Additionally, we acknowledge the drawbacks and challenges associated with our approach, emphasizing the areas that require further exploration and improvement. By addressing these limitations, we aim to inspire and guide future researchers towards developing more effective and innovative solutions for video summarization.
\end{ChapAbstract}

\section{Summary}
\label{section:conc-final}
    
    In this work, we present our novel self-supervised approach for the video summarization task, accompanied by an innovative evaluation pipeline specifically designed for video summarization. Our contributions are outlined as follows:
    
    \begin{itemize}
        \item \textbf{Novel Training-Free Framework}~~~We introduce a novel unsupervised framework that is specifically tailored for the video summarization task. By leveraging the intrinsic structure within videos, our model reduces the reliance on labor-intensive annotated data. This approach enables more efficient video summarization by exploiting the inherent relationships and patterns present in the video data itself. Our unsupervised model offers a promising alternative to traditional supervised methods, demonstrating improved efficiency and effectiveness in generating video summaries.
        \item \textbf{Human-Centric Evaluation Pipeline}~~~Recognizing the importance of real-life usability and user preferences, we propose an evaluation pipeline that is specifically designed to capture human-centric criteria. This pipeline moves beyond traditional evaluation metrics and incorporates aspects that are more relevant and meaningful to human viewers. By aligning the evaluation process with human preferences and expectations, we ensure that the generated video summaries cater to the needs and interests of the intended audience. Our evaluation pipeline provides a comprehensive and holistic assessment of the summarization results, enabling a more accurate evaluation of the algorithm's performance.
    \end{itemize}
    
    Through our experiments and evaluations, we demonstrate the superior performance of our proposed method against unsupervised approaches that are previously studied as well as its competitiveness to existing supervised models. The combination of our self-supervised model and human-centric evaluation pipeline contributes to advancements in the field of video summarization, offering improved efficiency, usability, and alignment with human preferences.

\section{Future Directions}
\label{section:conc-future}
    
    \subsection{Automatic Human-centric Evaluation}
        Currently, our evaluation process involves manually creating questions for each video set, which can be labor-intensive and time-consuming. To alleviate this challenge, we plan to explore the integration of large language models (LLMs) in automating the question generation process.
        
        By leveraging the capabilities of LLMs, such as their natural language processing and understanding abilities, we aim to develop an automated system that can generate tailored questions for each video set. This automated approach would significantly reduce the manual effort required to create questions and streamline the evaluation pipeline.
        
        Furthermore, the use of LLMs can provide additional benefits, such as generating more diverse and creative questions, adapting to user preferences, and allowing for personalized evaluation experiences. This automated human-centric evaluation approach has the potential to enhance the efficiency and scalability of our evaluation process, ultimately contributing to the advancement of video summarization research.
        
        In our future work, we will explore different techniques and methodologies to incorporate LLMs into the question generation process, fine-tuning the models on video summarization-specific data and evaluating their performance against manual question creation. We anticipate that this integration will not only save time and effort but also improve the quality and relevance of the evaluation questions.
    
    \subsection{Learnable Context-Aware Summarization}
    
        The current architecture of our video summarization approach consists of a pre-trained feature extraction module and a context-aware clustering module. As part of our future directions, we plan to enhance the context-aware clustering module by transforming it into a deep neural network with learnable parameters.
        
        Currently, the context-aware clustering module operates based on predefined rules and heuristics, utilizing the extracted features to group frames or segments into clusters. While this approach has shown promising results, introducing a learnable component can further improve the model's ability to capture complex patterns and adapt to different video datasets.
        
        By transforming the context-aware clustering module into a deep neural network, we can leverage the power of gradient-based optimization and enable the model to learn and refine its clustering process based on the specific characteristics of the input videos. This learnable module can automatically adapt its clustering strategy to different video types, lengths, and content, resulting in more accurate and context-aware video summarization.
        
        In our future work, we will explore different architectures and training methodologies for the learnable context-aware summarization module, evaluating its performance against the existing rule-based approach. We anticipate that this transition to a learnable module will lead to improved summarization results and enhanced adaptability to various video datasets.

\addtocontents{toc}{\protect\renewcommand\protect\cftchapfont{\protect\bfseries}}
\bibliographyown{publications}

A derivative paper of the research done in this thesis has been submitted to an international conference which publishes our proposed training-free framework. The information about this paper as well as the publishing conference is detailed below:

\begin{itemize}
    \item \textbf{Paper}: H. Huynh-Lam, N. Ho-Thi, M. Tran, and T. Le, "Cluster-based Video Summarization with Temporal Context Awareness", in \ul{Pacific-Rim Symposium on Image and Video Technology (PSIVT)}, \textit{Accepted for Publication}, 2023.
    \item \textbf{Conference}: Image and Video Technology - 11th Pacific-Rim Symposium, {PSIVT} 2023, Hybrid Event, November 22-24, 2023. Main page is available at \url{https://psivt2023.aut.ac.nz/}, conference ranks \textit{B}.
\end{itemize}


\addtocontents{toc}{\protect\cftpagenumberson{chap}}
\titlespacing*{\chapter}{0pt}{\dimexpr2.5cm-50pt}{\baselineskip}
\bibliography{references}
\titlespacing*{\chapter}{0pt}{-50pt}{\baselineskip}

\appendix



\section*{Supplementary material}

**Table 3.** Comparison of performance (F-Score(%)) with different embedding pre-trained models

| Setting | | F-Score |
|---|---|---|
| **Embedding Model** | **Best Config** | |
| dino-b16 | Euclidean PCA (34) + t-SNE (2) | **54.48** |
| clip-base-16 | Euclidean PCA (44) + t-SNE (3) | 52.33 |

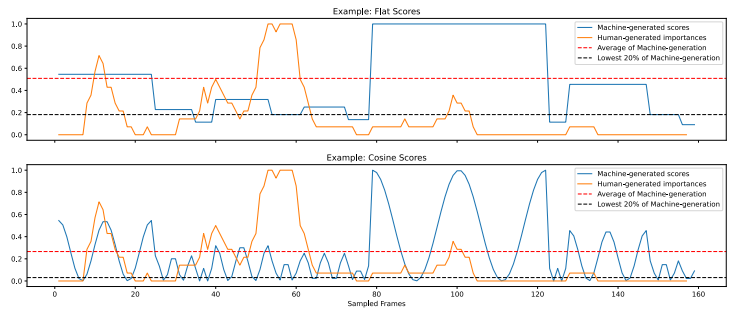

**Fig. 5.** Comparison of importance scores between user-annotated scores and scores generated by the proposed method under the unbiased `flat` rule as well as the biased `cosine` rule.

titioners to choose the most suitable pre-trained model based on their specific requirements and available resources.

### 4.5 Qualitative Assessment

To evaluate the interpretability of the proposed approach, we compared the automatically generated importance scores with those assigned by human annotators. Figure 5 displays the importance scores obtained through averaging human annotations as well as the scores generated by the proposed method. The flat result shows that each computed partition may be associated with one or several peaks in the user summaries, located at different positions within the partition. Longer partitions, which have higher flat scores according to the definition, tend to provide a more stable estimation of users' peaks. The experimental result also provides insights into the keyframe-biasing method employed in the proposed method, wherein higher importance is assigned to frames that are closer to keyframes. This figure reveals that the majority of the peaks in the `cosine` scores align with the peaks of the annotated importance. However, there are some peaks in the users' scores that are not captured by the `cosine` interpolation.

A visual inspection of our method's summarization results is conducted in which a reference video is analyzed against its summary generated through the approach. We present the inspection's result in Figure 6 with the original frames

Comparison between 5-second representatives and summarized keyframes of an input

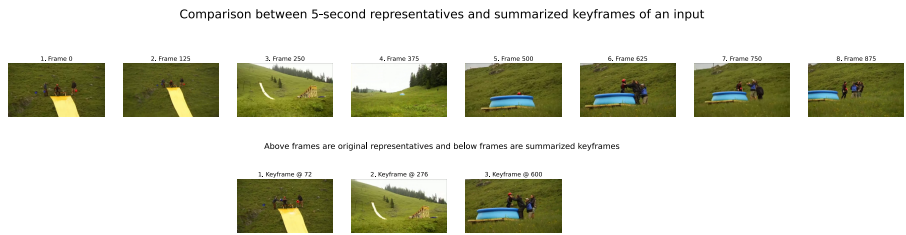

**Fig. 6.** Comparison between the representatives sampled from the original video with its summarization as a set of keyframes.

of the reference video and selected keyframes. The original frames are sampled every 5 seconds from the video, which shows a man playing a game of sliding down a slope and jumping into a pool of water. The keyframes are selected based on their importance scores, which are higher than the average on the video level. Our method preserves the main content and events of the video and selects diverse and representative keyframes that show different aspects of the video. Our method generates informative and expressive keyframes that convey the main theme, message, or story of the video.

### 4.6   Limitations

While the proposed method offers several advantages which have been already illustrated in the experimental results, it also has certain limitations that should be acknowledged. **Naive rules** for scoring and selection of keyframes are being used. Therefore, our current approach may not always accurately predict frame importance. Incorporating more sophisticated scoring mechanisms can enhance the summarization process. **Limited learnability** is demonstrated by our method as it lacks the ability to improve in a data-driven way due to its current reliance on predefined rules. Future research could explore integrating data-driven approaches like machine learning algorithms or attention mechanisms to enhance adaptability.

## 5   Conclusion

In this paper, we introduced TAC-SUM, an unsupervised video summarization approach that incorporates temporal context for generating concise and coherent summaries. The contextual clustering algorithm has successfully partitioned frames into meaningful segments, ensuring temporal coherence. Experimental results show that our method significantly outperforms traditional cluster-based approaches and even is competitive with state-of-the-art supervised methods on the SumMe dataset.

Despite its success, TAC-SUM has limitations related to pre-trained models and data-driven improvement. To address these limitations, future work will focus on integrating learnable components into the model to enhance the summarization process and improve adaptability to various video domains. This includes replacing the current algorithm for contextual clustering with a deep neural network having trainable parameters, enabling the model to capture more complex patterns and adapt to diverse video datasets. Additionally, various architectures and training techniques will be explored to transform the naive rules of importance into a data-driven scoring process, allowing complicated scores to be predicted.



\assignpagestyle{\chapter}{empty}

\end{document}